\PassOptionsToPackage{table}{xcolor}
\pdfoutput=1
\documentclass[11pt]{article}
\usepackage{float}
\usepackage{acl}
\usepackage[utf8]{inputenc}

\usepackage{pgfplots}
\usepackage{dsfont}

\DeclareUnicodeCharacter{2212}{−}
\usepgfplotslibrary{groupplots,dateplot}
\usetikzlibrary{patterns,shapes.arrows}
\pgfplotsset{compat=newest}
\pgfplotsset{width=10cm,compat=1.9}

\usepackage{tikzscale} 
\usepackage[utf8]{inputenc}
\usepackage{bbm}
\newcommand{\probP}{\text{I\kern-0.15em P}}

\usepackage{colortbl} % Include this in the preamble
\usepackage{makecell}

\usepackage{graphicx}

\usepackage{multirow}
\usepackage{array}
\usepackage{amssymb}
\usepackage{pifont}
\usepackage{CJKutf8}
\usepackage{tabularx}
\usepackage{placeins}
\usepackage[normalem]{ulem}
\usepackage{booktabs}
\usepackage[]{algpseudocode}
\usepackage[]{algorithm}
\algtext*{EndFor}%
\algtext*{EndProcedure}%
\usepackage[normalem]{ulem}
\usepackage{times}
\usepackage{latexsym}
\usepackage{adjustbox}

\usepackage[T1]{fontenc}

\usepackage{amsmath}
\usepackage{scalerel,xparse}

\usepackage{cleveref}
\usepackage{microtype}

\usepackage[normalem]{ulem}
\usepackage{fontawesome5}
\usepackage{color}
\usepackage{xspace}
\usepackage{todonotes}
\usepackage{enumitem}
\usepackage[most]{tcolorbox}
\usepackage{subfigure}

\definecolor{bggray}{rgb}{0.95, 0.95, 0.95}
\usepackage[%
    framemethod=tikz,
    skipbelow=\topskip,
    skipabove=\topskip
]{mdframed}
\mdfsetup{%
    leftmargin=0pt,
    rightmargin=0pt,
    backgroundcolor=bggray,
    middlelinecolor=black,
    roundcorner=3
}
\newtcolorbox[list inside=prompt,auto counter,number within=section]{prompt}[1][]{
    colbacktitle=black!60,
    fonttitle=\small,
    coltitle=white,
    fontupper=\footnotesize,
    boxsep=4pt,
    left=0pt,
    top=0pt,
    bottom=0pt,
    boxrule=1pt,
    #1,
}

\usepackage[T1]{fontenc}
\usepackage[russian,english]{babel}

\usepackage{soul}
\sethlcolor{orange!50}
\newcommand{\hlred}[1]{\sethlcolor{red!40}\hl{#1}\sethlcolor{orange!50}}

\title{\textsc{AskQE}: Question Answering as \\
Automatic Evaluation for Machine Translation}

\author{\textbf{Dayeon Ki}\textsuperscript{\ding{68}} \hspace{0.5cm} 
        \textbf{Kevin Duh}\textsuperscript{\ding{83}} \hspace{0.5cm} 
        \textbf{Marine Carpuat}\textsuperscript{\ding{68}} \vspace{0.1cm}  \\
  \textsuperscript{\ding{68}}University of Maryland \hspace{0.5cm}
  \textsuperscript{\ding{83}}Johns Hopkins University \hspace{0.5cm} \\
  \texttt{\{dayeonki,marine\}@umd.edu}
  \hspace{0.5cm}
  \texttt{kevinduh@cs.jhu.edu}
}

\newcommand{\askqe}{\textsc{AskQE} }
\newcommand{\tico}{\textsc{TICO-19} }

\begin{document}
\maketitle

% -------------- Abstract ---------------- %
\begin{abstract}
How can a monolingual English speaker determine whether an automatic translation in French is good enough to be shared? Existing MT error detection and quality estimation (QE) techniques do not address this practical scenario. We introduce \textsc{AskQE}, a question generation and answering framework designed to detect critical MT errors and provide actionable feedback, helping users decide whether to accept or reject MT outputs even without the knowledge of the target language. Using \textsc{ContraTICO}, a dataset of contrastive synthetic MT errors in the COVID-19 domain, we explore design choices for \askqe and develop an optimized version relying on \textsc{LLaMA-3 70b} and entailed facts to guide question generation. We evaluate the resulting system on the \textsc{BioMQM} dataset of naturally occurring MT errors, where \askqe has higher Kendall's $\tau$ correlation and decision accuracy with human ratings compared to other QE metrics.\footnote{We release our code and dataset at \url{https://github.com/dayeonki/askqe}.}

% Human study with monolingual source speakers also confirms that \askqe improves people's decision-making on the acceptance of MT compared to both backtranslation and error annotations.\footnote{Code and data will be publicly available upon publication.}
\end{abstract}

% -------------- Paper Content ---------------- %
\section{Introduction}

% Mistranslation in high stake is more critical, it's difficult for monolingual users to assess MT quality -> so we need QE metrics

% Amid the COVID-19 pandemic, a physician receives a machine-translated medical guideline in French and must decide whether it is safe to share with a French-speaking patient, despite not knowing the language. 

How can a monolingual English speaker determine whether an automatic translation of COVID-19 protocol in French is good enough to be shared? In such high-stakes settings, inaccurate translations can lead to confusion, conversation breakdowns \citep{yamashita-difficulties}, and even life-threatening risks \citep{berger2017israel, VieiraOHaganOSullivan2021, mehandru-reliable-safe}. For instance, even rare inaccurate translations in clinical instructions can pose significant risks to patient health \citep{khoong-demand}. However, accurately assessing Machine Translation (MT) quality is significantly more challenging for monolingual speakers than for bilinguals, as they rely on translations in a language they do not understand \citep{mehandru-etal-2023-physician}. To bridge this gap, we need Quality Estimation (QE) feedback to help users assess MT quality \citep{han-etal-2021-translation}.

\definecolor{acceptgreen}{RGB}{50, 168, 82}

\begin{figure}
    \centering
    \includegraphics[width=\linewidth]{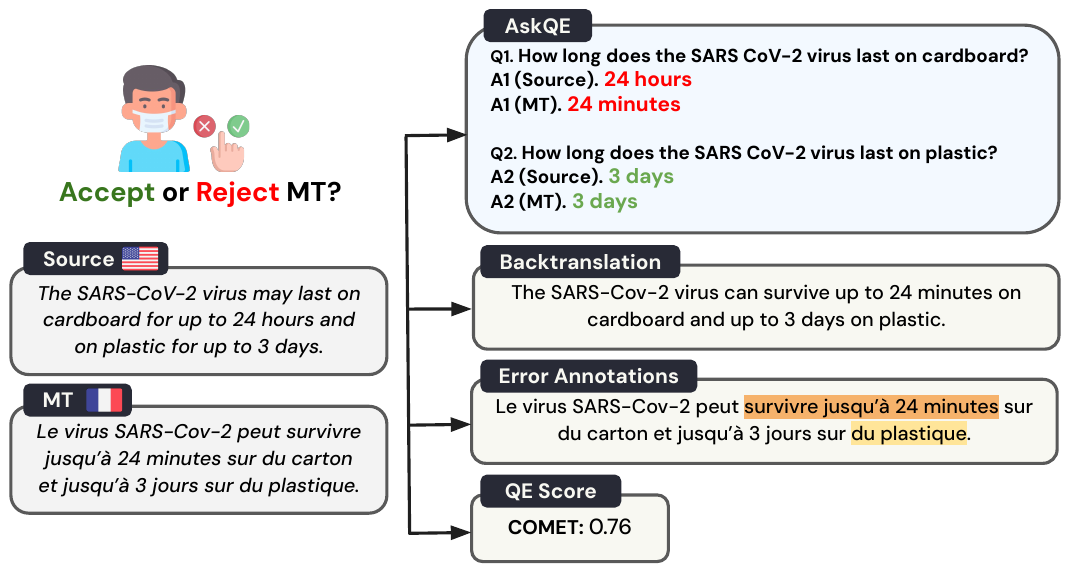}
    \caption{The goal of \textsc{AskQE} is to generate QA pairs that identify critical translation errors and help monolingual source speakers decide whether to \textcolor{acceptgreen}{accept} or \textcolor{red}{reject} MT.}
    \label{fig:teasure_figure}
\end{figure}

% QE feedback mostly lack explainability -> Motivate the problem and our RQ
Research in this space has largely relied on QE metrics that provide segment-level holistic assessments of MT quality, but these can be difficult to interpret and fail to explain how mistranslations impact users. Most QE metrics are primarily trained to produce either a single scalar score \citep{fernandes-etal-2023-devil, fu-etal-2024-gptscore} or error annotations that highlight problematic spans in the target language \citep{kocmi-federmann-2023-gemba, guerreiro-etal-2024-xcomet, lu-etal-2024-error}. This indicates that there is currently no feedback to use at the right granularity to support decision-making for users who do not understand the target language \citep{zouhar-bojar-2020-outbound}. These challenges lead to a key question: How can we identify critical translation errors and provide actionable feedback to help monolingual source speakers decide whether to accept or reject MT in high-stakes contexts?

% \citet{mehandru-etal-2023-physician} further highlight that while QE metrics generally perform well, they often overlook critical errors, and conversely, when they capture critical errors, overall QE suffers.

% Motivate why QG/QA is more natural in machine translation
We hypothesize that asking and answering questions about MT outputs is particularly well suited to address this question by providing a mechanism for explainable MT evaluation. It has been studied and shown effective for assessing factual consistency in related generation tasks such as summarization \citep{durmus-etal-2020-feqa, wang-etal-2020-asking, zhong-etal-2022-towards, deutsch-etal-2021-towards, fabbri-etal-2022-qafacteval}. For MT, questions and answers can provide functional explanations of MT quality, highlighting the functional consequences of potential errors rather than providing a mechanistic explanation of what is wrong \citep{mechanistic_functional}. This approach also aligns with the view of explanations as social, facilitating a knowledge transfer as part of an interaction where the user can weigh the evidence provided in the context of their beliefs \citep{explanation_in_ai}. Question and answers also have the potential to integrate well with techniques people use to estimate whether their interlocuator understood what was said, such as the teach-back techniques that physicians use in cross-lingual communication settings \citep{mehandru-reliable-safe}.

% Introduce AskQE and its main methodology
To explore this direction, we introduce \textbf{\textsc{AskQE}}, a question generation and answering framework based on the idea that a translation is unreliable if key questions about the source text yield different answers when derived from the source or the backtranslated MT. As illustrated in the example in Figure \ref{fig:teasure_figure}, the monolingual English speaker can see that Q1 is answered incorrectly while Q2 is answered correctly. \textsc{AskQE} consists of two key components: \textbf{1)} Question Generation (\textbf{QG}) conditioned on the source sentence and the entailed facts extracted from it (\S \ref{sec:qg}) and \textbf{2)} Question Answering (\textbf{QA}) based on the source and the backtranslated MT (\S \ref{sec:qa}) as in Figure \ref{fig:main_figure}.

% Main results and analysis (core findings of the paper)
First, we validate our approach and explore design choices using \textsc{ContraTICO}, a controlled synthetic dataset. We simulate MT errors by perturbing translations in the \tico dataset \citep{anastasopoulos-etal-2020-tico} across five language pairs (English to Spanish, French, Hindi, Tagalog, and Chinese), creating the \textsc{ContraTICO} dataset (\S \ref{sec:perturbation}). We test different variations of models and information given during QG, and propose an optimized version with \textsc{LLaMA-3 70b} \citep{llama3} and entailed facts to guide QG (\S \ref{sec:res_0}). We then show that \askqe effectively distinguishes minor errors from critical ones (\S \ref{sec:res_1}) and aligns well with established QE metrics (\S \ref{sec:res_2}). 
Second, we show that our findings generalize to naturally occurring MT errors and additional language pairs in \textsc{BioMQM} \citep{zouhar-etal-2024-fine}, achieving comparable Kendall's $\tau$ correlation with human judgments (\S \ref{sec:biomqm_generalize}). Given \textsc{AskQE}'s sensitivity to error severity and correlation with human judgments, we hypothesize that it provides actionable findings, which motivates our decision making simulation experiments. Here, we demonstrate that using \askqe feedback achieves higher decision accuracy than other QE metrics (\S \ref{sec:human_study}). 

% In future works, we envision integrating the QA pairs into interactive translation tools to help users assess and act on MT quality in real time.
\section{Background \& Related Work}

\subsection{QA as (MT) Evaluation}
Using Question Answering (QA) for evaluation has been predominantly studied in the context of summarization, where questions are generated from summaries, and answer pairs from the summary and the source document are compared to assess factual consistency between the original document and its summary \citep{durmus-etal-2020-feqa, wang-etal-2020-asking, zhong-etal-2022-towards, deutsch-etal-2021-towards, fabbri-etal-2022-qafacteval}. In parallel, Question Generation (QG) models have been explored to improve question quality, primarily through pre-training language models \cite{riabi-etal-2021-synthetic, shakeri-etal-2021-towards, dugan-etal-2022-feasibility}. More recently, \citet{fu-etal-2024-qgeval} shows that using Large Language Models (LLMs) such as \textsc{GPT-4} \citep{openai2024gpt4technicalreport} generate higher-quality questions compared to traditional QG approaches using pre-trained models.

On the other hand, the use of QA for MT evaluation has primarily focused on manual evaluation at the system level. Early approaches employ reading comprehension tests to assess the informativeness and usefulness of MT outputs \citep{tomita-etal-1993-evaluation, fuji1999evaluation, fuji2001evaluation} or their readability \citep{jones2005measuring, jones2005measuringa, jones2007ilr}. 
\citet{berka2011quiz} introduce yes/no type questions for manual MT evaluation in the English-Czech language pair, finding that different MT systems produce outputs with varying answer accuracy. \citet{weiss2012error} extend this approach to Polish-English translations, showing that MT outputs with more errors lead to lower answer accuracy. More recent studies focus on using QA for automatic MT evaluation \citep{sugiyama-etal-2015-investigation, scarton-specia-2016-reading, han-etal-2022-simqa, krubinski-etal-2021-just}. The closest work to ours is \textsc{MTEQA} \citep{krubinski-etal-2021-just}, which extracts answer spans and generates questions from the reference translation and compare answer overlap derived from the reference and MT. However, their approach relies on predefined answer spans and reference translations. In a contemporaneous pre-print, \citet{fernandes2025llmsunderstandtranslationsevaluating} show that QA-based evaluation outperforms both neural and LLM-based metrics for ranking paragraph-level translations. We depart from this by exploring the potential of QG/QA framework for more fine-grained, sentence-level MT evaluation with varying levels of error severity.

\begin{figure*}
    \centering
    \includegraphics[width=\linewidth]{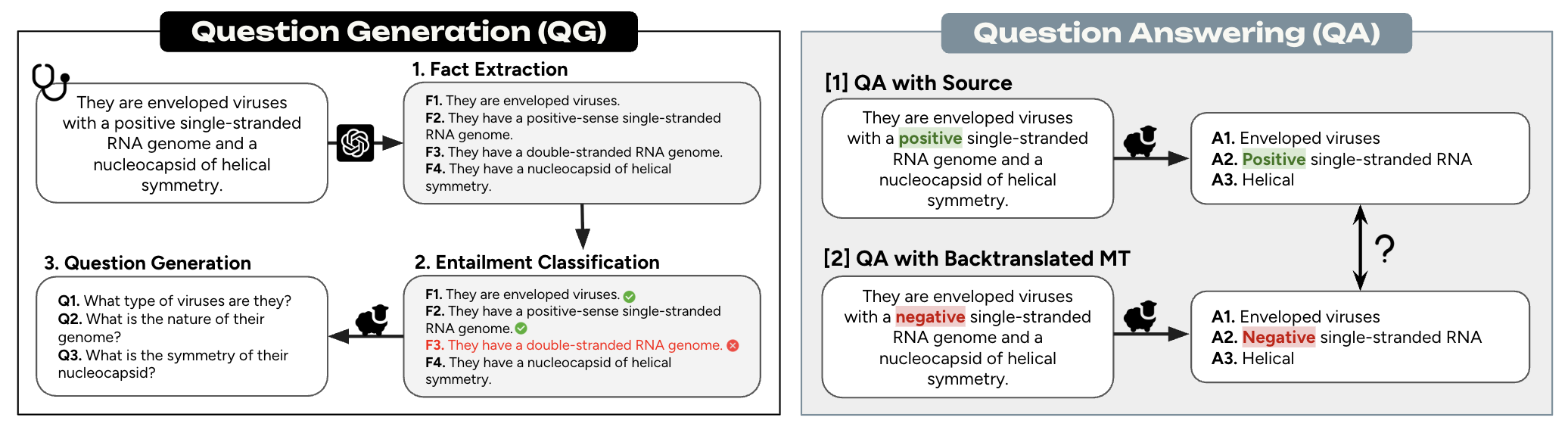}
    \caption{Overview of \textsc{AskQE}. Given source sentence, we \textbf{1) QG:} Generate questions based on the source and entailed atomic facts extracted from it, and \textbf{2) QA:} Answer each question using the source sentence and the backtranslated MT. Discrepancies in the answers are indicators of potential errors in the translation. The same model is used for both QG and QA.}
    \label{fig:main_figure}
\end{figure*}

\subsection{Quality Estimation}
\label{related_work:qe}
Quality Estimation (QE) is the task of automatically assessing MT output quality without relying on human references \citep{specia2018quality}. This form of feedback is particularly important for monolingual source speakers, who often lack both target language proficiency and domain expertise to evaluate MT quality as effectively as professional translators or bilinguals \citep{mehandru-etal-2023-physician}. Most existing QE research has focused on training models to produce a single scalar score \citep{fernandes-etal-2023-devil, fu-etal-2024-gptscore} or error annotations \citep{kocmi-federmann-2023-gemba, guerreiro-etal-2024-xcomet, lu-etal-2024-error} which output error spans in the target language. In parallel, the explainable QE shared task \citep{fomicheva-etal-2021-eval4nlp} frames translation error identification as an explainable QE task, where sentence-level quality judgments are explained by highlighting error-inducing words in the MT \citep{eksi-etal-2021-explaining, rubino-etal-2021-error} or providing textual explanations of metric outputs \citep{fomicheva-etal-2022-translation, jiang2024tigerscore, lu-etal-2024-error}. While we align with these efforts in pursuing more explainable evaluation metrics, we diverge from prior work by focusing on source side annotation through a QG/QA framework.

We compare our method against three established QE metrics: \textbf{1)} \textsc{xCOMET-QE} \citep{guerreiro-etal-2024-xcomet}, \textbf{2)} \textsc{MetricX-QE} \citep{juraska-etal-2024-metricx}, and \textbf{3)} BT-score \citep{agrawal-etal-2022-quality}. Both \textsc{xCOMET-QE} and \textsc{MetricX-QE} evaluate the source and MT output, with \textsc{xCOMET-QE} generating segment-level scores and error annotations and \textsc{MetricX-QE} generating scores. BT-Score assess the similarity between the source and backtranslated MT output using \textsc{BERTScore} \citep{bertscore} as the MT metric.

% \subsection{Biomedical MT}
% WMT biomedical shared tasks have driven the development of MT systems tailored for translating biomedical datasets \citep{neves-etal-2022-findings, neves-etal-2023-findings}. Along this line, several MT datasets in the biomedical domain have been developed, covering a range of subdomains, including COVID-19 pandemic-related texts \citep{anastasopoulos-etal-2020-tico}, clinical trial records\footnote{\url{https://ensaiosclinicos.gov.br/}}, and abstracts from scientific publications.\footnote{\url{https://www.nlm.nih.gov/medline/medline_overview.html}} To evaluate how \askqe identifies critical translation errors at the segment level across various language pairs, we focus on a scenario where monolingual users rely on MT to communicate about COVID-19-related content, using the \tico dataset as our testbed.

\section{\textsc{AskQE}}

We present an overview of \askqe in Figure \ref{fig:main_figure}. We first generate a list of questions conditioned on the source (\S \ref{sec:qg}), generate answers for each question based on the source or the backtranslated MT output (\S \ref{sec:qa}), and compute the answer overlap (\S \ref{sec:evaluation}). All prompts are in Appendix \ref{appendix:prompts}.

% 1. Question Generation
\subsection{Question Generation (QG)}
\label{sec:qg}
Given a source sentence $X_{\mathrm{src}}$, we generate a set of questions $Q_{\mathrm{src}}$ that can be answered based on the sentence. Before generating questions, we extract information from $X_{\mathrm{src}}$ on what to ask questions about and incorporate it as additional context in the prompt. Specifically, to ensure comprehensive coverage of the information from $X_{\mathrm{src}}$, we implement a two-step natural language inference (NLI) pipeline \citep{stacey-etal-2024-atomic}: \textbf{1) Fact extraction}, where we prompt \textsc{GPT-4}o\footnote{\url{https://openai.com/index/hello-gpt-4o/}} to extract atomic facts that can be inferred from the source sentence; \textbf{2) Entailment classification}, where we use an off-the-shelf NLI classifier\footnote{\url{https://huggingface.co/potsawee/deberta-v3-large-mnli}} to assess the binary entailment relationship (entailed or contradictory) between each extracted fact (as the hypothesis) and $X_{\mathrm{src}}$ (as the premise). We discard facts labeled as contradictory, potentially indicating that they cannot be reliably inferred from $X_{\mathrm{src}}$.\footnote{On average, each instance yields 3.61 facts, with 3.08 retained after entailment classification. We show the effect of entailment classification in Appendix \ref{appendix:entailment}.} Finally, we prompt an LLM to generate questions given $X_{\mathrm{src}}$ and the filtered set of entailed atomic facts. Details on other tested QG variants beyond the NLI pipeline are provided in Appendix \ref{appendix:qg_variants}.

% 2. Question Answering
\subsection{Question Answering (QA)}
\label{sec:qa}
We generate answers for each question in $Q_{\mathrm{src}}$ using two different contexts: source sentence and the backtranslated MT output.

\paragraph{QA with Source.}
We provide the source sentence $X_{\mathrm{src}}$ and each question in $Q_{\mathrm{src}}$ as context and prompt an LLM to generate \textit{reference} answers $A_{\mathrm{src}}$, serving as the ground truth for evaluation.

\paragraph{QA with Backtranslated MT.}
Comparing answers derived from the source ($A_{\mathrm{src}}$) and the MT output $Y_{\mathrm{tgt}}$ requires a cross-lingual QA system. However, cross-lingual QA systems may be less accurate than English QA systems, leading to potential disagreements in answer pairs that stem from differences between QA systems rather than translation errors. To mitigate this, we use a monolingual English QA system and instead rely on backtranslation to obtain an English representation of the MT output $Y_{\mathrm{bt}}$. While backtranslation may introduce some noise, we hypothesize that with a high-quality MT system, it is unlikely to mask errors present in the original MT. At worst, it may introduce new errors, potentially making the system overly cautious rather than overlooking critical errors. To this end, we generate \textit{predicted} answers $A_{\mathrm{bt}}$ from $Y_{\mathrm{bt}}$.\footnote{We show that the English QA system outperforms its cross-lingual counterpart in Appendix \ref{appendix:crosslingual_qa}.}

% and compare them against $A_{\mathrm{src}}$.

% We use the backtranslated MT output $Y_{\mathrm{bt}}$ and each question in $Q_{\mathrm{src}}$ as context to generate \textit{predicted} answers $A_{\mathrm{bt}}$, which are then compared against $A_{\mathrm{src}}$. Since $Y_{\mathrm{bt}}$ originates from potentially erroneous MT outputs, we expect $A_{\mathrm{bt}}$ to reflect the errors present in $Y_{\mathrm{bt}}$. We apply backtranlsation to standardize the context language to English before performing QA to ensure that both the reference answers ($A_{\mathrm{src}}$) and predicted answers ($A_{\mathrm{bt}}$) are generated in the same language for direct comparison and reduce varying QA performance across languages \citep{jin2023betteraskenglishcrosslingual} since LLMs often have stronger QA performance in English due to more extensive English data exposed during training \citep{touvron2023llamaopenefficientfoundation, bandarkar-etal-2024-belebele}.

\subsection{\askqe Outputs}
\label{sec:evaluation}

Given the \askqe framework, we can present the QE information in multiple ways. On the one hand, we could simply list all the questions and answer pairs or only those where the answers differ. On the other hand, we could use these pairs to compute a score. While other variants exist, a full exploration is left for future work. In this study, we validate \askqe using a two-step process: \textbf{1)} measuring answer overlap between $A_{\mathrm{src}}$ and $A_{\mathrm{bt}}$ using a similarity metric, and \textbf{2)} aggregating question-answer similarities into a segment-level metric. 
For \textbf{1)}, following \citet{krubinski-etal-2021-just}, we explore several similarity metrics commonly used in QA and MT evaluation. We consider both string-comparison metrics for lexical overlap and neural metric for understanding semantic similarity:
\begin{itemize}[leftmargin=*, itemsep=2pt, parsep=-1pt]
 \item \textbf{Word-level F1} and \textbf{Exact Match (\textsc{EM})}, following prior QA evaluation works \citep{rajpurkar-etal-2016-squad, chen2019evaluating, durmus-etal-2020-feqa}.
 \item \textbf{\textsc{BLEU}} \citep{papineni-etal-2002-bleu} and \textsc{\textbf{chrF}} \citep{popovic-2015-chrf}, widely used in MT evaluation.
 \item \textbf{\textsc{SentenceBERT}} for wordpiece-level embedding similarities \citep{reimers-gurevych-2019-sentence}.
\end{itemize}
For \textbf{2)}, we compute the final \askqe score for a given translation $Y_{\mathrm{tgt}}$ by averaging the similarity scores $D(\cdot,\cdot)$ across all $N$ generated questions:
\begin{equation}
    \text{AskQE}(Y_{\mathrm{tgt}}) = \sum_{i=1}^{N}{\frac{D(A_{\mathrm{src}}, A_{\mathrm{bt}})}{N}}
\end{equation}

%  \item \textbf{\textsc{xCOMET-QE}} \citep{guerreiro-etal-2024-xcomet} as neural metric fine-tuned on MT tasks

\begin{figure*}
    \centering
    \includegraphics[width=\linewidth]{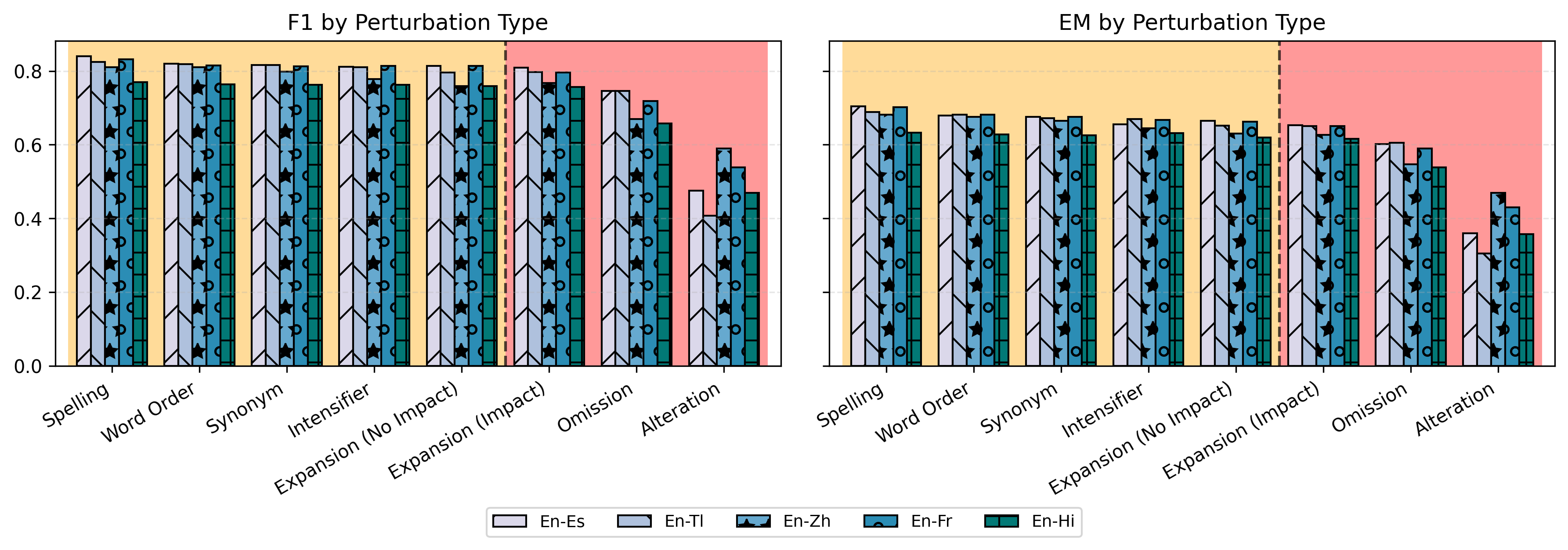}
    \caption{\textsc{AskQE} (\textsc{LLaMA-3 70b} with NLI) using \textsc{F1} and \textsc{EM} metric. In each subplot, the $x$-axis represents perturbation types, while the $y$-axis denotes the corresponding scores. Due to space constraints, we report only the two metrics common in QA research \citep{rajpurkar-etal-2016-squad, deutsch-etal-2021-towards}. Full results for all metrics and models are provided in Appendix \ref{appendix:detailed_askqe}.}
    \label{fig:res_detect_errors}
\end{figure*}

\section{\textsc{ContraTICO}: Controlled Error Generation by Perturbation}
\label{sec:perturbation}

To simulate high-stakes settings, we use \textsc{TICO-19} \cite{anastasopoulos-etal-2020-tico}, a MT dataset containing COVID-19-related content translated from English into 36 languages.\footnote{Detailed dataset statistics, categorized by data source, are outlined in Appendix Table \ref{tab:tico_stats}.} Since the original \tico dataset only provides English source and reference translations in the target language, we construct a dataset with eight synthetic perturbations across five language pairs, \textsc{ContraTICO}, to assess the impact of \askqe design in a controlled setting. Formally, given a reference translation $Y_{\mathrm{ref}}$, we prompt \textsc{GPT-4}o to perturb $Y_{\mathrm{ref}}$ with specific error, which results in a perturbed translation $Y_{\mathrm{tgt}}$. We define eight linguistic perturbations, categorized by their level of severity into either \hl{Minor} or \hlred{Critical}, based on the potential implications of the translation error in practice \citep{freitag-etal-2021-experts}. Typological errors, for instance, have  minimal influence on MT quality \citep{sai-etal-2021-perturbation}, whereas errors that alter the semantics of the translation, such as those introducing misleading information, have a significantly greater impact \citep{karpinska-etal-2022-demetr}. All eight perturbations are applied to each reference translation. We show detailed examples for each perturbation in Appendix Table \ref{tab:perturbation_type}.\footnote{The overlap ratio of perturbed translations between perturbations is reported in Appendix Figure \ref{fig:overlap_ratio}.}

\paragraph{\hl{Minor.}}
These errors do not lead to loss of meaning but introduce small inaccuracies or stylistic inconsistencies that might marginally affect clarity. We carefully design perturbations as minimal pairs with the reference translation (i.e., differ in only one specific aspect) \citep{warstadt-etal-2020-blimp-benchmark}. We define five types of minor perturbations:

\begin{itemize}[leftmargin=*, itemsep=2pt, parsep=-1pt]
 \item \textbf{Spelling:} Misspell one to two words.
 \item \textbf{Word Order:} Reorder words in the sentence. 
 \item \textbf{Synonym:} Replace one word to its synonym.
 \item \textbf{Intensifier:} Modify the intensity of an adjective or an adverb (e.g., \textit{small} to \textit{very small}).
 \item \textbf{Expansion (No Impact):} Expand a word or phrase by adding contextually implied details without introducing new meaning.
\end{itemize}

% While minor errors do not significantly distort the core meaning of the original text, they introduce small inaccuracies or stylistic inconsistencies that might marginally affect clarity. 

%  and are unlikely to confuse or mislead the reader when making decisions based on MT quality

\begin{figure*}
    \centering
    \includegraphics[width=\linewidth]{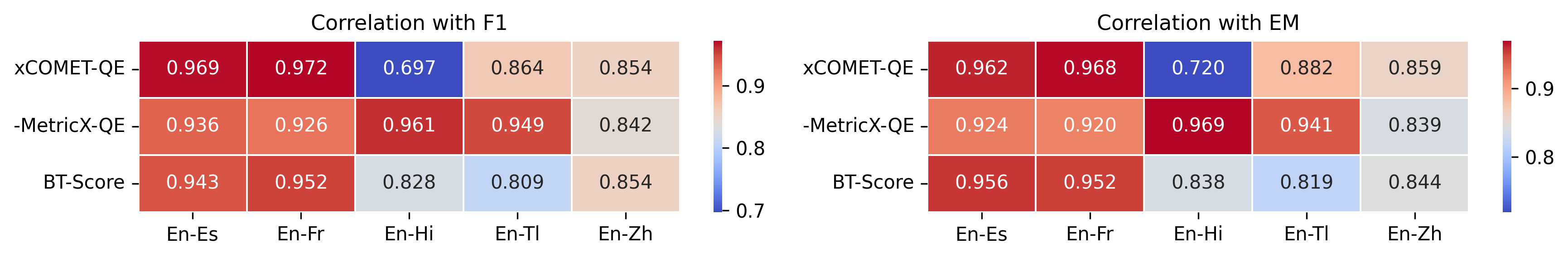}
    \caption{Pearson correlation coefficients between the \textsc{F1} and \textsc{EM} metrics of \textsc{AskQE} and three established QE metrics: \textsc{xCOMET-QE}, \textsc{MetricX-QE}, and BT-Score. We take negatives of \textsc{MetricX-QE}. \askqe exhibits strong correlations with all three QE metrics, confirming its effectiveness as a QE measure. Full numerical results are provided in Appendix \ref{appendix:correlation}.} 
    \label{fig:res_correlation}
\end{figure*}

\paragraph{\hlred{Critical.}}
These errors significantly changes the original meaning and usually appear in a highly visible or important part of the content. We define three types of critical perturbations:

\begin{itemize}[leftmargin=*, itemsep=2pt, parsep=-1pt]
 \item \textbf{Expansion (Impact):} Expand a word or phrase by introducing new meaning.
 \item \textbf{Omission:} Omit a word or phrase. 
 \item \textbf{Alteration:} Alter a word or phrase by changing its original meaning.
\end{itemize}
To validate the impact of different perturbations, we show that \textsc{xCOMET-QE} \citep{guerreiro-etal-2024-xcomet} and cosine similarity scores between the original source $X_{\mathrm{src}}$ and the perturbed translation $Y_{\mathrm{tgt}}$ decrease as severity increases in Appendix \ref{appendix:perturbation_effective}.

% They often occur in visible or key parts of the text. 

\section{Experiment Setup}

\subsection{Dataset}
We use \textsc{ContraTICO} and \textsc{BioMQM} \citep{zouhar-etal-2024-fine} as our testbed.\footnote{Note that we do not consider general MT benchmarks, since our focus is specifically on high-stakes contexts.} \textsc{BioMQM} is a biomedical domain MT dataset with error annotations by professional translators based on the multidimensional quality metrics (MQM) \citep{freitag-etal-2021-experts}.\footnote{Detailed dataset statistics are in Appendix \ref{appendix:biomqm_stats}.} For \textsc{ContraTICO}, we evaluate whether \askqe effectively detects errors, is more sensitive to critical over minor errors, and compare outputs to existing QE methods (\S \ref{sec:res_contratico}). With the \textsc{BioMQM} dataset, we compare \askqe outputs with human error annotations and benchmark its performance against existing QE methods (\S \ref{sec:res_biomqm}).

\paragraph{Language Pairs.}
To align with practical settings, we select language pairs that are high in demand in the United States healthcare system \citep{khoong-demand, amn2021healthcare, systematic-review-impact, pragmatic-assessment}. For \textsc{ContraTICO}, we use English-Spanish (\textsc{en-es}), French (\textsc{en-fr}), Hindi (\textsc{en-hi}), Tagalog (\textsc{en-tl}), and Chinese (\textsc{en-zh}) and for \textsc{BioMQM}, we use English-German (\textsc{en-de}), Spanish (\textsc{en-es}), French (\textsc{en-fr}), Russian (\textsc{en-ru}), and Chinese (\textsc{en-zh}). We fix English as the source language to reflect real-world scenarios where non-English monolinguals rely on translated COVID-19 content originally written in English.

\subsection{Models}
\label{sec:models}

\paragraph{QG/QA.}
We benchmark five English-centric, open-weights LLMs across model sizes and families: \textsc{LLaMA-3 8B} and \textsc{70B} \citep{llama3}, \textsc{Gemma-2 9B} and \textsc{27B} \citep{gemma2}, and \textsc{Yi-1.5 9B} \citep{yi}. Since all prompting in the \askqe pipeline is conducted in English, we hypothesize that English-centric models are better suited for this task than multilingual models.\footnote{HuggingFace model names are in Appendix Table \ref{tab:huggingface_api}.}

\paragraph{Backtranslation.}
We use the Google Translate API for backtranslation due to its efficiency in both time and computation. While backtranslation can introduce noise and has been shown to be an unreliable standalone measure of translation quality \cite{agrawal-etal-2022-quality}, we believe that it minimizes noise relative to other options, as supported by: \textbf{1}) a reasonable average \textsc{xCOMET-QE} score (0.748) between the perturbed translation $Y_{\mathrm{tgt}}$ and its corresponding backtranslation $Y_{\mathrm{bt}}$ (Appendix \ref{appendix:backtranslation_effective}), and \textbf{2}) superior performance compared to a cross-lingual QA system, particularly in distinguishing minor from critical errors (Appendix \ref{appendix:crosslingual_qa}).\footnote{Additionally, we report the Google Translate quality of MT pair in the \tico dataset in Appendix \ref{appendix:tico_gt}.}
\section{Results on \textsc{ContraTICO}}
\label{sec:res_contratico}

% Optimized version of askQE -> llama-70b + NLI
We first optimize the \askqe metric by evaluating five LLMs across three QG variants and selecting \textsc{LLaMA-3 70b} \citep{llama3} with NLI as the best-performing configuration (\S \ref{sec:res_0}). Next, we validate this setup by demonstrating that metric scores decrease for more severe perturbations (\S \ref{sec:res_1}) and exhibits strong correlations with established QE metrics (\S \ref{sec:res_2}).

% Optimizing AskQE
\subsection{Optimizing \textsc{AskQE}}
\label{sec:res_0}
In Appendix Tables \ref{tab:detailed_gemma_9b} to \ref{tab:detailed_yi_9b}, we present detailed results of \askqe by evaluating across five LLMs across three QG variants (Appendix \ref{appendix:qg_variants}), resulting in a total of 15 configurations. We identify the best-performing configuration based on the following criteria: \textbf{1)} Minimal \askqe metric score differences \textit{within} the same error severity level (\hl{Minor}, \hlred{Critical}); \textbf{2)} Large \askqe metric score differences \textit{between} different error severity levels; \textbf{3)} High Correlations with existing QE metrics (Appendix \ref{appendix:correlation}); \textbf{4)} Strong performance in desiderata evaluation, confirming the quality of the generated questions (Appendix \ref{appendix:desiderata}). Each configuration is ranked across these axes, and we select \textsc{LLaMA-3 70b} with NLI as the best-performing method.

% Main results with llama-70b + NLI
\subsection{\askqe can Detect Critical MT Errors}
\label{sec:res_1}
We first validate whether the optimized \askqe with \textsc{LLaMA-3 70b} can effectively differentiate between different levels of perturbation (\hl{Minor}, \hlred{Critical}). As shown in Figure \ref{fig:res_detect_errors}, \askqe consistently assigns lower scores to critical errors compared to minor ones. Among the eight linguistic perturbations examined, spelling errors yield the highest \askqe scores (Avg. \textsc{F1}: 0.815, Avg. \textsc{EM}: 0.682), while alterations show the lowest scores (Avg. \textsc{F1}: 0.496, Avg. \textsc{EM}: 0.384). This trend is consistent across all five language pairs and five metrics used for computing \textsc{AskQE}. In Appendix Tables \ref{tab:qualitative_es} and \ref{tab:qualitative_zh}, we provide qualitative examples for each perturbation level for the \textsc{En-Es} and \textsc{En-Zh} language pair.

% The findings are consistent with prior works showing typological errors have a minimal impact compared to semantic errors having a greater impact on MT quality \citep{sai-etal-2021-perturbation, karpinska-etal-2022-demetr}.

% Correlation with existing QE metrics (xCOMET-QE, MetricX, BT)
\subsection{\askqe Correlates Well with QE Metrics}
\label{sec:res_2}
To validate that \askqe can function as other established QE metrics, we measure Pearson correlation coefficients between \askqe and three existing QE metrics: \textbf{1)} \textsc{xCOMET-QE}, \textbf{2)} \textsc{MetricX-QE}, and \textbf{3)} BT-score (\S \ref{related_work:qe}). As shown in Figure \ref{fig:res_correlation}, both \textsc{F1} and \textsc{EM} scores of \askqe exhibit strong, statistically significant correlations with all three QE metrics. The average correlations are 0.871 and 0.878 with \textsc{xCOMET-QE}, -0.923 and -0.919 with \textsc{MetricX-QE}, and 0.877 and 0.882 with BT-Score for \textsc{F1} and \textsc{EM}, respectively. We report the raw scores for each QE metric in Appendix \ref{appendix:raw_qe}.

Overall, our empirical results in the simulated setting are promising indicators that \askqe effectively detects critical translation errors (\S \ref{sec:res_1}) and correlates well with established QE metrics (\S \ref{sec:res_2}).

\definecolor{mygray}{RGB}{240,240,240} % Light gray background

\begin{table*}
\centering
\resizebox{\linewidth}{!}{%
    \begin{tabular}{p{0.6\textwidth} c m{0.75\textwidth}}  % Changed p{} to m{} for vertical alignment
    \specialrule{1.3pt}{0pt}{0pt}
    \textbf{Question Type} & \textbf{\% Q} & \textbf{Examples} \\
    \toprule
    
    \rowcolor{mygray} \raisebox{-0.2em}{\includegraphics[height=1.1em]{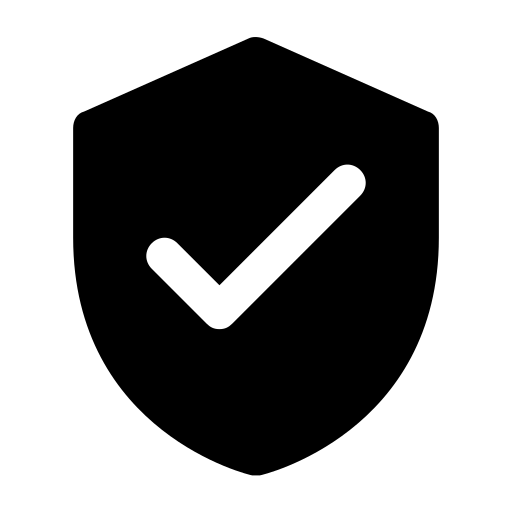}}~\textbf{Verification.} Asking for the truthfulness of an event or a concept. 
    & 15.7 & 
    \begin{tabular}{@{}l@{}} 
        \textbullet~Are many international borders closed? \\ 
        \textbullet~Is Europe the new epicenter of the pandemic? 
    \end{tabular} \\

    \raisebox{-0.2em}{\includegraphics[height=1.1em]{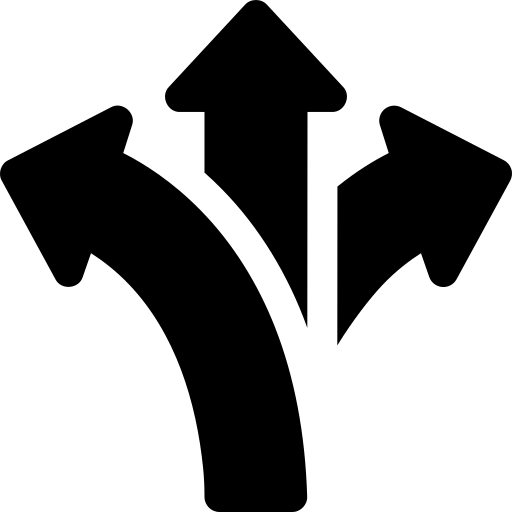}}~\textbf{Disjunctive.} Asking for the true one given multiple events or concepts, where comparison among options is not needed. 
    & 0.71 & 
    \begin{tabular}{@{}l@{}} 
        \textbullet~Is COVID-19 caused by a virus or bacteria? \\ 
        \textbullet~Should I wear a mask indoors or outdoors? 
    \end{tabular} \\

    \rowcolor{mygray} \raisebox{-0.2em}{\includegraphics[height=1.1em]{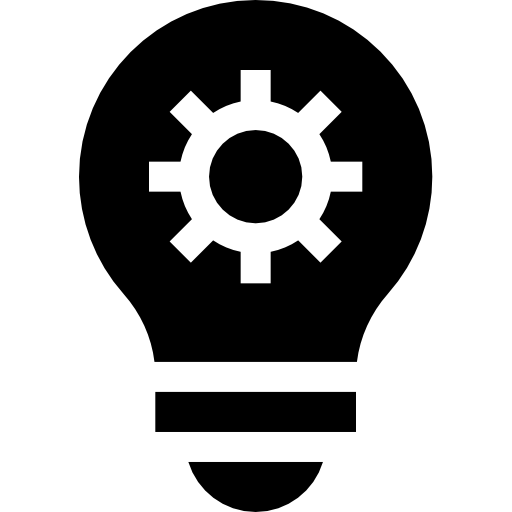}}~\textbf{Concept.} Asking for a definition of an event or a concept. 
    & 23.6 & 
    \begin{tabular}{@{}l@{}} 
        \textbullet~What kind of restrictions have countries imposed on arriving travelers? \\ 
        \textbullet~What is another name for paracetamol?
    \end{tabular} \\

    \raisebox{-0.2em}{\includegraphics[height=1.1em]{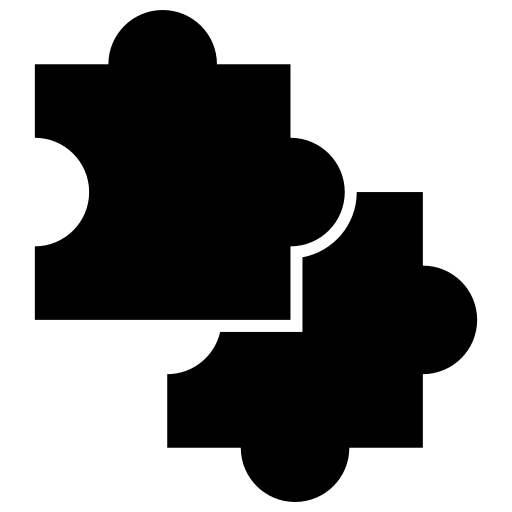}}~\textbf{Extent.} Asking for the extent or quantity of an event or a concept. 
    & 24.6 & 
    \begin{tabular}{@{}l@{}} 
        \textbullet~How many COVID-19 cases were reported today? \\ 
        \textbullet~In how many countries has the healthcare system been stretched?
    \end{tabular} \\

    \rowcolor{mygray} \raisebox{-0.2em}{\includegraphics[height=1.1em]{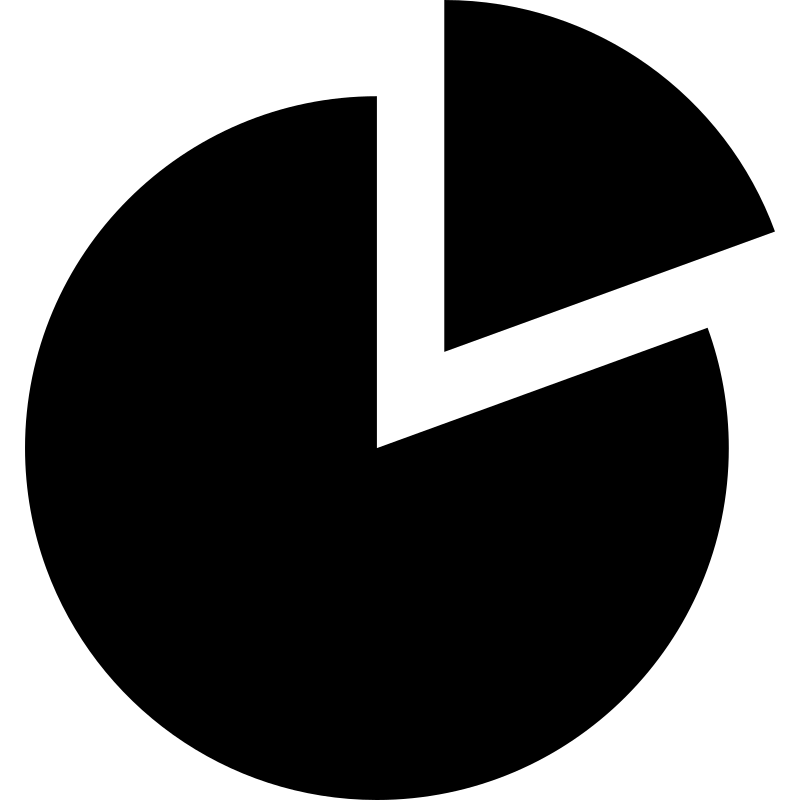}}~\textbf{Example.} Asking for example(s) or instance(s) of an event or a concept. 
    & 1.88 & 
    \begin{tabular}{@{}l@{}} 
        \textbullet~What types of events are being canceled worldwide? \\ 
        \textbullet~What are the sources for further information on the coronavirus outbreak?
    \end{tabular} \\

    \raisebox{-0.2em}{\includegraphics[height=1.1em]{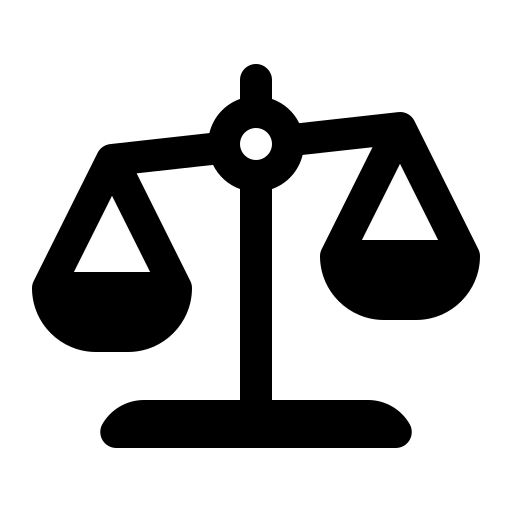}}~\textbf{Comparison.} Asking for comparison among multiple events or concepts. 
    & 1.68 & 
    \begin{tabular}{@{}l@{}} 
        \textbullet~How do countries differ in their testing practices? \\ 
        \textbullet~What is the difference between formal education and informal education?
    \end{tabular} \\

    \rowcolor{mygray} \raisebox{-0.2em}{\includegraphics[height=1.1em]{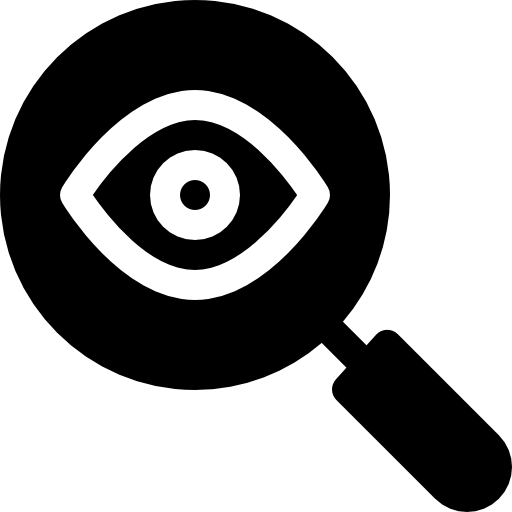}}~\textbf{Cause.} Asking for the cause or reason for an event or a concept. 
    & 14.4 & 
    \begin{tabular}{@{}l@{}} 
        \textbullet~What is the cause of COVID-19? \\ 
        \textbullet~What is the cause of the severe shortage of test kits in many countries?
    \end{tabular} \\

    \raisebox{-0.2em}{\includegraphics[height=1.1em]{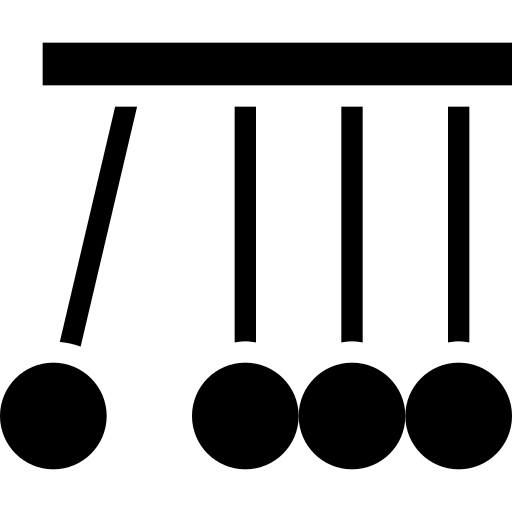}}~\textbf{Consequence.} Asking for the consequences or results of an event. 
    & 2.72 & 
    \begin{tabular}{@{}l@{}} 
        \textbullet~What are serious complications of the disease? \\ 
        \textbullet~What happens to the people who were sitting near an infected person?
    \end{tabular} \\

    \rowcolor{mygray} \raisebox{-0.2em}{\includegraphics[height=1.1em]{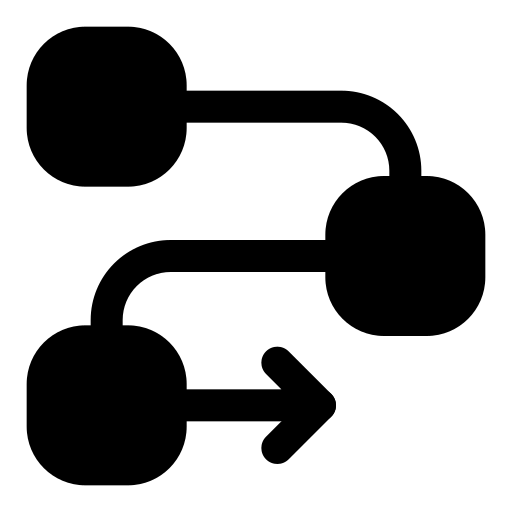}}~\textbf{Procedural.} Asking for the procedures, tools, or methods by which a certain outcome is achieved. 
    & 14.0 & 
    \begin{tabular}{@{}l@{}} 
        \textbullet~What hygiene practices should be followed on a plane? \\ 
        \textbullet~What steps is the Office of Emergency Services directed to take?
    \end{tabular} \\

    \raisebox{-0.2em}{\includegraphics[height=1.1em]{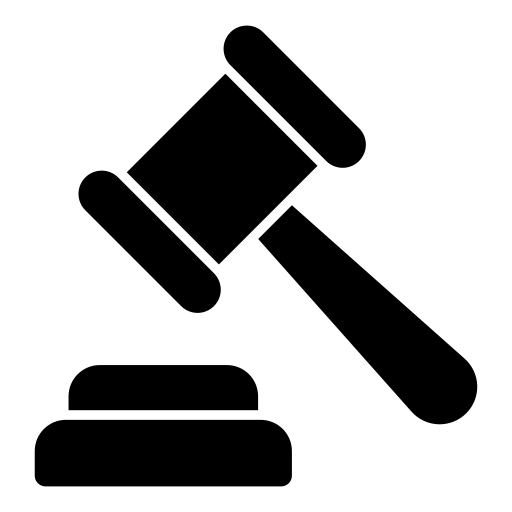}}~\textbf{Judgmental.} Asking for the opinions of the answerer's own. 
    & 0.66 & 
    \begin{tabular}{@{}l@{}} 
        \textbullet~Should hourly data be our current approach? \\ 
        \textbullet~Can you still experience events without traveling?
    \end{tabular} \\    

    \specialrule{1.3pt}{0pt}{0pt}
    \end{tabular}
}
\caption{Example questions from \askqe classified according to the question taxonomy from \citet{cao-wang-2021-controllable}.} 
\label{tab:question_categorization}
\end{table*}

\subsection{Question Analysis}
\label{sec:question_categorization}

% What types of questions are generated? - according to Q typology
\paragraph{Type Categorization.}
We categorize the types of questions generated during QG according to the 10 pragmatic function types defined by \citet{cao-wang-2021-controllable}, using a few-shot LLM prompting classifier.\footnote{Following \citet{trienes-etal-2024-infolossqa}, we prompt \textsc{GPT-4}o with annotation guidelines and few-shot examples from \citet{cao-wang-2021-controllable} (Appendix \ref{appendix:categorization_prompt}).} We present the distribution of question types, along with their definitions and examples in Table \ref{tab:question_categorization}. The two most common question types are \textbf{1) \raisebox{-0.2em}{\includegraphics[height=1.1em]{figure/logos/extent.png}} Extent:} questions asking about the extent or quantity of an event or a concept (24.6\%), and \textbf{2) \raisebox{-0.2em}{\includegraphics[height=1.1em]{figure/logos/concept.png}} Concept:} questions seeking the definition of an event or a concept (23.6\%). This distribution suggests a strong domain effect, as the \tico dataset primarily consists of COVID-19-related articles and public announcements (Appendix Table \ref{tab:tico_stats}). These texts frequently report numerical data (e.g., case numbers, percentages) and define key medical and epidemiological terms, naturally leading to a higher proportion of Extent (e.g., ``\textit{How many COVID-19 cases were reported today}?'') and Concept (e.g., ``\textit{What is another name for paracetamol}?'') questions. Conversely, there are fewer \raisebox{-0.2em}{\includegraphics[height=1.1em]{figure/logos/judgmental.png}} Judgmental (0.66\%) and \raisebox{-0.2em}{\includegraphics[height=1.1em]{figure/logos/disjunctive.png}} Disjunctive (0.71\%) questions, which aligns with the fact that news articles and official announcements tend to be objective and neutral, rather than expressing subjective opinions or presenting binary choices.

% Intermediate analysis measuring quality of generated Qs
\paragraph{Quality.}
We further conduct a fine-grained analysis to quantitatively assess the quality of the questions generated by \textsc{AskQE}. To this end, we define five quality desiderata that measures the correctness, diversity, readability, and answerability of the questions, as detailed in Appendix Table \ref{tab:desiderata}. On average, 3.37 questions are generated per instance\footnote{The average instance length is 24.79 words.}, with no cases of empty sets or duplicate questions. The diversity of questions within each instance, measured by the average \textsc{SentenceBERT} similarity \citep{reimers-gurevych-2019-sentence}, is 0.634. Readability, measured using the Flesch Reading Ease score \citep{Flesch1948}, is 65.89 with fairly standard level despite the presence of domain-specific terminology. The average answerability score of each question with respect to the source sentence, measured by \textsc{SelfCheckGPT} \citep{manakul-etal-2023-selfcheckgpt}, is 92.92, indicating that the generated questions remain highly faithful to the source context.
% We extend our analysis with a fine-grained examination of the type and quality of generated questions (\S \ref{sec:question_categorization}), evaluate \textsc{AskQE}'s generalizability in a more naturally occurring MT error setting (\S \ref{sec:biomqm}), and simulate human decision-making based on different types of QE feedback (\S \ref{sec:human_study}).
\definecolor{neutralyellow}{RGB}{252, 186, 3}
\definecolor{minororange}{RGB}{247, 148, 0}
\definecolor{majorred}{RGB}{214, 0, 0}
\definecolor{criticalbrown}{RGB}{107, 0, 12}

\section{Results on \textsc{BioMQM}}
\label{sec:res_biomqm}

Our initial experiments on the \textsc{ContraTICO} dataset were conducted in a controlled setting. We extend this analysis with the \textsc{BioMQM} dataset with more naturally occurring translation errors and additional language pairs.

\subsection{\askqe Generalizes to \textsc{BioMQM}}
\label{sec:biomqm_generalize}
\paragraph{Error Detection.}
Since \textsc{BioMQM} dataset contains multiple errors (1.95 on average) per segment, we assign the error severity of an MT output based on its highest level error (e.g., if the MT contains both critical and minor errors, it is categorized as critical). As shown in Table \ref{tab:biomqm_main}, the average \askqe scores with all metrics progressively decreases as error severity increases from \textcolor{neutralyellow}{\textbf{Neutral}} to \textcolor{criticalbrown}{\textbf{Critical}}. On average, 3.40 questions are generated per instance, with an average diversity score of 0.656 and an answerability score of 91.72, measured using the same metrics as in Section \ref{sec:question_categorization}. We further confirm that this trend holds across individual language pairs, as detailed in Appendix \ref{appendix:biomqm_detailed}. In sum, our findings confirm that \textsc{AskQE} effectively distinguishes between different error severity levels even with more naturally occurring MT errors.

\definecolor{neutralyellow}{RGB}{252, 186, 3}
\definecolor{minororange}{RGB}{247, 142, 5}
\definecolor{majorred}{RGB}{214, 0, 0}
\definecolor{criticalbrown}{RGB}{107, 0, 12}

\begin{table}
\centering
\resizebox{\linewidth}{!}{%
    \begin{tabular}{llllllll}
    \specialrule{1.3pt}{0pt}{0pt}
    \textbf{Severity} & \textbf{\textsc{F1}} & \textbf{\textsc{EM}} & \textbf{\textsc{chrF}} & \textbf{\textsc{BLEU}} & \textbf{\textsc{SBERT}} \\
    \toprule

    \textcolor{neutralyellow}{\textbf{Neutral}} & 0.736 & 0.426 & 78.41 & 61.63 & 0.830 \\
    \textcolor{minororange}{\textbf{Minor}} & 0.720 & 0.414 & 76.24 & 58.44 & 0.825 \\
    \textcolor{majorred}{\textbf{Major}} & 0.700 & 0.406 & 75.27 & 56.59 & 0.823 \\
    \textcolor{criticalbrown}{\textbf{Critical}} & 0.688 & 0.382 & 74.32 & 54.75 & 0.819 \\
    
    \specialrule{1.3pt}{0pt}{0pt}
    \end{tabular}
}
\caption{\textsc{AskQE} scores using different metrics evaluated on \textsc{BioMQM} per error severity. \textbf{\textsc{SBERT}}: \textsc{SentenceBERT}.}
\label{tab:biomqm_main}
\end{table}

\begin{figure}[t!]
    \centering
    \includegraphics[width=\linewidth]{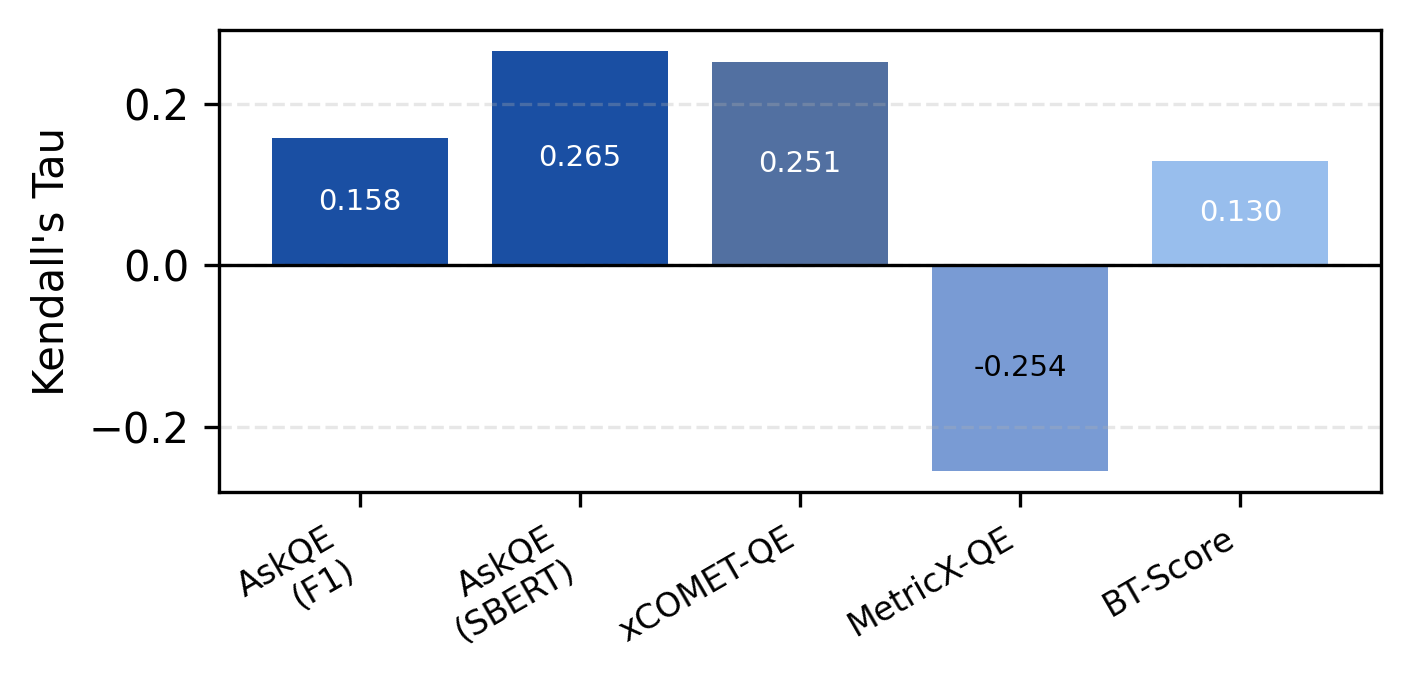}
    \caption{Segment-level correlation (Kendall's $\tau$) between QE metrics and human judgments in the \textsc{BioMQM} dataset.}
    \label{fig:kendalls_tau}
\end{figure}

\paragraph{Correlation.}
Following \citet{zouhar-etal-2024-fine}, we measure segment-level Kendall's $\tau$ correlation between QE metrics and human judgments. We use the professional error annotations from \textsc{BioMQM}, and compute a human judgment score based on the schema from \citet{freitag-etal-2021-experts}. We compare the same neural QE metrics against \textsc{AskQE} as in our \textsc{ContraTICO} evaluation (\S \ref{sec:res_2}): \textbf{1)} Segment-level scores from \textsc{xCOMET-QE}, \textbf{2)} \textsc{MetricX-QE}, and \textbf{3)} BT-Score. As shown in Figure \ref{fig:kendalls_tau}, \askqe with \textsc{SentenceBERT} achieves the highest correlation with human judgments (0.265). This further suggests that \askqe aligns well with human evaluations of MT quality, motivating its use in a more human-centered application \---\ guiding decisions on whether to accept or reject MT output (\S \ref{sec:human_study}).

\subsection{Decision Making Simulation: How Actionable is \askqe Feedback?}
\label{sec:human_study}

\paragraph{Experiment Setup.}
We simulate a real-world scenario in which individuals decide whether to accept or reject an MT output $Y_{tgt}$ based on specific QE feedback, as illustrated in Figure \ref{fig:teasure_figure}. To formalize this binary decision-making process, we define a segment-level decision threshold, which depends on the type of QE feedback. When feedback consists of error annotations, such as \textsc{BioMQM} human judgments or MQM annotations from \textsc{xCOMET-QE}, we apply the following rule, where $e$ represents the highest error severity level:
\begin{equation}
    \small
    \text{Decision}(Y_{\text{tgt}}) =
    \begin{cases}
        \text{Accept}, & \text{if } e \in \{\textcolor{neutralyellow}{\text{\textbf{Neutral}}}, \textcolor{minororange}{\text{\textbf{Minor}}}\} \\
        \text{Reject}, & \text{otherwise}
    \end{cases}
\end{equation}

Conversely, when feedback is given as a single scalar score, such as outputs from \textsc{AskQE}, segment scores from \textsc{xCOMET-QE} (DA), \textsc{MetricX-QE}, or BT-Score, we fit a two-component Gaussian Mixture Model (GMM) for each QE metric. This model clusters $N$ segments into ``accept'' or ``reject'' groups based on score distribution.\footnote{Detailed description of the process is in Appendix \ref{appendix:gmm_detailed}.} We then compare the predicted segment-level decision labels $\hat{l}$ from each QE metric to the \textsc{BioMQM} human judgment labels $l$ and compute decision accuracy:
\begin{equation}
    \text{Decision Acc.(QE)} = \frac{\sum_{N}^{i=1}\mathbbm{1}(l_i = \hat{l}_i)}{N}
\end{equation}

Additionally, for \textsc{AskQE}, we introduce a simple baseline where decisions are based on the number of mismatches between reference and predicted answer pairs. Specifically, an MT output is rejected if the mismatch count exceeds a predefined threshold.

\begin{table}
\centering
\resizebox{\linewidth}{!}{%
    \begin{tabular}{llll}
    \specialrule{1.3pt}{0pt}{0pt}
    \textbf{QE Metric} & \textbf{Decision Acc. (\%)} \\
    \toprule

    \textbf{\textsc{AskQE}} (\textsc{F1}) & \textbf{75.75} \\ 
    \textbf{\textsc{AskQE}} (\textsc{SBERT}) & 63.77 \\ 
    % \textbf{\textsc{AskQE}} (\textsc{xCOMET-QE} (DA)) & 54.37 \\ 
    % \textbf{\textsc{AskQE}} (\textsc{xCOMET-QE} (MQM)) & 66.95 \\
    \textbf{\textsc{AskQE}} (Num. Mismatch=1) & 42.16 \\
    \textbf{\textsc{AskQE}} (Num. Mismatch=3) & 68.37 \\
    \textbf{\textsc{AskQE}} (Num. Mismatch=5) & 74.19 \\
    \textbf{\textsc{xCOMET-QE}} (DA) & 63.77 \\
    \textbf{\textsc{xCOMET-QE}} (MQM) & 61.00 \\
    \textbf{\textsc{MetricX-QE}*} & 59.57 \\
    \textbf{BT-Score} & 68.85 \\
    
    \specialrule{1.3pt}{0pt}{0pt}
    \end{tabular}
}
\caption{Decision accuracy (\%) of each QE metric in the \textsc{BioMQM} human simulation experiment. Best score in \textbf{bold}. \textbf{*}: We subtract the original \textsc{MetricX-QE} score from 25 to align is interpretation with other metrics.}
\label{tab:human_simulation}
\end{table}

\paragraph{Results.}
As shown in Table \ref{tab:human_simulation}, \askqe achieves decision accuracies comparable to or higher than other QE baselines. Qualitatively, \askqe using \textsc{F1} mostly disagrees with human ratings on translations with major linguistic convention errors, such as spelling or mistranslation errors, and disagrees with other QE metrics on minor linguistic convention errors. Compared to other QE metrics, \askqe is less effective at detecting spelling errors but more effective at identifying translations with stylistic issues (e.g., non-fluent) or mistranslations, leading to the highest decision accuracy (75.75\%). We show detailed qualitative analysis in Appendix \ref{appendix:qualitative_analysis}. Overall, these results highlight \textsc{AskQE}'s potential not only for MT quality assessment but also as actionable feedback to support real-world decision making in high-stakes contexts.

\section{Conclusion}

In this work, we introduce \textsc{AskQE}, a question generation and answering framework designed to identify critical translation errors and provide actionable feedback for users relying on MT outputs in a language they do not understand. Using the \textsc{ContraTICO} dataset (\S \ref{sec:perturbation}) across five language pairs, we validate the effectiveness of our method for critical error detection (\S \ref{sec:res_1}) and correlation with established QE metrics (\S \ref{sec:res_2}). Analysis of generated questions shows that most focus on extent or definition while remaining faithful to the source context (\S \ref{sec:question_categorization}). We further show that \askqe generalizes well to realistic scenarios with naturally occurring translation errors in the \textsc{BioMQM} dataset, achieving stronger correlations with human judgments (\S \ref{sec:biomqm_generalize}) and improving MT acceptance decisions compared to other QE metrics (\S \ref{sec:human_study}). 

These findings highlight the promise of QG/QA framework for MT quality assessment and QE feedback, calling for future work to explore optimal strategies such as enhancing context integration during QG and expanding to multi-turn QA.

% In future works, we envision integrating the QA pairs into interactive translation tools to help users assess and act on MT quality in real time.
\section{Limitations}

Our experiment setup is as comprehensive as our computational budget allows, while we could not cover every possible variant. The scope of our study is limited to out-of-English language pairs, as generating questions with LLMs has been more extensively studied in English \citep{li-zhang-2024-planning, fu-etal-2024-qgeval}, and using English as the source language benefits performance from its prevalence in LLM training data \citep{llama3}. This leaves open questions on how to design an optimal QG/QA framework by exploring various combinations of of LLMs, language pairs, and contextual inputs for QG, which we leave for future work.

% Added limitation - errors that are addition to the content of the source
As part of the construction process of the \textsc{ContraTICO} dataset, we introduce several type of perturbations that adds new information to the content of the source. As shown in Figure \ref{fig:res_detect_errors}, \textsc{AskQE} can detect critical perturbation that introduces a new meaning (Expansion (Impact)) but finds it difficult to detect errors that only modify the intensity of an adjective or an adverb (Intensifier) or add contextually implied details without introducing new meaning (Expansion (No Impact)).

% Our human study focuses on monolingual source speakers whom self-identified as proficient in English but not in the target language, reflecting a scenario where non-English natives encounter MT outputs of COVID-19 articles originally written in English. An important complementary study remains \---\ evaluating with the monolingual target speakers. This would require modifying our current setup to generate questions from the backtranslated source and derive answers from both the backtranslated source and the perturbed MT output. Future work can explore this variant through a human-centered study to assess the usefulness of \askqe for monolingual target speakers.
\section*{Acknowledgments}

We thank the anonymous reviewers and the members of the \textsc{clip} lab at University of Maryland for their constructive feedback. This work was supported in part by the Human Language Technology Center of Excellence at Johns Hopkins University, by NSF Fairness in AI Grant 2147292, and by the Institute for Trustworthy AI in Law and Society (TRAILS), which is supported by the National Science Foundation under Award No. 2229885. The views and conclusions contained herein are those of the authors and should not be interpreted as necessarily representing the official policies, either expressed or implied, of NSF or the U.S. Government. The U.S. Government is authorized to reproduce and distribute reprints for governmental purposes notwithstanding any copyright annotation therein.

\bibliography{custom}

% -------------- Appendix ---------------- %
\appendix

\section{Prompts}
\label{appendix:prompts}
\subsection{Perturbation}
We show prompt templates for each perturbation types for English-Spanish (\textsc{En-Es}) language pair. For other target languages, the 2-shot examples are replaced accordingly.

\begin{prompt}[title={Prompt A.1.1: Perturbation (Synonym)}]
\textbf{Task:} You will be given a \texttt{\{target lang\}} sentence. Your goal is to perturb the sentence by replacing one or two words (noun, verb, adjective or adverb) to its synonym. Please make sure it does not changes the meaning in a significant way. Output only the perturbed \texttt{\{target lang\}} sentence without giving any additional explanation. \\ \\
*** Example Starts *** \\
\textbf{Original:} Puede tratarse de un miembro de la secretaría o del personal clínico, según el protocolo de cada consultorio. \\
\textbf{Perturbed:} Puede tratarse de un miembro de la secretaría o del personal médico, según el protocolo de cada consultorio. \\ \\
\textbf{Original:} Además, reclutaremos nuevos consultorios de monitoreo. \\
\textbf{Perturbed:} Asimismo, contrataremos nuevos consultorios de seguimiento. \\
*** Example Ends *** \\ \\
\textbf{Original:} \texttt{\{original\}} \\
\textbf{Perturbed:}
\end{prompt}
\begin{prompt}[title={Prompt A.1.2: Perturbation (Word Order)}]
\textbf{Task:} You will be given a \texttt{\{target lang\}} sentence. Your goal is to perturb the sentence by changing the word order. Please make sure it does not changes the meaning in a significant way. Output only the perturbed \texttt{\{target lang\}} sentence without giving any additional explanation. \\ \\
*** Example Starts *** \\
\textbf{Original:} Puede tratarse de un miembro de la secretaría o del personal clínico, según el protocolo de cada consultorio. \\
\textbf{Perturbed:} Puede tratarse de un miembro del personal clínico o de la secretaría, según el protocolo de cada consultorio. \\ \\
\textbf{Original:} Desarrollaremos un observatorio para presentar los datos a nivel nacional, así como un tablero de control para hacer observaciones a los consultorios sobre la calidad de sus datos y la recolección de muestras virológicas y serológicas. \\
\textbf{Perturbed:} Se desarrollará un observatorio para presentar los datos a nivel nacional, junto con un tablero de control destinado a proporcionar observaciones a los consultorios sobre la calidad de los datos y la recolección de muestras virológicas y serológicas. \\
*** Example Ends *** \\ \\
\textbf{Original:} \texttt{\{original\}} \\
\textbf{Perturbed:}
\end{prompt}
\begin{prompt}[title={Prompt A.1.3: Perturbation (Spelling)}]
\textbf{Task:} You will be given a \texttt{\{target lang\}} sentence. Your goal is to perturb the sentence by misspelling one or two words. The misspelled words should be important in the sentence. Output only the perturbed \texttt{\{target lang\}} sentence without giving any additional explanation. \\ \\
*** Example Starts *** \\
\textbf{Original:} Puede tratarse de un miembro de la secretaría o del personal clínico, según el protocolo de cada consultorio. \\
\textbf{Perturbed:} Puede tratarse de un miembro de la secretaría o del persnal clínico, según el protcolo de cada consultorio. \\ \\
\textbf{Original:} Además, reclutaremos nuevos consultorios de monitoreo. \\
\textbf{Perturbed:} Además, reclutaremos nuevos cosultorios de monitoreo. \\
*** Example Ends *** \\ \\
\textbf{Original:} \texttt{\{original\}} \\
\textbf{Perturbed:}
\end{prompt}
\begin{prompt}[title={Prompt A.1.4: Perturbation (Intensifier)}]
\textbf{Task:} You will be given a \texttt{\{target lang\}} sentence. Your goal is to perturb the sentence by adding one or two words that changes the intensity of the existing word. Please make sure that the added word does not disturb the grammaticality of the sentence. Output only the perturbed \texttt{\{target lang\}} sentence without giving any additional explanation. \\ \\
*** Example Starts *** \\
\textbf{Original:} Los síntomas comunes incluyen fiebre, tos seca y fatiga. \\
\textbf{Perturbed:} Los síntomas comunes incluyen fiebre alta, tos seca y fatiga. \\ \\
\textbf{Original:} La dificultad para respirar, el dolor de garganta, los dolores de cabeza, las molestias corporales o la producción de esputo se encuentran entre los otros síntomas menos comunes. \\
\textbf{Perturbed:} La dificultad para respirar severa, el dolor de garganta, los dolores de cabeza, las molestias corporales o la producción de esputo se encuentran entre los otros síntomas menos comunes. \\
*** Example Ends *** \\ \\ 
\textbf{Original:} \texttt{\{original\}} \\
\textbf{Perturbed:}
\end{prompt}
\begin{prompt}[title={Prompt A.1.5: Perturbation (Expansion (No Impact))}]
\textbf{Task:} You will be given a \texttt{\{target lang\}} sentence. Your goal is to perturb the sentence by adding one or two words in the sentence. Please make sure that the added word does not disturb the grammaticality of the sentence and does not changes the meaning in a significant way. The added words would add more context that was already obvious from the sentence. Output only the perturbed \texttt{\{target lang\}} sentence without giving any additional explanation. \\ \\
*** Example Starts *** \\
\textbf{Original:} si crees que tus síntomas o problemas justifican un examen más detallado. \\
\textbf{Perturbed:} si crees que tus síntomas o problemas justifican un examen médico más detallado. \\ \\
\textbf{Original:} En caso de respuesta afirmativa a estas preguntas de filtro, se debe pedir al paciente que no acuda al consultorio y que siga el esquema de PHE en su lugar. \\
\textbf{Perturbed:} En caso de respuesta afirmativa a estas preguntas de filtro, se debe pedir al paciente adulto que no acuda al consultorio y que siga el esquema de PHE en su lugar. \\
*** Example Ends *** \\ \\ 
\textbf{Original:} \texttt{\{original\}} \\
\textbf{Perturbed:}
\end{prompt}

\begin{prompt}[title={Prompt A.1.6: Perturbation (Expansion (Impact))}]
\textbf{Task:} You will be given a \texttt{\{target lang\}} sentence. Your goal is to perturb the sentence by adding words in the sentence. Please make sure that the added word does not disturb the grammaticality of the sentence but should change the meaning in a significant way. Output only the perturbed \texttt{\{target lang\}} sentence without giving any additional explanation. \\ \\
*** Example Starts *** \\
\textbf{Original:} Los síntomas comunes incluyen fiebre, tos seca y fatiga. \\
\textbf{Perturbed:} Los síntomas comunes incluyen fiebre y dolores musculares, tos seca y fatiga. \\ \\
\textbf{Original:} La dificultad para respirar, el dolor de garganta, los dolores de cabeza, las molestias corporales o la producción de esputo se encuentran entre los otros síntomas menos comunes. \\
\textbf{Perturbed:} La dificultad para respirar, el dolor de garganta, los dolores de cabeza, las molestias corporales, la producción de esputo y trastornos digestivos se encuentran entre los otros síntomas menos comunes. \\
*** Example Ends *** \\ \\ 
\textbf{Original:} \texttt{\{original\}} \\
\textbf{Perturbed:}
\end{prompt}
\begin{prompt}[title={Prompt A.1.7: Perturbation (Omission)}]
\textbf{Task:} You will be given a \texttt{\{target lang\}} sentence. Your goal is to perturb the sentence by removing information from the sentence. Remove only one or two words from the sentence. Please make sure that the removed information does not disturb the grammaticality of the sentence but should change the meaning in a significant way. Output only the perturbed \texttt{\{target lang\}} sentence without giving any additional explanation. \\ \\
*** Example Starts *** \\
\textbf{Original:} Los síntomas comunes incluyen fiebre, tos seca y fatiga. \\
\textbf{Perturbed:} Los síntomas comunes incluyen fatiga y fatiga. \\ \\
\textbf{Original:} Se están realizando investigaciones sobre una vacuna o un tratamiento antiviral específico. \\
\textbf{Perturbed:} Se están realizando investigaciones sobre un tratamiento antiviral específico. \\
*** Example Ends *** \\ \\
\textbf{Original:} \texttt{\{original\}} \\
\textbf{Perturbed:}
\end{prompt}
\begin{prompt}[title={Prompt A.1.8: Perturbation (Alteration)}]
\textbf{Task:} You will be given a \texttt{\{target lang\}} sentence. Your goal is to perturb the sentence by changing the affirmative sentence into negation (vice versa) or changing one word (noun, verb, adjective or adverb) to its antonym. Please make sure that the perturbation does not disturb the grammaticality of the sentence but should change the meaning in a significant way. Output only the perturbed \texttt{\{target lang\}} sentence without giving any additional explanation. \\ \\
*** Example Starts *** \\
\textbf{Original:} No logró aliviar el dolor con medicamentos que la competencia prohíbe a los participantes. \\
\textbf{Perturbed:} No logró aliviar el placer con medicamentos que la competencia prohíbe a los participantes. \\ \\
\textbf{Original:} El mes pasado, un comité presidencial recomendó la renuncia del antiguo CEP como parte de medidas para llevar al país a nuevas elecciones. \\
\textbf{Perturbed:} El mes pasado, un comité presidencial no recomendó la renuncia del antiguo CEP como parte de medidas para llevar al país a nuevas elecciones. \\
*** Example Ends *** \\ \\
\textbf{Original:} \texttt{\{original\}} \\
\textbf{Perturbed:}
\end{prompt}

\subsection{Question Generation (QG)}
\begin{prompt}[title={Prompt A.2.1: Question Generation (Vanilla)}]
\textbf{Task:} You will be given an English sentence. Your goal is to generate a list of relevant questions based on the sentence. Output only the list of questions in Python list format without giving any additional explanation. \\ \\
*** Example Starts *** \\
\textbf{Sentence:} but if you have the cough \\
\textbf{Questions:} [``What condition is being referred to?''] \\ \\
\textbf{Sentence:} The number of accessory proteins and their function is unique depending on the specific coronavirus. \\
\textbf{Questions:} [``What is unique depending on the specific coronavirus?'', ``What is unique about the function of accessory proteins?''] \\
*** Example Ends *** \\ \\
\textbf{Sentence:} \texttt{\{sentence\}} \\
\textbf{Questions:}
\end{prompt}
\begin{prompt}[title={Prompt A.2.2: Question Generation (NLI)}]
\textbf{Task:} You will be given an English sentence and a list of atomic facts, which are short sentences conveying one piece of information. Your goal is to generate a list of relevant questions based on the sentence. Output the list of questions in Python list format without giving any additional explanation. \\ \\
*** Example Starts *** \\
\textbf{Sentence:} but if you have the cough \\
\textbf{Atomic facts:} [``You have the cough.''] \\
\textbf{Questions:} [``What condition is being referred to?''] \\ \\
\textbf{Sentence:} The number of accessory proteins and their function is unique depending on the specific coronavirus. \\
\textbf{Atomic facts:} [``The number of accessory proteins is unique depending on the specific coronavirus.'', ``The function of accessory proteins is unique depending on the specific coronavirus.''] \\
\textbf{Questions:} [``What is unique depending on the specific coronavirus?'', ``What is unique about the function of accessory proteins?''] \\
*** Example Ends *** \\ \\
\textbf{Sentence:} \texttt{\{sentence\}} \\
\textbf{Atomic facts:} \texttt{\{atomic facts\}} \\
\textbf{Questions:}
\end{prompt}
\begin{prompt}[title={Prompt A.2.3: Question Generation (SRL)}]
\textbf{Task:} You will be given an English sentence and a dictionary of semantic roles in the sentence. Your goal is to generate a list of relevant questions based on the sentence. Output the list of questions in Python list format without giving any additional explanation. \\ \\
*** Example Starts *** \\
\textbf{Sentence:} but if you have the cough \\
\textbf{Semantic roles:} \{`Verb': `have', `ARG0': `you', `ARG1': `the cough'\} \\
\textbf{Questions:} [``What condition is being referred to?''] \\ \\
\textbf{Sentence:} The number of accessory proteins and their function is unique depending on the specific coronavirus. \\
\textbf{Semantic roles:} \{`Verb': `is', `ARG1': `The number of accessory proteins and their function', `MNR': `unique', `TMP': `depending on the specific coronavirus'\} \\
\textbf{Questions:} [``What is unique depending on the specific coronavirus?'', ``What is unique about the function of accessory proteins?''] \\
*** Example Ends *** \\ \\
\textbf{Sentence:} \texttt{\{sentence\}} \\
\textbf{Semantic roles:} \texttt{\{semantic roles\}} \\
\textbf{Questions:}
\end{prompt}

\subsection{Question Answering (QA)}
\begin{prompt}[title={Prompt A.3: Question Answering}]
\textbf{Task:} You will be given an English sentence and a list of relevant questions. Your goal is to generate a list of answers to the questions based on the sentence. Output only the list of answers in Python list format without giving any additional explanation. \\ \\
*** Example Starts *** \\
\textbf{Sentence:} and does this pain move from your chest? \\
\textbf{Questions:} [``What moves from your chest?'', ``Where does the pain move from?''] \\
\textbf{Answers:} [``The pain'', ``Your chest''] \\ \\
\textbf{Sentence:} Diabetes mellitus (784, 10.9\%), chronic lung disease (656, 9.2\%), and cardiovascular disease (647, 9.0\%) were the most frequently reported conditions among all cases. \\
\textbf{Questions:} [``What were the most frequently reported conditions among all cases?'', ``What percentage of cases reported diabetes mellitus?'', ``What percentage of cases reported chronic lung disease?'', ``What percentage of cases reported cardiovascular disease?''] \\
\textbf{Answers:} [``Diabetes mellitus, chronic lung disease, and cardiovascular disease'', ``10.9\%'', ``9.2\%'', ``9.0\%''] \\
*** Example Ends *** \\ \\
\textbf{Sentence:} \texttt{\{sentence\}} \\
\textbf{Questions:} \texttt{\{questions\}} \\
\textbf{Answers:} 
\end{prompt}

\section{Translation Quality of \textsc{TICO-19}}
\label{appendix:tico_gt}
As shown in Table \ref{tab:tico_gt}, we measure Google Translate quality of source sentences in the \tico dataset by computing \textsc{xCOMET-QE} scores between the source and reference translation. On average, translation quality is lower for lower-resource languages (\textsc{En-Hi} and \textsc{En-Tl}) than for higher-resource languages (\textsc{En-Es}, \textsc{En-Fr}, and \textsc{En-Zh}).

\begin{table}
\centering
\resizebox{\linewidth}{!}{%
    \begin{tabular}{ll}
    \specialrule{1.3pt}{0pt}{0pt}
    \textbf{Model} & \textbf{HuggingFace Model Name} \\
    \toprule
    
    \textbf{\textsc{Gemma-2 9b}} & \texttt{google/gemma-2-9b-it} \\
    \textbf{\textsc{Gemma-2 27b}} & \texttt{google/gemma-2-27b-it} \\
    \textbf{\textsc{LLaMA-3 8b}} & \texttt{meta-llama/Llama-3.1-8B-Instruct} \\
    \textbf{\textsc{LLaMA-3 70b}} & \texttt{meta-llama/Llama-3.1-70B-Instruct} \\
    \textbf{\textsc{Yi-1.5 9b}} & \texttt{01-ai/Yi-1.5-9B-Chat} \\
    
    \specialrule{1.3pt}{0pt}{0pt}
    \end{tabular}
}
\caption{HuggingFace model names for all tested LLMs.} 
\label{tab:huggingface_api}
\end{table}
\begin{table}
\centering
\resizebox{140pt}{!}{%
    \begin{tabular}{llll}
    \specialrule{1.3pt}{0pt}{0pt}
    \textbf{Language Pair} & \textbf{\textsc{xCOMET-QE}} \\
    \toprule

    \textbf{\textsc{En-Es}} & 0.886 \\
    \textbf{\textsc{En-Fr}} & 0.855 \\
    \textbf{\textsc{En-Hi}} & 0.564 \\
    \textbf{\textsc{En-Tl}} & 0.704 \\
    \textbf{\textsc{En-Zh}} & 0.730 \\
    
    \specialrule{1.3pt}{0pt}{0pt}
    \end{tabular}
}
\caption{Google Translate quality for the \tico dataset. We measure the \textsc{xCOMET-QE} scores between the source sentence and the reference translation.}
\label{tab:tico_gt}
\end{table}

\section{Effectiveness of \askqe Design}
\subsection{Perturbation}
\label{appendix:perturbation_effective}
We evaluate perturbation effectiveness by measuring \textsc{xCOMET-QE} \citep{guerreiro-etal-2024-xcomet} and cosine similarity scores between the original source $X_{\mathrm{src}}$ and the perturbed translation $Y_{\mathrm{tgt}}$ in Table \ref{tab:effectiveness} (\textit{right}). Lower metric scores for more severe perturbations provide empirical support for the validity of our perturbation strategy.

\subsection{Backtranslation Quality}
\label{appendix:backtranslation_effective}
We evaluate the \textsc{xCOMET-QE} scores between the perturbed translation $Y_{\mathrm{tgt}}$ and its corresponding backtranslation $Y_{\mathrm{bt}}$, as shown in Table \ref{tab:effectiveness} (\textit{left}). Higher scores indicate better backtranslation quality. On average, \textsc{xCOMET-QE} scores are 0.757, 0.782, 0.671, 0.762, and 0.767 for \textsc{En-Es}, \textsc{En-Fr}, \textsc{En-Hi}, \textsc{En-Tl}, and \textsc{En-Zh}, respectively, resulting in an overall average of 0.748, which suggests a reasonable level of backtranslation quality.

\subsection{Comparison to Cross-lingual QA System}
\label{appendix:crosslingual_qa}
We compared the monolingual English QA system used in \textsc{AskQE} to a cross-lingual QA system that bypasses backtranslation and directly generates answers using the original source $X_{\mathrm{src}}$ and the perturbed translation $Y_{\mathrm{tgt}}$. Since the outputs are in different languages, we use a neural metric \textsc{SentenceBERT} instead of string-based comparison metrics. As shown in Table \ref{tab:crosslingual_qa}, our English QA system outperforms the cross-lingual QA system in differentiating between minor and critical errors, for all language pairs. These results further support the importance of incorporating backtranslation into our pipeline.

\subsection{Impact of Backtranslation on Critical Errors}
\label{appendix:impact_critical}
We hypothesize that with a high-quality MT system, it is unlikely to mask errors present in the original MT. To test this, we conduct an error analysis to examine whether any critical errors in the original MT go undetected because they are erased during backtranslation. For each language pair, we prompt \textsc{GPT-4} with the reference translation $Y_{\mathrm{ref}}$, the perturbed translation $Y_{\mathrm{tgt}}$, and its backtranslation $Y_{\mathrm{bt}}$, asking whether the specific error introduced in the perturbed translation remains identifiable in the backtranslation. As shown in Table \ref{tab:impact_critical}, the average percentage of backtranslations in which the critical error remains detectable is high: 87.76\% for Expansion (No Impact), 86.61\% for Omission, and 90.18\% for Alteration errors.

\begin{table}
\centering
\resizebox{\linewidth}{!}{%
    \begin{tabular}{llllllll}
    \specialrule{1.3pt}{0pt}{0pt}
    \textbf{Language pair} & \textbf{Error Type} & \textbf{Identified (\%)}\\
    \toprule

    \multirow{3}{*}{\textsc{\textbf{En-Es}}} & Expansion (Impact) & 90.42 \\
    & Omission & 86.86 \\
    & Alteration & 87.02 \\
    \midrule

    \multirow{3}{*}{\textsc{\textbf{En-Fr}}} & Expansion (Impact) & 88.88 \\
    & Omission & 89.94 \\
    & Alteration & 91.25 \\
    \midrule

    \multirow{3}{*}{\textsc{\textbf{En-Hi}}} & Expansion (Impact) & 86.32 \\
    & Omission & 87.78 \\
    & Alteration & 93.85 \\
    \midrule

    \multirow{3}{*}{\textsc{\textbf{En-Tl}}} & Expansion (Impact) & 87.65 \\
    & Omission & 82.78 \\
    & Alteration & 86.11 \\
    \midrule

    \multirow{3}{*}{\textsc{\textbf{En-Zh}}} & Expansion (Impact) & 85.55 \\
    & Omission & 85.68 \\
    & Alteration & 92.67 \\
    
    \specialrule{1.3pt}{0pt}{0pt}
    \end{tabular}
}
\caption{Error analysis on the impact of backtranslation on critical errors per language pair.}
\label{tab:impact_critical}
\end{table}

\section{Details of \textsc{AskQE}}

\subsection{Detailed Results}
\label{appendix:detailed_askqe}
From Tables \ref{tab:detailed_gemma_9b} to \ref{tab:detailed_yi_9b}, we present detailed results of \askqe using different metrics: \textsc{F1}, \textsc{EM}, \textsc{chrF}, \textsc{BLEU}, and \textsc{SentenceBERT} per language pair, severity level, and perturbation type.

\subsection{Variants of QG}
\label{appendix:qg_variants}
We decompose the standard QG process and extract relevant information from the source sentence to enhance the prompt with additional context. We introduce three variants: \textbf{1)} Vanilla, \textbf{2)} NLI, and \textbf{3)} SRL. When no supplementary information is provided beyond the source sentence, we refer to this approach as \textbf{Vanilla}. The \textbf{NLI}-based process is detailed in Section \ref{sec:qg}. Further, we incorporate semantic role labeling (\textbf{SRL}) to generate questions that target the most important parts of the source sentence. We prompt \textsc{GPT-4}o to annotate semantic roles in the source following the PropBank framework \citep{kingsbury-palmer-2002-treebank, palmer-etal-2005-proposition}, covering both the core and non-core roles \citep{bonial2010propbank}. We then prompt an LLM to generate questions given the source that focus on the extracted semantic roles. On average, 1.56 core roles are labeled per instance. Detailed distribution of non-core roles are outlined in Table \ref{tab:srl_stats}.

\begin{table}
\centering
\resizebox{150pt}{!}{%
    \begin{tabular}{llll}
    \specialrule{1.3pt}{0pt}{0pt}
    \textbf{Role} & \textbf{Description} & \textbf{Count} \\
    \toprule

    \textbf{TMP} & Temporal & 303 \\
    \textbf{LOC} & Locative & 227 \\
    \textbf{MNR} & Manner & 223 \\
    \textbf{PRP} & Purpose or Reason & 127 \\
    \textbf{MOD} & Modality & 116 \\
    \textbf{CAU} & Causal & 47 \\
    \textbf{COM} & Comitative & 29 \\
    \textbf{DIR} & Directional & 27 \\
    \textbf{GOL} & Goal & 14 \\
    \textbf{EXT} & Extent & 14 \\
    \textbf{NEG} & Negation & 7 \\
    \textbf{CON} & Conditional & 1 \\

    \specialrule{1.3pt}{0pt}{0pt}
    \end{tabular}
}
\caption{Dataset distribution of non-core roles in SRL.} 
\label{tab:srl_stats}
\end{table}

\subsection{Effect of Entailment Classification}
\label{appendix:entailment}
As detailed in \S \ref{sec:qg}, each instance in the \textsc{ContraTICO} dataset yields an average of 3.61 facts, of which 3.08 are retained after entailment filtering, using a \textsc{DeBERTa-v3 L} (340M) NLI classifier. This component is lightweight, requiring 01:13 (MM:SS) on a single NVIDIA RTX A5000 GPU to process 971 English sentences. To assess the impact of skipping this step, we conduct an ablation where all generated facts are used directly during QA without NLI filtering. As shown in Table \ref{tab:entailment}, removing the entailment classification step consistently lowers average F1 and EM scores and reduces the score gap between minor and critical errors. This demonstrates that entailment filtering enhances both overall QA accuracy and sensitivity to error severity, justifying its inclusion despite minimal computational cost.

\begin{table}[!htp]
\centering
\resizebox{\linewidth}{!}{%
    \begin{tabular}{llllllll}
    \specialrule{1.3pt}{0pt}{0pt}
    
    \textbf{Severity} & \textbf{Perturbation} & \textbf{F1 (w/o)} & \textbf{F1 (w/)} & \textbf{EM (w/o)} & \textbf{EM (w/)}\\
    \toprule

    \multirow{5}{*}{\hl{\textbf{Minor}}} & Spelling & 0.819 & 0.840 & 0.675 & 0.704 \\
    & Word Order & 0.811 & 0.820 & 0.634 & 0.679 \\
    & Synonym & 0.809 & 0.816 & 0.635 & 0.676 \\
    & Intensifier & 0.798 & 0.812 & 0.621 & 0.655 \\
    & Expansion (No Impact) & 0.790 & 0.814 & 0.622 & 0.665 \\
    \midrule
    
    \multirow{3}{*}{\hlred{\textbf{Critical}}} & Expansion (Impact) & 0.776 & 0.809 & 0.618 & 0.653 \\
    & Omission & 0.743 & 0.746 & 0.603 & 0.602 \\
    & Alteration & 0.468 & 0.475 & 0.329 & 0.360 \\

    \specialrule{1.3pt}{0pt}{0pt}
    \end{tabular}
}
\caption{Average F1 and EM scores with \textbf{(w/)} and without \textbf{(w/o)} entailment classification step during the NLI-based question generation process.}
\label{tab:entailment}
\end{table}

\section{Correlation Analysis}
\subsection{Detailed Results}
\label{appendix:correlation}
As a sanity check for \textsc{AskQE}, we conduct a correlation analysis with three established QE metrics: \textbf{1)} \textsc{xCOMET-QE} \citep{guerreiro-etal-2024-xcomet} and \textbf{2)} \textsc{MetricX-QE} \citep{juraska-etal-2024-metricx} between the source $X_{\mathrm{src}}$ and the perturbed MT output $Y_{\mathrm{tgt}}$, and \textbf{3)} BT-score \citep{agrawal-etal-2022-quality} between the source $X_{\mathrm{src}}$ and the backtranslated MT output $Y_{\mathrm{bt}}$. We use \textsc{BERTScore} \citep{bertscore} as the MT metric for \textbf{3)}. As shown in Tables \ref{tab:correlation_gemma_9b} to \ref{tab:correlation_yi_9b}, we observe that \askqe exhibit strong, statistically significant correlations with all three QE baselines, supporting the validity of our approach. Among the 15 LLM configurations, \textsc{LLaMA-3 70b} with NLI achieves the highest average correlation.

\subsection{QE Metric Scores}
\label{appendix:raw_qe}
We report the raw scores for each QE metrics (\textsc{xCOMET-QE}, \textsc{MetricX-QE}, BT-Score) in Table \ref{tab:raw_qe} for each language pair and perturbation. We observe that the QE metric scores decrease as the severity level increases, showing similar trends as \textsc{AskQE} scores.

\section{Desiderata Evaluation}
\label{appendix:desiderata}
We perform a fine-grained analysis of each LLM configuration to better understand the quality of the generated questions from \textsc{AskQE}. A detailed description of each desideratum is provided in Table \ref{tab:desiderata}, and results across all configurations are reported in Table \ref{tab:detailed_desiderata}. On average, we observe that 3.15 questions are generated per instance, with an answerability rate of 89.51\%, indicating that the questions are both faithful and answerable using the source sentence. Based on these findings, along with the metric scores of \textsc{AsKQE}, we select \textsc{LLaMA-3 70b} with NLI as our best-performing method, which shows the highest diversity and answerability scores.

\section{\textsc{Details of BioMQM}}
\subsection{Dataset Statistics}
\label{appendix:biomqm_stats}
We use \textsc{BioMQM} \citep{zouhar-etal-2024-fine} as our testbed, a biomedical domain MT dataset containing abstracts of scientific publication and medical texts with error annotations by professional translators. We detail dataset statistics of \textsc{BioMQM} per error category and subcategory in Table \ref{tab:biomqm_stats_1}, per language pair and error severity in Table \ref{tab:biomqm_stats_2}.

We preprocess the development split by filtering instances where: \textbf{1)} the source sentence contains no errors and \textbf{2)} the source language is English. This results in a total of 5,216 instances with an average instance length is 22.45 words across five language pairs: English-German (\textsc{en-de}), English-Spanish (\textsc{en-es}), English-French (\textsc{en-fr}), English-Russian (\textsc{en-ru}), and English-Chinese (\textsc{en-zh}).

\begin{table}
\centering
\resizebox{\linewidth}{!}{%
    \begin{tabular}{llll}
    \specialrule{1.3pt}{0pt}{0pt}
    \textbf{Category} & \textbf{Subcategory} & \textbf{Count} \\
    \toprule

    \multirow{4}{*}{{\textbf{Linguistic Conventions}}} & Grammar & 555 \\
    & Spelling & 1575 \\
    & Punctuation & 393 \\
    & Register & 11 \\
    \midrule
    
    \multirow{3}{*}{{\textbf{Accuracy}}} & Mistranslation & 1297 \\
    & Untranslated & 228 \\
    & Addition & 44 \\
    \midrule

    \multirow{2}{*}{{\textbf{Terminology}}} & Inconsistent & 134 \\
    & Wrong term & 771 \\
    \midrule

    \multirow{3}{*}{{\textbf{Locale Conventions}}} & Date Format & 9 \\
    & Number Format & 38 \\
    & Measurement Format & 76 \\
    \midrule

    \textbf{Style} & Non Fluent & 1447 \\
    \midrule
    \textbf{Unintelligible} & - & 88 \\
    \midrule
    \textbf{Other} & - & 71 \\
    \midrule
    
    \textbf{Total} & & \textbf{6737} \\
    \specialrule{1.3pt}{0pt}{0pt}
    \end{tabular}
}
\caption{Dataset statistics for \textsc{BioMQM} \citep{zouhar-etal-2024-fine} per error category and subcategories.}
\label{tab:biomqm_stats_1}
\end{table}
\definecolor{neutralyellow}{RGB}{252, 186, 3}
\definecolor{minororange}{RGB}{247, 142, 5}
\definecolor{majorred}{RGB}{214, 0, 0}
\definecolor{criticalbrown}{RGB}{107, 0, 12}

\begin{table}
\centering
\resizebox{160pt}{!}{%
    \begin{tabular}{llll}
    \specialrule{1.3pt}{0pt}{0pt}
    \textbf{Language pair} & \textbf{Error Severity} & \textbf{Count} \\
    \toprule

    \multirow{4}{*}{\textsc{\textbf{En-De}}}
    & \textcolor{neutralyellow}{\textbf{Neutral}} & 50 \\
    & \textcolor{minororange}{\textbf{Minor}} & 572 \\
    & \textcolor{majorred}{\textbf{Major}} & 1520 \\
    & \textcolor{criticalbrown}{\textbf{Critical}} & 13 \\
    \midrule

    \multirow{4}{*}{\textsc{\textbf{En-Es}}}
    & \textcolor{neutralyellow}{\textbf{Neutral}} & 169 \\
    & \textcolor{minororange}{\textbf{Minor}} & 1376 \\
    & \textcolor{majorred}{\textbf{Major}} & 74 \\
    & \textcolor{criticalbrown}{\textbf{Critical}} & 94 \\
    \midrule

    \multirow{4}{*}{\textsc{\textbf{En-Fr}}}
    & \textcolor{neutralyellow}{\textbf{Neutral}} & 14 \\
    & \textcolor{minororange}{\textbf{Minor}} & 597 \\
    & \textcolor{majorred}{\textbf{Major}} & 140 \\
    & \textcolor{criticalbrown}{\textbf{Critical}} & 11 \\
    \midrule

    \multirow{4}{*}{\textsc{\textbf{En-Ru}}}
    & \textcolor{neutralyellow}{\textbf{Neutral}} & 216 \\
    & \textcolor{minororange}{\textbf{Minor}} & 205 \\
    & \textcolor{majorred}{\textbf{Major}} & 74 \\
    & \textcolor{criticalbrown}{\textbf{Critical}} & 20 \\
    \midrule

    \multirow{4}{*}{\textsc{\textbf{En-Zh}}}
    & \textcolor{neutralyellow}{\textbf{Neutral}} & 61 \\
    & \textcolor{minororange}{\textbf{Minor}} & 1210 \\
    & \textcolor{majorred}{\textbf{Major}} & 318 \\
    & \textcolor{criticalbrown}{\textbf{Critical}} & 3 \\
    
    \midrule
    \textbf{Total} & & \textbf{6737} \\
    \specialrule{1.3pt}{0pt}{0pt}
    \end{tabular}
}
\caption{Dataset statistics for \textsc{BioMQM} \citep{zouhar-etal-2024-fine} per language pair and error severity level.}
\label{tab:biomqm_stats_2}
\end{table}
\definecolor{neutralyellow}{RGB}{252, 186, 3}
\definecolor{minororange}{RGB}{247, 142, 5}
\definecolor{majorred}{RGB}{214, 0, 0}
\definecolor{criticalbrown}{RGB}{107, 0, 12}

\begin{table}
\centering
\resizebox{\linewidth}{!}{%
    \begin{tabular}{llllllll}
    \specialrule{1.3pt}{0pt}{0pt}
    \textbf{Language} & \textbf{Severity} & \textbf{\textsc{F1}} & \textbf{\textsc{EM}} & \textbf{\textsc{chrF}} & \textbf{\textsc{BLEU}} & \textbf{\textsc{SBERT}} \\
    \toprule

    \multirow{3}{*}{{\textbf{\textsc{En-De}}}} & \textcolor{neutralyellow}{\textbf{Neutral}} & 0.791 & 0.500 & 85.30 & 69.96 & 0.900 \\
    & \textcolor{minororange}{\textbf{Minor}} & 0.785 & 0.529 & 82.92 & 68.26 & 0.854 \\
    & \textcolor{criticalbrown}{\textbf{C+M}} & 0.721 & 0.431 & 77.94 & 60.94 & 0.848 \\
    \midrule

    \multirow{3}{*}{{\textbf{\textsc{En-Es}}}} & \textcolor{neutralyellow}{\textbf{Neutral}} & 0.785 & 0.549 & 81.16 & 69.00 & 0.853 \\
    & \textcolor{minororange}{\textbf{Minor}} & 0.735 & 0.431 & 79.12 & 60.98 & 0.827 \\
    & \textcolor{criticalbrown}{\textbf{C+M}} & 0.728 & 0.418 & 77.33 & 61.01 & 0.825 \\
    \midrule

    \multirow{3}{*}{{\textbf{\textsc{En-Fr}}}} & \textcolor{neutralyellow}{\textbf{Neutral}} & 0.843 & 0.567 & 87.51 & 62.93 & 0.829 \\
    & \textcolor{minororange}{\textbf{Minor}} & 0.721 & 0.486 & 77.25 & 62.90 & 0.825 \\
    & \textcolor{criticalbrown}{\textbf{C+M}} & 0.695 & 0.389 & 74.99 & 57.93 & 0.605 \\
    \midrule

    \multirow{3}{*}{{\textbf{\textsc{En-Ru}}}} & \textcolor{neutralyellow}{\textbf{Neutral}} & 0.694 & 0.340 & 73.59 & 54.35 & 0.821 \\
    & \textcolor{minororange}{\textbf{Minor}} & 0.703 & 0.363 & 74.08 & 56.32 & 0.698 \\
    & \textcolor{criticalbrown}{\textbf{C+M}} & 0.648 & 0.238 & 67.92 & 47.37 & 0.603 \\
    \midrule

    \multirow{3}{*}{{\textbf{\textsc{En-Zh}}}} & \textcolor{neutralyellow}{\textbf{Neutral}} & 0.705 & 0.404 & 75.60 & 56.13 & 0.812 \\
    & \textcolor{minororange}{\textbf{Minor}} & 0.661 & 0.317 & 71.40 & 51.00 & 0.808 \\
    & \textcolor{criticalbrown}{\textbf{C+M}} & 0.618 & 0.296 & 67.07 & 48.15 & 0.790 \\
    
    \specialrule{1.3pt}{0pt}{0pt}
    \end{tabular}
}
\caption{Disaggregated results of \askqe evaluated on \textsc{BioMQM} dataset per language pair and error severity. Note that we combine \textcolor{majorred}{\textbf{Major}} and \textcolor{criticalbrown}{\textbf{Critical}} errors since the number of comparisons are very limited for \textcolor{criticalbrown}{\textbf{Critical}} (shown as \textcolor{criticalbrown}{\textbf{C+M}}).}
\label{tab:biomqm_disagg}
\end{table}

\definecolor{neutralyellow}{RGB}{252, 186, 3}
\definecolor{minororange}{RGB}{247, 142, 5}
\definecolor{majorred}{RGB}{214, 0, 0}
\definecolor{criticalbrown}{RGB}{107, 0, 12}

\subsection{Detailed Results}
\label{appendix:biomqm_detailed}
We present detailed results of \askqe evaluated on \textsc{BioMQM} by language pair and error severity in Table \ref{tab:biomqm_disagg}. To ensure fair comparisons, we match the number of instances per row to the minimum available count. Given the limited number of \textcolor{criticalbrown}{\textbf{Critical}} errors, we merge \textcolor{majorred}{\textbf{Major}} and \textcolor{criticalbrown}{\textbf{Critical}} errors (\textcolor{criticalbrown}{\textbf{C+M}}) for analysis. Our findings from \textsc{ContraTICO} generalize to \textsc{BioMQM}, confirming that \askqe detects errors across different severity levels.

\subsection{Human Simulation: GMM}
\label{appendix:gmm_detailed}
\begin{figure*}[htp!]
    \centering
    % First Row
    \subfigure[\textsc{AskQE} (\textsc{F1})]{\includegraphics[width=0.45\textwidth]{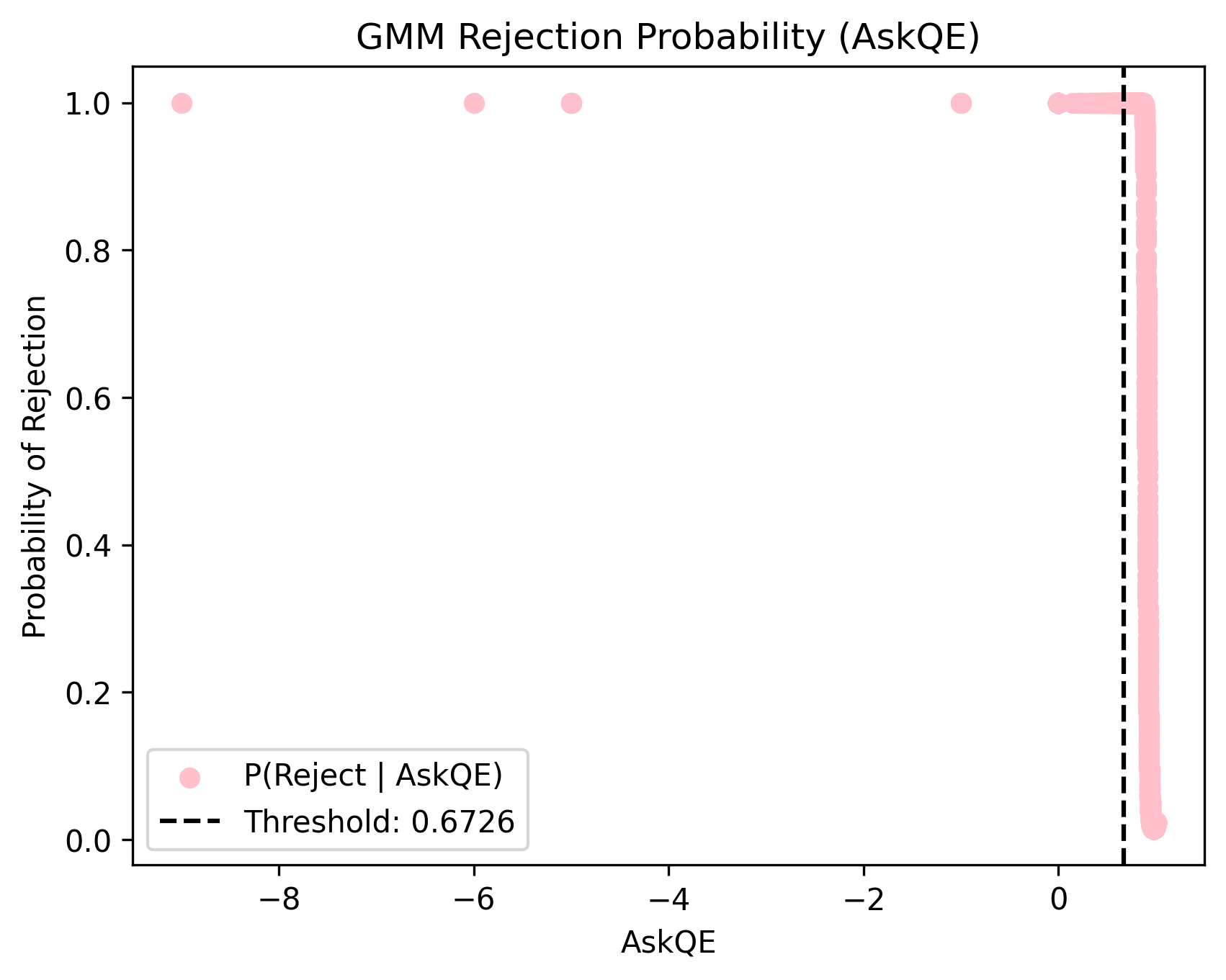}}
    \subfigure[\textsc{AskQE} (\textsc{SBERT})]{\includegraphics[width=0.45\textwidth]{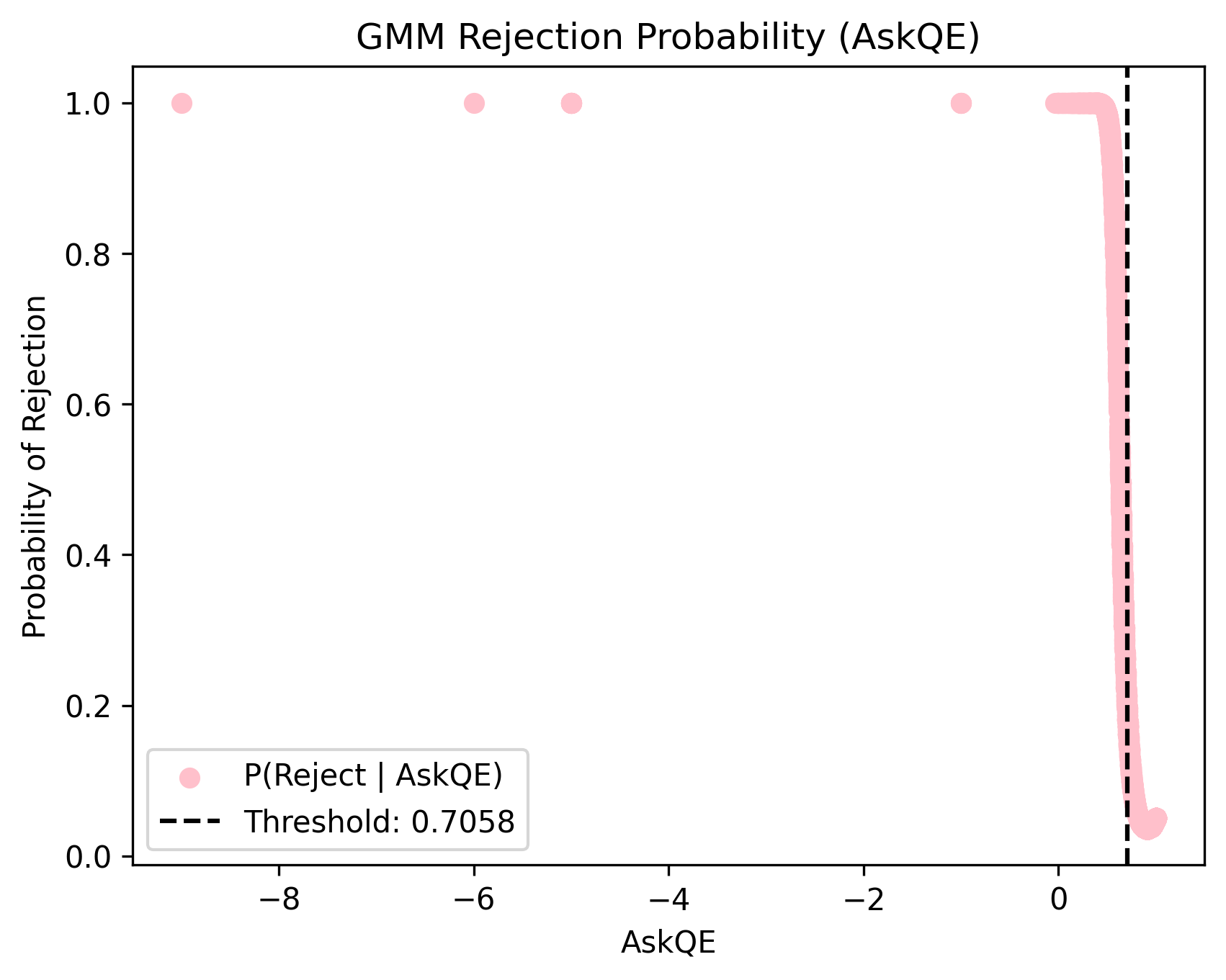}}

    % Second Row
    \subfigure[\textsc{xCOMET-QE} (DA)]{\includegraphics[width=0.45\textwidth]{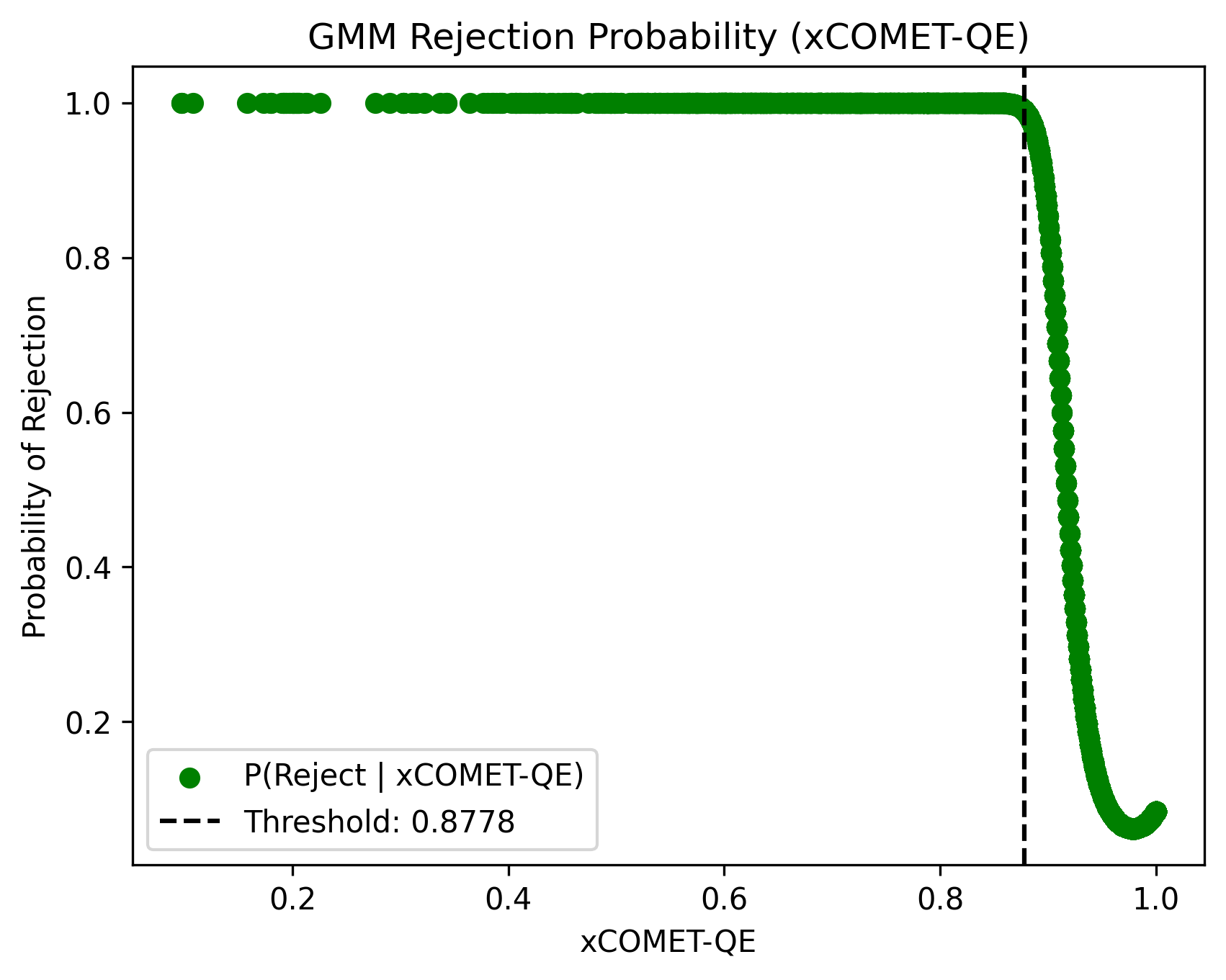}}
    \subfigure[\textsc{MetricX-QE}]{\includegraphics[width=0.45\textwidth]{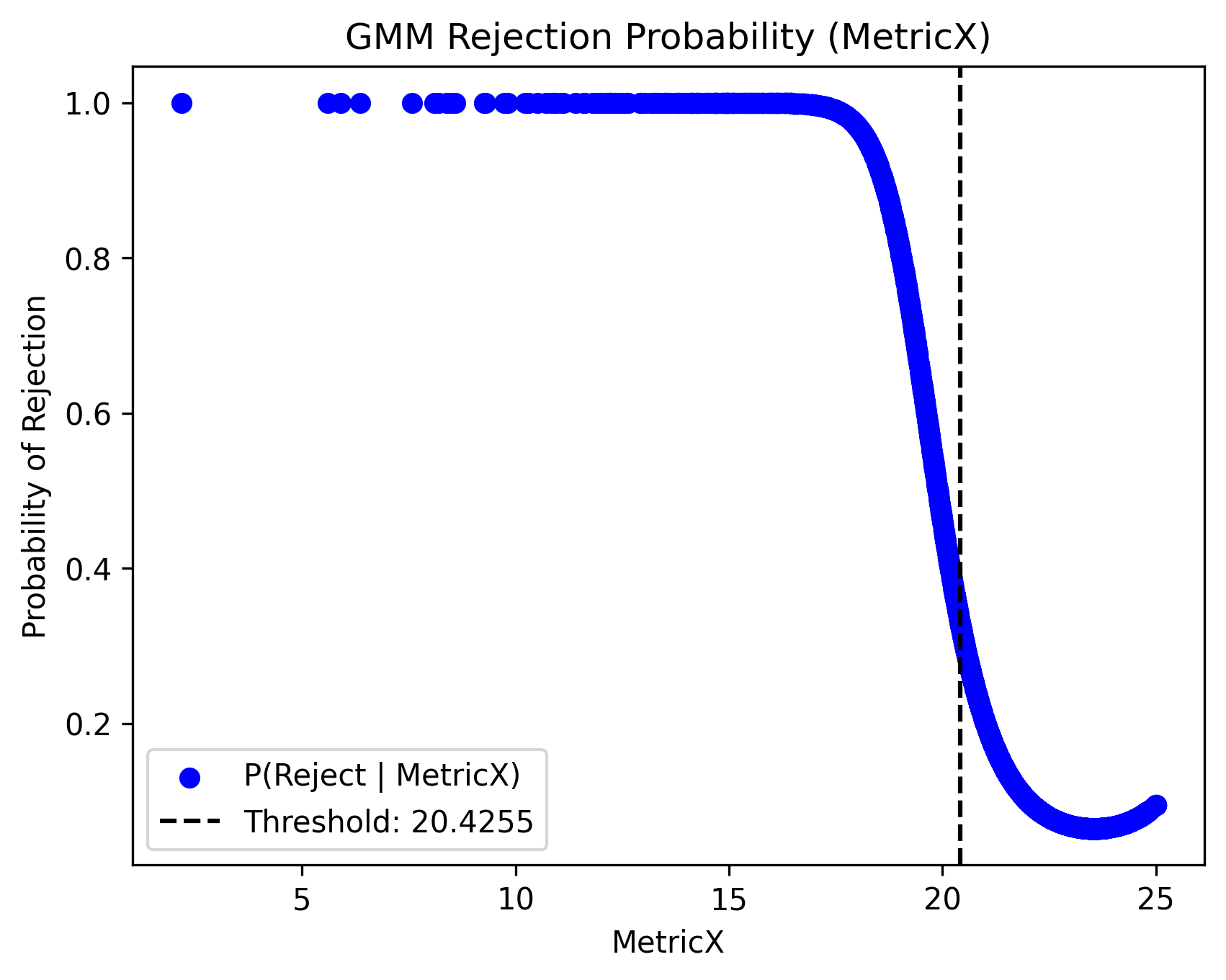}}

    % Third Row
    \subfigure[BT-Score]{\includegraphics[width=0.45\textwidth]{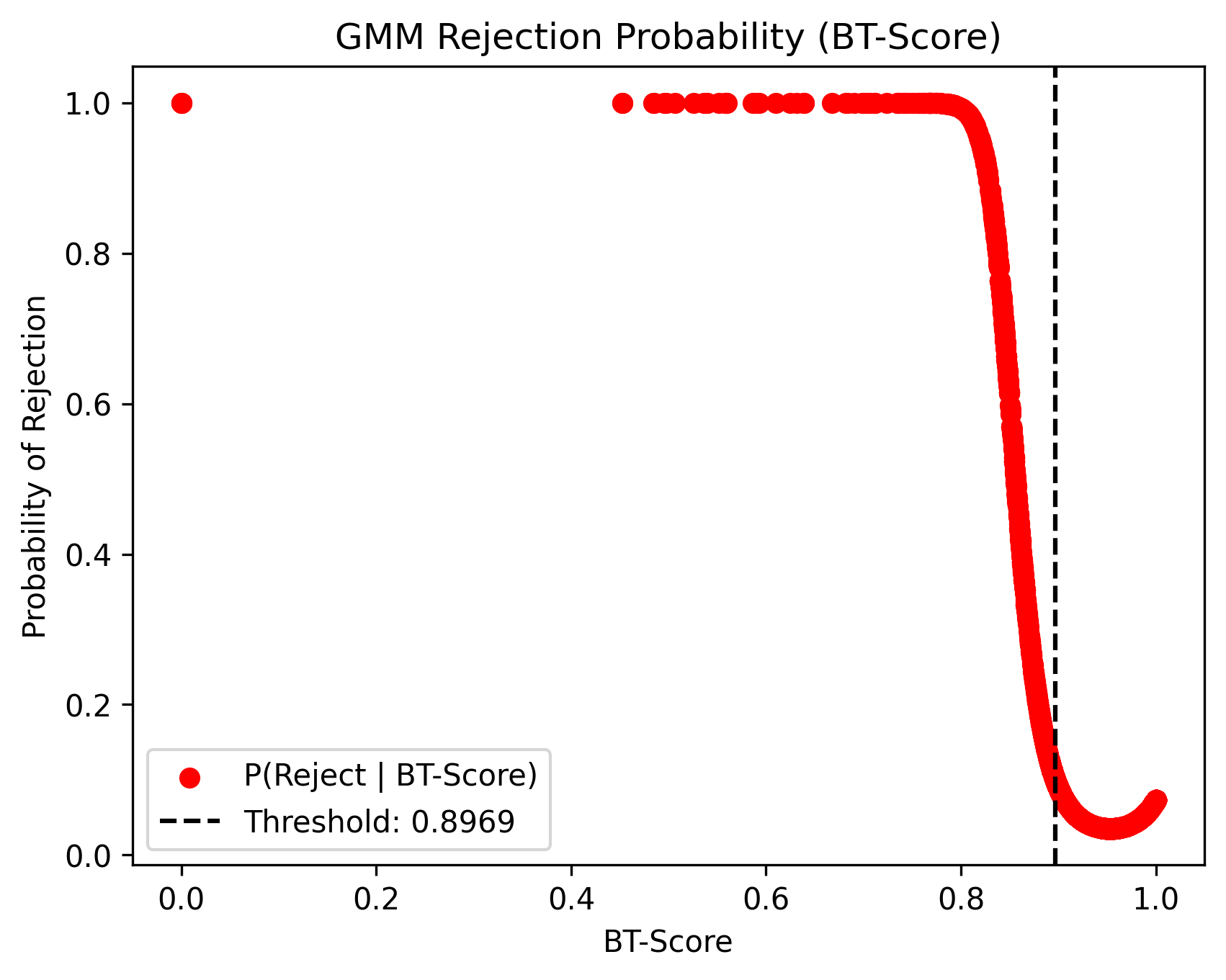}}

    \caption{Gaussian Mixture Model clustering for each QE metric. $x$-axis: QE metric scores; $y$-axis: Probability of rejection.}
    \label{fig:gmm}
\end{figure*}

For each QE metric baseline that outputs a single scalar score, we fit a two-component Gaussian Mixture Model (GMM).\footnote{\url{https://scikit-learn.org/stable/modules/mixture.html}} GMM is a probabilistic clustering model that assumes data points are generated from a mixture of Gaussian distributions, assigning a probability to each data point for belonging to a specific cluster. We hypothesize two clusters: \textbf{1)} high-quality translations with higher scores and \textbf{2)} low-quality translations with lower scores. Each segment in the \textsc{BioMQM} dataset is assigned a probability of belonging to the low-score cluster (rejection). Figure \ref{fig:gmm} illustrates the distribution of QE metric scores relative to their probability of rejection.

\subsection{Detailed Qualitative Analysis}
\label{appendix:qualitative_analysis}
In this section, we present the detailed qualitative findings from the human simulation study as briefly discussed in \S \ref{sec:human_study}. As shown in Tables \ref{tab:disagreement_severity} and \ref{tab:disagreement_category}, \textsc{AskQE (F1)} tends to disagree with human ratings primarily on major errors in terms of error severity, particularly spelling errors, followed by mistranslation errors in terms of error type. Further in Table \ref{tab:qual_comparison}, we compare \textsc{AskQE} to other QE metrics by highlighting the top-3 error types where \textsc{AskQE} is most effective and the bottom-3 where it is least effective. We find that AskQE is less effective at detecting spelling errors, but more effective at identifying issues related to style and mistranslation.

\definecolor{neutralyellow}{RGB}{252, 186, 3}
\definecolor{minororange}{RGB}{247, 142, 5}
\definecolor{majorred}{RGB}{214, 0, 0}
\definecolor{criticalbrown}{RGB}{107, 0, 12}

\begin{table}[!htp]
\centering
\resizebox{0.8\linewidth}{!}{%
    \begin{tabular}{llllllll}
    \specialrule{1.3pt}{0pt}{0pt}
    
    \textbf{Error Severity} & \textbf{Disagreement Count}\\
    \toprule

    \textcolor{neutralyellow}{\textbf{Neutral}} & 70 \\
    \textcolor{minororange}{\textbf{Minor}} & 751 \\
    \textcolor{majorred}{\textbf{Major}} & 2102 \\
    \textcolor{criticalbrown}{\textbf{Critical}} & 138 \\

    \specialrule{1.3pt}{0pt}{0pt}
    \end{tabular}
}
\caption{Disagreement count for human ratings and \textsc{AskQE (F1)} evaluated on \textsc{BioMQM} dataset per error severity.}
\label{tab:disagreement_severity}
\end{table}
\definecolor{neutralyellow}{RGB}{252, 186, 3}
\definecolor{minororange}{RGB}{247, 142, 5}
\definecolor{majorred}{RGB}{214, 0, 0}
\definecolor{criticalbrown}{RGB}{107, 0, 12}

\begin{table}[!htp]
\centering
\resizebox{\linewidth}{!}{%
    \begin{tabular}{llllllll}
    \specialrule{1.3pt}{0pt}{0pt}
    
    \textbf{Category} & \textbf{Subcategory} & \textbf{Disagreement Count}\\
    \toprule

    \multirow{4}{*}{{\textbf{Linguistic Conventions}}} & Grammar & 153 \\
    & Spelling & 1073 \\
    & Punctuation & 87 \\
    \midrule
    
    \multirow{3}{*}{{\textbf{Accuracy}}} & Mistranslation & 641 \\
    & Untranslated & 72 \\
    & Addition & 11 \\
    & Unspecified & 88 \\
    \midrule

    \multirow{2}{*}{{\textbf{Terminology}}} & Inconsistent & 44 \\
    & Wrong term & 344 \\
    \midrule

    \multirow{3}{*}{{\textbf{Locale Conventions}}} & Date Format & 4 \\
    & Number Format & 20 \\
    & Measurement Format & 14 \\
    \midrule

    \textbf{Style} & Non Fluent & 490 \\
    \midrule
    \textbf{Other} & - & 20 \\

    \specialrule{1.3pt}{0pt}{0pt}
    \end{tabular}
}
\caption{Disagreement count for human ratings and \textsc{AskQE (F1)} evaluated on \textsc{BioMQM} dataset per error category and subcategories.}
\label{tab:disagreement_category}
\end{table}
\begin{table*}[!htp]
\centering
\resizebox{\linewidth}{!}{%
    \begin{tabular}{llllllll}
    \specialrule{1.3pt}{0pt}{0pt}
    
    \textbf{QE Metric} & \textbf{Top-3} & \textbf{Bottom-3}\\
    \toprule

    \multirow{3}{*}{\textbf{\textsc{xCOMET-QE (DA)}}} & (1) Accuracy / Mistranslation / 401 & (1) Linguistic conventions / Spelling / 344 \\
    & (2) Style / Non-fluent / 248 & (2) Accuracy / Mistranslation / 275 \\
    & (3) Linguistic conventions / Spelling / 237 & (3) Style / Non-fluent / 87 \\
    \midrule

    \multirow{3}{*}{\textbf{\textsc{xCOMET-QE (MQM)}}} & (1) Accuracy / Mistranslation / 421 & (1) Linguistic conventions / Spelling / 249 \\
    & (2) Style / Non-fluent / 194 & (2) Accuracy / Mistranslation / 81 \\
    & (3) Linguistic conventions / Spelling / 156 & (3) Linguistic conventions / Punctuation / 73 \\
    \midrule

    \multirow{3}{*}{\textbf{\textsc{MetricX-QE}}} & (1) Style / Non-fluent / 537 & (1) Linguistic conventions / Spelling / 373 \\
    & (2) Accuracy / Mistranslation / 419 & (2) Accuracy / Mistranslation / 245 \\
    & (3) Linguistic conventions / Spelling / 301 & (3) Style / Non-fluent / 226 \\
    \midrule

    \multirow{3}{*}{\textbf{BT-Score}} & (1) Accuracy / Mistranslation / 197 & (1) Linguistic conventions / Spelling / 218 \\
    & (2) Style / Non-fluent / 128 & (2) Accuracy / Mistranslation / 172 \\
    & (3) Linguistic conventions / Spelling / 82 & (3) Style / Non-fluent / 72 \\

    \specialrule{1.3pt}{0pt}{0pt}
    \end{tabular}
}
\caption{Top-3 and Bottom-3 error types between \textsc{AskQE} and other QE metrics. \textbf{Top-3:} \textsc{AskQE} is most effective; \textbf{Bottom-3:} \textsc{AskQE} is least effective. Each cell presents in the format as error category/subcategory/count.}
\label{tab:qual_comparison}
\end{table*}

\section{Computational \& Time Efficiency}
\label{appendix:efficiency}
We compare the average computational, time, and cost efficiency for computing \textsc{AskQE} and baseline QE methods, as shown in Table \ref{tab:efficiency}, on 971 \textsc{En-Es} sentence pairs from \textsc{ContraTICO}. Notably, \textsc{AskQE} using \textsc{LLaMA-3 70b} achieves faster runtime than \textsc{xCOMET-QE}, and fact extraction step costs less than \$1 using \textsc{GPT-4}o due to the short average sentence length (24.79 words). We also propose two lower-cost variants of \textsc{AskQE}: \textbf{1)} Using smaller models for question answering (\S \ref{appendix:detailed_askqe}) and \textbf{2)} Substituting \textsc{GPT-4}o with lighter variants for question generation (\S \ref{appendix:qg_variants}). Having established this, future works could focus on optimizing and deploying such a tool, particularly in user-facing applications. \textsc{AskQE} could be deployed selectively, for instance using a standard segment-level QE score as a first pass and apply \textsc{AskQE}’s fine-grained evaluation only to translations that fall below a certain quality threshold.

\begin{table*}
\centering
\resizebox{\linewidth}{!}{%
    \begin{tabular}{llllllll}
    \specialrule{1.3pt}{0pt}{0pt}
    \textbf{Method} & \textbf{Model (size)} & \textbf{Computation (GPU)} & \textbf{Avg. Time (\textit{hh:mm})} & \textbf{Avg. Cost (\$)} \\
    \toprule

    \textbf{\textsc{xCOMET-QE}} & \textsc{xCOMET-XL (3b)} & 1 $\times$ NVIDIA RTX A5000 & 02:06 & 0 \\
    \textbf{\textsc{MetricX-QE}} & \textsc{MetricX-24 XL (3b)} & 1 $\times$ NVIDIA RTX A5000 & 00:02 & 0 \\
    \textbf{BT-Score} & \textsc{DeBERTa-v3 L (340M)} & 1 $\times$ NVIDIA RTX A5000 & 00:02 & 0 \\
    \textbf{\textsc{AskQE}} & \textsc{GPT-4}o for fact extraction & - & 00:20 & 0.979 \\
    \textbf{\textsc{AskQE}} & \textsc{LLaMA-3 7b} & 8 $\times$ NVIDIA RTX A5000 & 01:16 & 0 \\
    
    \specialrule{1.3pt}{0pt}{0pt}
    \end{tabular}
}
\caption{Average computational, time, and cost efficiency for tested methods. \textsc{GPT-4}o pricing is based on current OpenAI API rates at \url{https://openai.com/api/pricing/}.} 
\label{tab:efficiency}
\end{table*}

\begin{table*}[!htp]
\centering
\resizebox{\linewidth}{!}{%
    \begin{tabular}{l l p{16cm}}
    \specialrule{1.3pt}{0pt}{0pt}
    \textbf{Severity} & \textbf{Perturbation} & \textbf{Example} \\
    \toprule

    \multirow{14}{*}{\hl{\textbf{Minor}}} & \multirow{4}{*}{Spelling} & \textbf{Original:} Assurez-vous que toutes les informations et tous les conseils que vous obtenez ont été confirmés par des médecins et scientifiques de bonne réputation. \\
    & & \textbf{Perturbed:} Assurez-vous que toutes les informations et tous les conseils que vous obtenez ont été confirmés par des \textit{médins} et \textit{scientfiques} de bonne réputation. \\
    \cmidrule{2-3}

    & \multirow{2}{*}{Word Order} & \textbf{Original:} Dans la mesure du possible, évitez les endroits très fréquentés. \\
    & & \textbf{Perturbed:} Dans la mesure du possible, \textit{les endroits très fréquentés doivent être évités.} \\
    \cmidrule{2-3}

    & \multirow{2}{*}{Synonym} & \textbf{Original:} mais j’ai le rhume des foins aussi \\
    & & \textbf{Perturbed}: mais j’ai le rhume des \textit{prés} aussi \\
    \cmidrule{2-3}

    & \multirow{4}{*}{Intensifier} & \textbf{Original:} Le 11-mars-2020, l’Organisation mondiale de la santé déclarait la maladie à coronavirus-2019 (COVID-19) comme étant une pandémie. \\
    & & \textbf{Perturbed:} Le 11 mars 2020, l’Organisation mondiale de la santé déclarait la \textit{grave} maladie à coronavirus 2019 (COVID-19) comme étant une pandémie. \\
    \cmidrule{2-3}

    & \multirow{2}{*}{Expansion (No Impact)} & \textbf{Original:} si vous pensez que vos symptômes ou problèmes justifient un examen plus approfondi. \\
    & & \textbf{Perturbed:} si vous pensez que vos symptômes ou problèmes justifient un examen \textit{médical} plus approfondi. \\
    \midrule

    \multirow{9}{*}{\hlred{\textbf{Critical}}} & \multirow{3}{*}{Expansion (Impact)} & \textbf{Original:} Les symptômes courants comprennent la fièvre, une toux sèche et la fatigue. \\
    & & \textbf{Perturbed}: Les symptômes courants comprennent la fièvre et des douleurs musculaires, \textit{une toux sèche et la fatigue}. \\
    \cmidrule{2-3}

    & \multirow{2}{*}{Omission} & \textbf{Original:} Des recherches sur \textit{un vaccin ou} un traitement antiviral spécifique sont en cours. \\
    & & \textbf{Perturbed}: Des recherches sur un traitement antiviral spécifique sont en cours. \\
    \cmidrule{2-3}

    & \multirow{4}{*}{Alteration} & \textbf{Original:} Il n'a pas réussi à soulager la douleur avec des médicaments que la compétition interdit aux concurrents de prendre. \\
    & & \textbf{Perturbed}: Il n'a pas réussi à soulager le \textit{plaisir} avec des médicaments, que la compétition interdit aux concurrents de prendre. \\

    \specialrule{1.3pt}{0pt}{0pt}
    \end{tabular}
}
\caption{Detailed examples for each perturbation type, divided by severity (\hl{Minor}, \hlred{Critical}) for French (\textsc{Fr}) as target language. Perturbed parts of the reference translation (\textbf{Original}) are in \textit{italic} in \textbf{Perturbed}.}
\label{tab:perturbation_type}
\end{table*}

\clearpage

\begin{table*}[!htp]
\centering
\resizebox{\linewidth}{!}{%
    \begin{tabular}{c l l l l l}
    \specialrule{1.3pt}{0pt}{0pt}
    \textbf{Language Pair} & \textbf{Severity} & \textbf{Perturbation} & \textbf{\textsc{xCOMET($X_{\mathrm{src}}$, $Y_{\mathrm{tgt}}$)}} & \textbf{\textsc{Sim($X_{\mathrm{src}}$, $Y_{\mathrm{tgt}}$)}} & \textbf{\textsc{xCOMET($Y_{\mathrm{tgt}}$, $Y_{\mathrm{bt}}$)}} \\
    \toprule

    \multirow{9}{*}{\textbf{\textsc{En-Es}}} & Original & - & 0.762 & 0.908 & - \\
    \cmidrule{2-6}
    & \multirow{5}{*}{\hl{Minor}} & Spelling & 0.758 & 0.900 & 0.766 \\
    & & Word Order & 0.738 & 0.899 & 0.756 \\
    & & Synonym & 0.725 & 0.897 & 0.749 \\
    & & Intensifier & 0.740 & 0.886 & 0.760 \\
    & & Expansion (No Impact) & 0.748 & 0.886 & 0.760 \\
    \cmidrule{2-6}
    & \multirow{3}{*}{\hlred{Critical}} & Expansion (Impact) & 0.708 & 0.868 & 0.766 \\
    & & Omission & 0.619 & 0.877 & 0.764 \\
    & & Alteration & 0.473 & 0.863 & 0.728 \\
    \midrule

    \multirow{9}{*}{\textbf{\textsc{En-Fr}}} & Original & - & 0.786 & 0.897 & - \\
    \cmidrule{2-6}
    & \multirow{5}{*}{\hl{Minor}} & Spelling & 0.784 & 0.891 & 0.789 \\
    & & Word Order & 0.766 & 0.889 & 0.784 \\
    & & Synonym & 0.751 & 0.886 & 0.775 \\
    & & Intensifier & 0.765 & 0.880 & 0.787 \\
    & & Expansion (No Impact) & 0.774 & 0.878 & 0.791 \\
    \cmidrule{2-6}
    & \multirow{3}{*}{\hlred{Critical}} & Expansion (Impact) & 0.736 & 0.858 & 0.790 \\
    & & Omission & 0.635 & 0.865 & 0.784 \\
    & & Alteration & 0.547 & 0.842 & 0.756 \\
    \midrule

    \multirow{9}{*}{\textbf{\textsc{En-Hi}}} & Original & - & 0.632 & 0.870 & - \\
    \cmidrule{2-6}
    & \multirow{5}{*}{\hl{Minor}} & Spelling & 0.630 & 0.858 & 0.677 \\
    & & Word Order & 0.609 & 0.849 & 0.669 \\
    & & Synonym & 0.619 & 0.864 & 0.671 \\
    & & Intensifier & 0.624 & 0.852 & 0.675 \\
    & & Expansion (No Impact) & 0.617 & 0.849 & 0.674 \\
    \cmidrule{2-6}
    & \multirow{3}{*}{\hlred{Critical}} & Expansion (Impact) & 0.574 & 0.826 & 0.672 \\
    & & Omission & 0.348 & 0.821 & 0.686 \\
    & & Alteration & 0.459 & 0.835 & 0.642 \\
    \midrule

    \multirow{9}{*}{\textbf{\textsc{En-Tl}}} & Original & - & 0.725 & 0.898 & - \\
    \cmidrule{2-6}
    & \multirow{5}{*}{\hl{Minor}} & Spelling & 0.739 & 0.887 & 0.770 \\
    & & Word Order & 0.722 & 0.878 & 0.764 \\
    & & Synonym & 0.714 & 0.887 & 0.755 \\
    & & Intensifier & 0.726 & 0.873 & 0.768 \\
    & & Expansion (No Impact) & 0.716 & 0.882 & 0.762 \\
    \cmidrule{2-6}
    & \multirow{3}{*}{\hlred{Critical}} & Expansion (Impact) & 0.693 & 0.851 & 0.771 \\
    & & Omission & 0.609 & 0.848 & 0.766 \\
    & & Alteration & 0.581 & 0.858 & 0.743 \\
    \midrule

    \multirow{9}{*}{\textbf{\textsc{En-Zh}}} & Original & - & 0.767 & 0.868 & - \\
    \cmidrule{2-6}
    & \multirow{5}{*}{\hl{Minor}} & Spelling & 0.764 & 0.861 & 0.774 \\
    & & Word Order & 0.726 & 0.854 & 0.763 \\
    & & Synonym & 0.724 & 0.851 & 0.756 \\
    & & Intensifier & 0.763 & 0.857 & 0.776 \\
    & & Expansion (No Impact) & 0.749 & 0.853 & 0.772 \\
    \cmidrule{2-6}
    & \multirow{3}{*}{\hlred{Critical}} & Expansion (Impact) & 0.718 & 0.840 & 0.778 \\
    & & Omission & 0.613 & 0.815 & 0.772 \\
    & & Alteration & 0.583 & 0.829 & 0.741 \\    

    \specialrule{1.3pt}{0pt}{0pt}
    \end{tabular}
}
\caption{Perturbation and backtranslation quality measured per language pair. \textbf{\textsc{xCOMET($X_{\mathrm{src}}$, $Y_{\mathrm{tgt}}$)}:} \textsc{xCOMET-QE} scores between the original source and the perturbed translation (↓ as more severe perturbation). \textbf{\textsc{Sim($X_{\mathrm{src}}$, $Y_{\mathrm{tgt}}$)}:} Cosine similarity scores between the original source and the perturbed translation (↓ as more severe perturbation). \textbf{\textsc{xCOMET($Y_{\mathrm{tgt}}$, $Y_{\mathrm{bt}}$)}:} \textsc{xCOMET-QE} scores between the perturbed translation and the respective backtranslation (↑). The first two metrics evaluate the perturbation effectiveness, while the last one assesses backtranslation quality. For Original, we replace $Y_{\mathrm{tgt}}$ to $Y_{\mathrm{ref}}$.}
\label{tab:effectiveness}
\end{table*}

\clearpage

\begin{figure*}[htp!]
    \centering
    % First Row
    \subfigure[\textsc{En-Es}]{\includegraphics[width=0.45\textwidth]{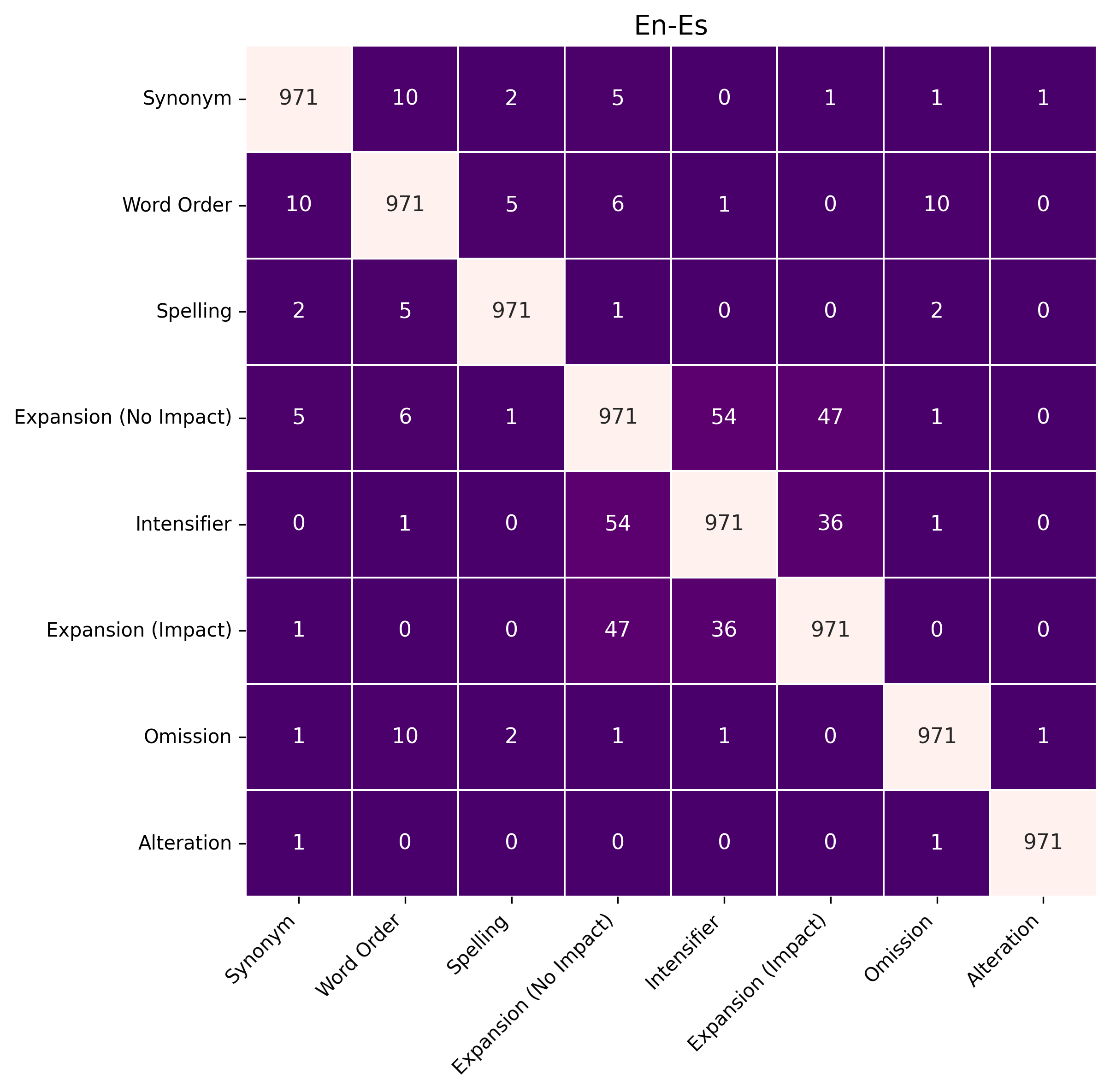}}
    \subfigure[\textsc{En-Fr}]{\includegraphics[width=0.45\textwidth]{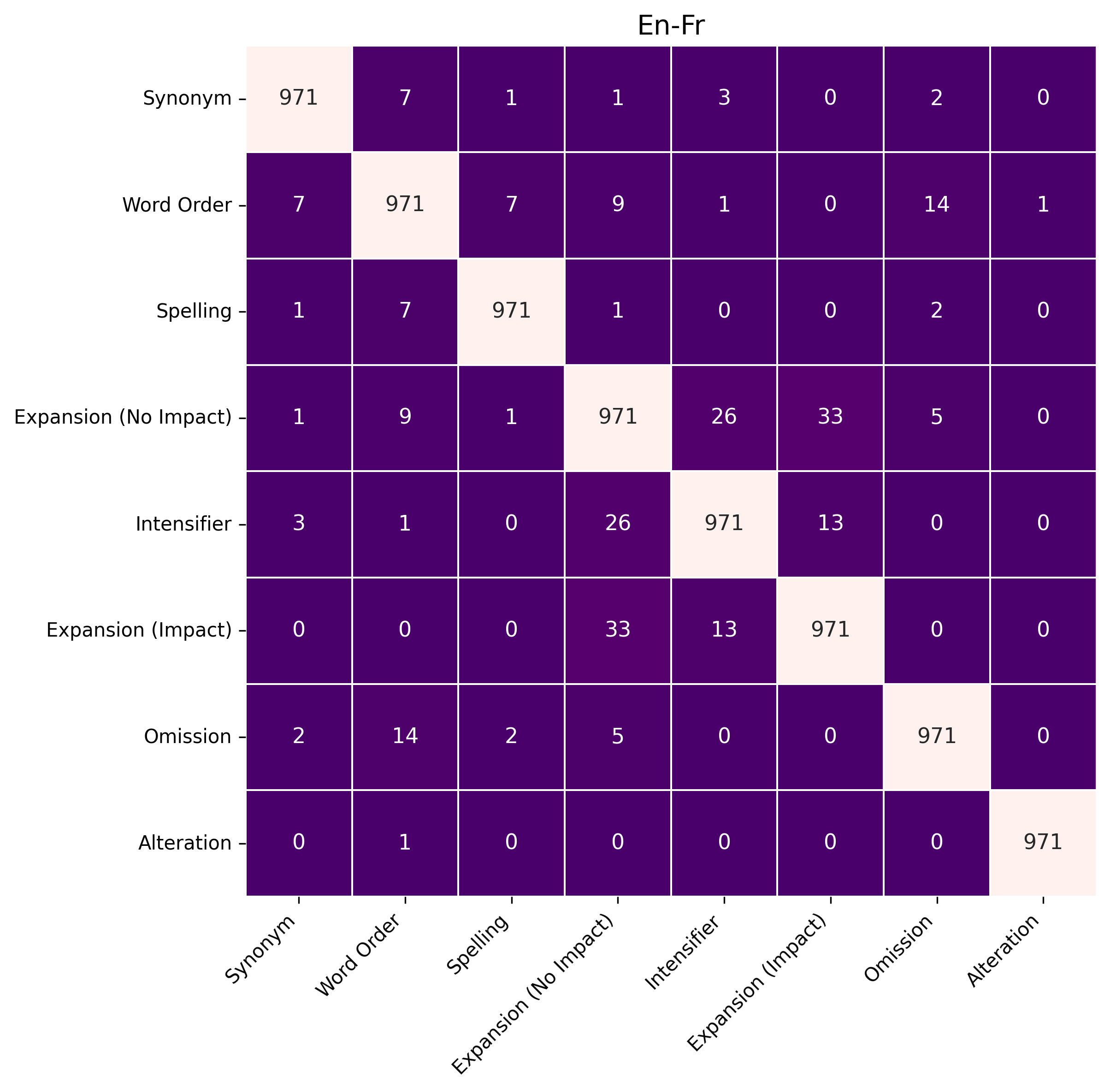}}

    % Second Row
    \subfigure[\textsc{En-Hi}]{\includegraphics[width=0.45\textwidth]{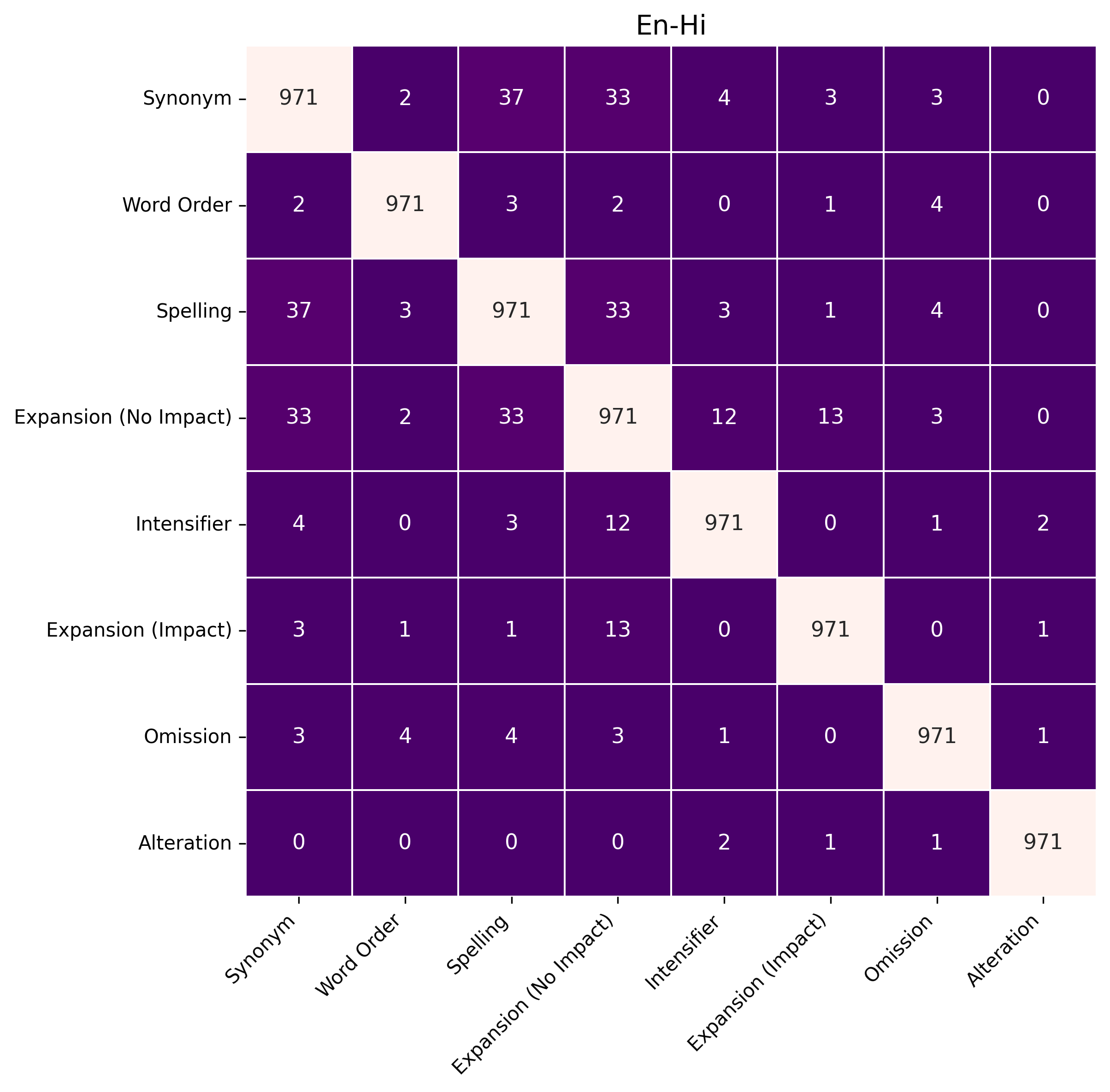}}
    \subfigure[\textsc{En-Tl}]{\includegraphics[width=0.45\textwidth]{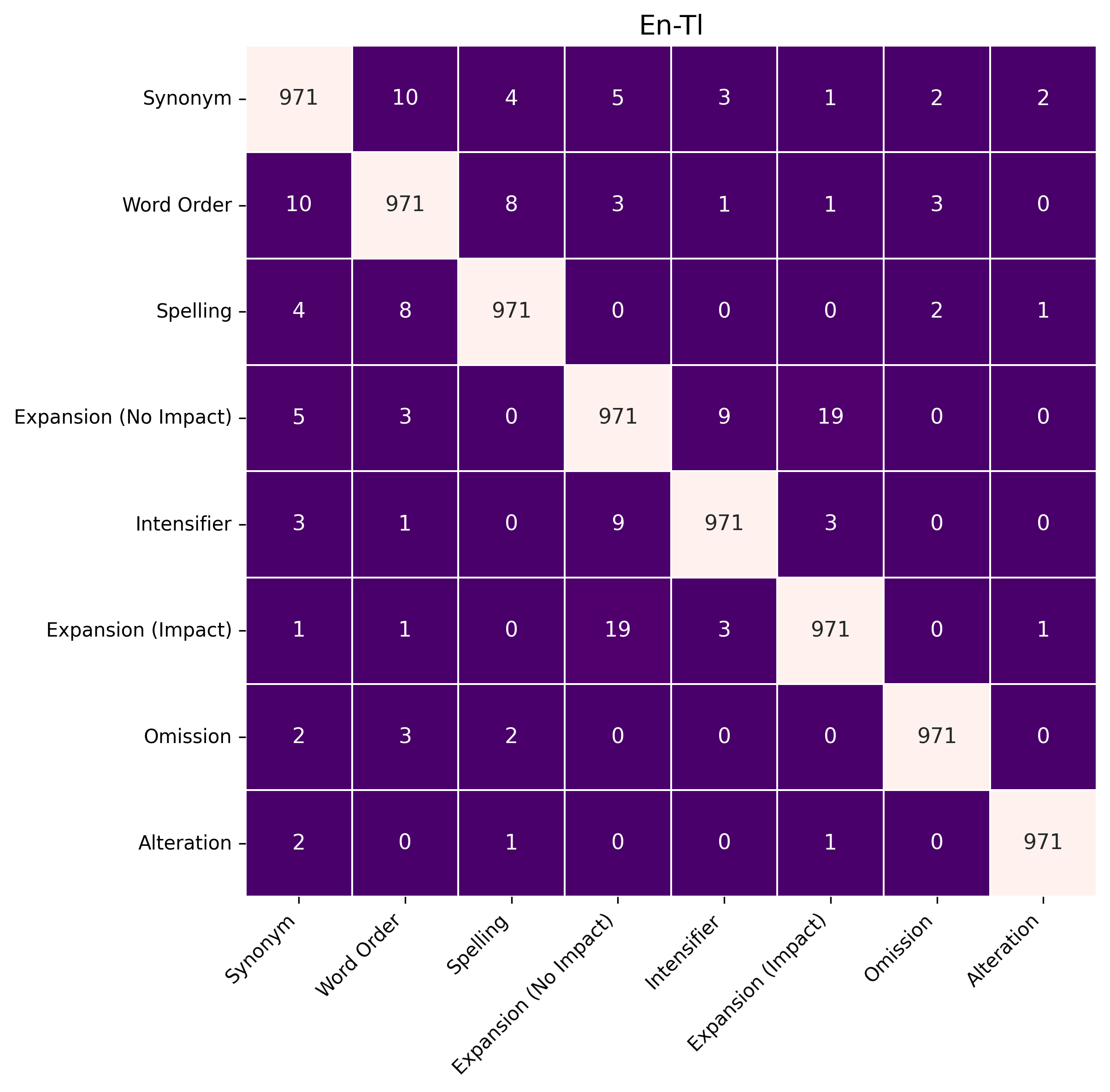}}

    % Third Row
    \subfigure[\textsc{En-Zh}]{\includegraphics[width=0.45\textwidth]{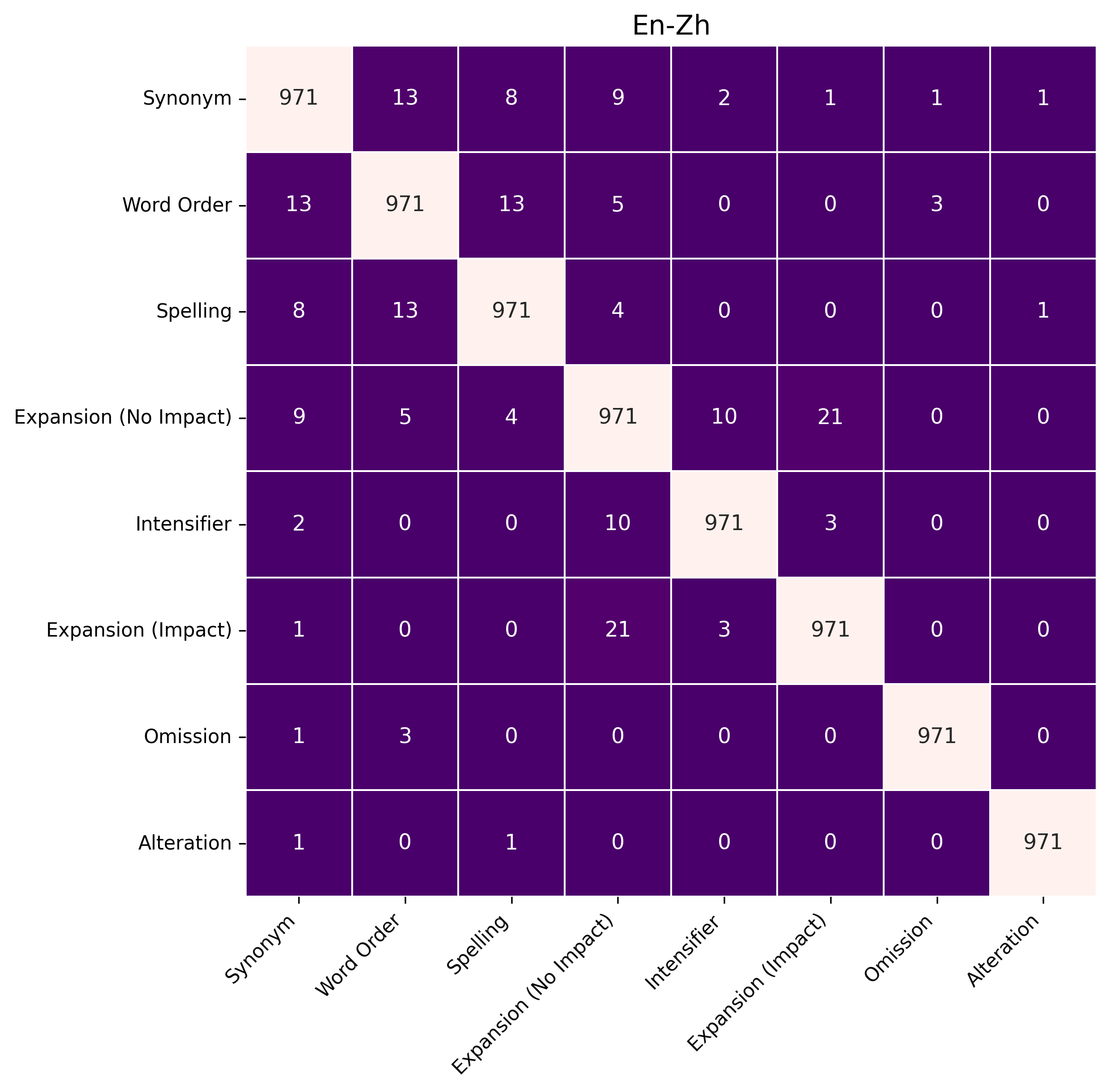}}

    \caption{Overlap ratio between perturbed translations from different perturbation types. We show that there are mostly no overlaps.with one exception of Expansion (No Impact) and Intensifier.}
    \label{fig:overlap_ratio}
\end{figure*}
\clearpage

\begin{table*}
\centering
\resizebox{0.7\linewidth}{!}{%
    \begin{tabular}{llllllll}
    \specialrule{1.3pt}{0pt}{0pt}
    \textbf{Language} & \textbf{Severity} & \textbf{Perturbation} & \textbf{English QA} & \textbf{Cross-lingual QA} \\
    \toprule

    \multirow{8}{*}{\textsc{\textbf{En-Es}}} & \multirow{5}{*}{\hl{Minor}} & Spelling & 0.916 & 0.263 \\
     &  & Word Order  & 0.904 & 0.254 \\
     &  & Synonym & 0.904 & 0.255 \\
     &  & Intensifier & 0.900 & 0.251 \\
     &  & Expansion (No Impact) & 0.897 & 0.247 \\
     \cmidrule{2-5}
     & \multirow{3}{*}{\hlred{Critical}} & Expansion (Impact) & 0.896 & 0.241 \\
     &  & Omission & 0.864 & 0.244 \\
     &  & Alteration & 0.762 & 0.232 \\
     \midrule
     
    \multirow{8}{*}{\textsc{\textbf{En-Fr}}} & \multirow{5}{*}{\hl{Minor}} & Spelling & 0.907 & 0.248 \\
     &  & Word Order  & 0.902 & 0.239 \\
     &  & Synonym & 0.892 & 0.242 \\
     &  & Intensifier & 0.897 & 0.234 \\
     &  & Expansion (No Impact) & 0.903 & 0.232 \\
     \cmidrule{2-5}
     & \multirow{3}{*}{\hlred{Critical}} & Expansion (Impact) & 0.890 & 0.226 \\
     &  & Omission & 0.863 & 0.228 \\
     &  & Alteration & 0.729 & 0.228 \\
     \midrule
    
    \multirow{8}{*}{\textsc{\textbf{En-Hi}}} & \multirow{5}{*}{\hl{Minor}} & Spelling & 0.907 & 0.167 \\
     &  & Word Order  & 0.866 & 0.156 \\
     &  & Synonym & 0.879 & 0.138 \\
     &  & Intensifier & 0.893 & 0.141 \\
     &  & Expansion (No Impact) & 0.901 & 0.139 \\
     \cmidrule{2-5}
     & \multirow{3}{*}{\hlred{Critical}} & Expansion (Impact) & 0.879 & 0.134 \\
     &  & Omission & 0.825 & 0.132 \\
     &  & Alteration & 0.809 & 0.136 \\
     \midrule
     
    \multirow{8}{*}{\textsc{\textbf{En-Tl}}} & \multirow{5}{*}{\hl{Minor}} & Spelling & 0.906 & 0.254 \\
     &  & Word Order  & 0.902 & 0.238 \\
     &  & Synonym & 0.900 & 0.232 \\
     &  & Intensifier & 0.900 & 0.227 \\
     &  & Expansion (No Impact) & 0.898 & 0.230 \\
     \cmidrule{2-5}
     & \multirow{3}{*}{\hlred{Critical}} & Expansion (Impact) & 0.888 & 0.228 \\
     &  & Omission & 0.847 & 0.211 \\
     &  & Alteration & 0.778 & 0.209 \\
     \midrule
     
    \multirow{8}{*}{\textsc{\textbf{En-Zh}}} & \multirow{5}{*}{\hl{Minor}} & Spelling & 0.880 & 0.158 \\
     &  & Word Order  & 0.879 & 0.162 \\
     &  & Synonym & 0.876 & 0.165 \\
     &  & Intensifier & 0.879 & 0.156 \\
     &  & Expansion (No Impact) & 0.876 & 0.157 \\
     \cmidrule{2-5}
     & \multirow{3}{*}{\hlred{Critical}} & Expansion (Impact) & 0.872 & 0.147 \\
     &  & Omission & 0.817 & 0.130 \\
     &  & Alteration & 0.758 & 0.138 \\

    \specialrule{1.3pt}{0pt}{0pt}
    \end{tabular}
}
\caption{\textsc{AskQE} scores using \textsc{SentenceBERT} for English and cross-lingual QA system.} 
\label{tab:crosslingual_qa}
\end{table*}
\clearpage

\begin{table*}
\centering
\resizebox{\linewidth}{!}{%
    \begin{tabular}{llll}
    \specialrule{1.3pt}{0pt}{0pt}
    \textbf{Source} & \textbf{Domain} & \textbf{\# Sent} & \textbf{Avg. Length} \\
    \toprule

    \textbf{CMU} & Medical, Conversational & 379 & 8.86 \\
    \textbf{PubMed} & Medical, Scientific (COVID-19 related articles) & 263 & 22.98 \\
    \textbf{Wikinews} & News (COVID-19 related news articles) & 21 & 20.10 \\
    \textbf{Wikivoyage} & Travel (Summary of travel restrictions) & 243 & 18.61 \\
    \textbf{Wikipedia} & General (COVID-19 related articles) & 383 & 23.57 \\
    \textbf{Wikisource} & Announcements (Government/organization announcements) & 24 & 29.25 \\
    \midrule
    \textbf{Total} & & \textbf{971} & \textbf{24.79} \\
    \specialrule{1.3pt}{0pt}{0pt}
    \end{tabular}
}
\caption{Dataset statistics for \textsc{TICO-19} evaluation dataset \cite{anastasopoulos-etal-2020-tico}. We categorize the development split according to data source. \textbf{\# Sent:} Number of sentences; \textbf{Avg. Length:} Average length of sentences.} 
\label{tab:tico_stats}
\end{table*}
\begin{table*}
\centering
\resizebox{\linewidth}{!}{%
    \begin{tabular}{c p{0.12\textwidth} p{0.3\textwidth} p{0.3\textwidth} p{0.3\textwidth} p{0.25\textwidth} p{0.25\textwidth}}
    \specialrule{1.3pt}{0pt}{0pt}
    \textbf{Severity} & \textbf{Perturbation} & \textbf{$X_{\mathrm{src}}$} & \textbf{$Y_{\mathrm{tgt}}$} & \textbf{$Q_{\mathrm{src}}$} & \textbf{$A_{\mathrm{src}}$} & \textbf{$A_{\mathrm{bt}}$} \\
    \toprule

    \multirow{25}{*}{\hl{\textbf{Minor}}} & Spelling & it is right in the center of my chest & es justo en el centro de mi puedo & [`Where is it in relation to your chest?', `Is it exactly in the center of your chest?'] & [`In the center of my chest', `Yes'] & [`In the center of my can', `No'] \\
    \cmidrule{2-7}

    & Word Order & Disinfect the surfaces with a suitable disinfectant, such as diluted household bleach. & Desinfecte las superficies con lejía de uso doméstico diluida, un desinfectante adecuado. & [`What disinfectant should be used to disinfect the surfaces?', `What should be used to disinfect the surfaces?'] & [`A suitable disinfectant', `A suitable disinfectant, such as diluted household bleach'] & [`Diluted household bleach', `Diluted household bleach'] \\
    \cmidrule{2-7}

    & Synonym & Remove it from the back, throw it away, and then wash your hands. & Quíteselo por la parte trasera, deséchelo y límpiese las manos. & [`What should you do after throwing it away?', `What should you do after removing it from the back?'] & [`Wash your hands', `Throw it away'] & [`Clean your hands', `Discard it'] \\
    \cmidrule{2-7}

    & Intensifier & or if you have high blood pressure & o si tiene presión arterial muy alta & [`What might does he have?'] & [`high blood pressure'] & [`very high blood pressure'] \\
    \cmidrule{2-7}

    & Expansion (No Impact) & With the COVID-19 outbreak, PHE and RCGP RSC have adapted existing influenza surveillance to monitor the spread of COVID-19 in the community, and this protocol sets out the basis for that collaboration. & Con el brote de la COVID-19, PHE y el RSC del RCGP han adaptado la actual vigilancia de la influenza para controlar la propagación de la COVID-19 en la comunidad, y este protocolo conjunto sienta las bases para esa colaboración. & [`What is the basis for the collaboration between PHE and RCGP RSC?', `Why have PHE and RCGP RSC adapted existing influenza surveillance?', `What is the purpose of adapting existing influenza surveillance?', `What is the collaboration between PHE and RCGP RSC about?'] & [`This protocol', `The COVID-19 outbreak', `To monitor the spread of COVID-19 in the community', `The spread of COVID-19 in the community'] & [`The joint protocol', `The outbreak of COVID-19', `To monitor the spread of COVID-19 in the community', `Monitoring the spread of COVID-19 in the community'] \\
    \midrule

    \multirow{16}{*}{\hlred{\textbf{Critical}}} & Expansion (Impact) & mandatory real-name registration for public transit cards & registro obligatorio con el nombre real y la dirección para tarjetas de transporte público & [`Is real-name registration mandatory for public transit cards?', `What is mandatory for public transit cards?'] & [`Yes', `Real-name registration'] & [`Yes', `Registration with real name and address'] \\
    \cmidrule{2-7}

    & Omission & The genome size for coronaviruses ranges from 26.4 to 31.7 kilobases. & El tamaño del genoma de los coronavirus varía de 31,7 kilobases. & [`What is the range of genome size for coronaviruses?', `What is the lower limit of genome size for coronaviruses?', `What is the upper limit of genome size for coronaviruses?'] & [`26.4 to 31.7 kilobases', `26.4 kilobases', `31.7 kilobases'] & [`31.7 kilobases', `31.7 kilobases', `31.7 kilobases'] \\
    \cmidrule{2-7}

    & Alteration & Among these 7,162 cases, 2,692 (37.6\%) patients had one or more underlying health condition or risk factor, and 4,470 (62.4\%) had none of these conditions reported. & Entre estos 7162 casos, 2692 (37,6 \%) pacientes no tenían una o más afecciones médicas subyacentes o factores de riesgo, y 4470 (62,4 \%) tenían alguna de estas afecciones informadas. & [`How many cases are there?', `How many patients had one or more underlying health condition or risk factor?', `What percentage of patients had one or more underlying health condition or risk factor?', `How many patients had none of these conditions reported?', `What percentage of patients had none of these conditions reported?'] & [`7,162', `2,692', `37.6\%', `4,470', `62.4\%'] & [`7162', `4470', `62.4\%', `2692', `37.6\%'] \\

    \specialrule{1.3pt}{0pt}{0pt}
    \end{tabular}
}
\caption{Qualitative Examples of questions and answer pairs generated by \textsc{AskQE} for \textsc{En-Es} language pair. \textbf{$X_{\mathrm{src}}$}: Source sentence; \textbf{$Y_{\mathrm{tgt}}$}: Perturbed MT output; \textbf{$Q_{\mathrm{src}}$}: Questions generated from source; \textbf{$A_{\mathrm{src}}$}: Answers generated from source; \textbf{$A_{\mathrm{bt}}$}: Answers generated from the backtranslated MT output.} 
\label{tab:qualitative_es}
\end{table*}
\begin{CJK*}{UTF8}{gbsn}

\begin{table*}
\centering
\resizebox{\linewidth}{!}{%
    \begin{tabular}{c p{0.12\textwidth} p{0.3\textwidth} p{0.3\textwidth} p{0.3\textwidth} p{0.25\textwidth} p{0.25\textwidth}}
    \specialrule{1.3pt}{0pt}{0pt}
    \textbf{Severity} & \textbf{Perturbation} & \textbf{$X_{\mathrm{src}}$} & \textbf{$Y_{\mathrm{tgt}}$} & \textbf{$Q_{\mathrm{src}}$} & \textbf{$A_{\mathrm{src}}$} & \textbf{$A_{\mathrm{bt}}$} \\
    \toprule

    \multirow{25}{*}{\hl{\textbf{Minor}}} & Spelling & Thailand is using an app and SIM cards for all travelers to enforce their quarantine. & 泰国利用手机应用程序和 SIN 卡对所有游客实施隔离。 & [`What is Thailand using for all travelers?', `What is Thailand using to enforce travelers' quarantine?']& [`An app and SIM cards', `An app and SIM cards'] & [`A mobile app and SIN cards', `A mobile app and SIN cards'] \\
    \cmidrule{2-7}

    & Word Order & Some patients have very mild symptoms, similar to a cold. & 症状较轻的有些患者，类似感冒。 & [`What symptoms do some patients have?', `What are these symptoms similar to?'] & [`Very mild symptoms', `A cold'] & [`Milder symptoms', `A cold'] \\
    \cmidrule{2-7}

    & Synonym & Phase II trials are used to establish an initial reading of efficacy and further explore safety in small numbers of people having the disease targeted by the NCE. & II 期试验用于确定疗效的初始读数，并进一步探索对于一些 NCE 目标人群的安全性。 & [`What is the purpose of Phase II trials?', `What is further explored in Phase II trials?', `How many people are involved in Phase II trials?'] & [`To establish an initial reading of efficacy and further explore safety', `Safety', `Small numbers'] & [`An initial readout of efficacy and to further explore safety', `Safety', `Unknown'] \\
    \cmidrule{2-7}

    & Intensifier & Many of these underlying health conditions are common in the United States: based on self-reported 2018 data, the prevalence of diagnosed diabetes among U.S. adults was 10.1\%, and the U.S. age-adjusted prevalence of all types of heart disease (excluding hypertension without other heart disease) was 10.6\% in 2017. & 许多这些基础疾病在美国很常见：根据 2018 年自我报告的数据，美国成年人中诊断为高糖尿病的患病率为 10.1\%，美国所有类型心脏病（不伴其他心脏病的高血压除外）的年龄调整患病率在 2017 年为 10.6\%。& [`What are the underlying health conditions that are common in the United States?', `What percentage of U.S. adults have diagnosed diabetes based on self-reported 2018 data?', `What was the age-adjusted prevalence of all types of heart disease in the U.S. in 2017?'] & [`Diabetes and heart disease', `10.1\%', `10.6\%'] & [`Diagnosed high diabetes and all types of heart disease', `10.1\%', `10.6\%'] \\
    \cmidrule{2-7}

    & Expansion (No Impact) & now i send you an image & 现在我发给您一张详细图像 & [`What do you send?', `When do you send the image?'] & [`An image', `Now'] & [`A detailed image', `Now'] \\
    \midrule

    \multirow{16}{*}{\hlred{\textbf{Critical}}} & Expansion (Impact) & mandatory real-name registration for public transit cards & registro obligatorio con el nombre real y la dirección para tarjetas de transporte público & [`Is real-name registration mandatory for public transit cards?', `What is mandatory for public transit cards?'] & [`Yes', `Real-name registration'] & [`Yes', `Registration with real name and address'] \\
    \cmidrule{2-7}

    & Omission & Others in early-stage Phase II trials or numerous treatment candidates in Phase I trials, are also excluded. & 早期 II 期试验的其他药物或 I 期试验中的大量候选药物也不含在此列。 & [`Who are excluded from early-stage Phase II trials?', `Who are excluded from Phase I trials?'] & [`Others', `Numerous treatment candidates'] & [`Other drugs', `Large numbers of drug candidates'] \\
    \cmidrule{2-7}

    & Alteration & Phylogentically, mouse hepatitis virus (Murine coronavirus), which infects the mouse's liver and the central nervous system, is related to  human coronavirus OC43 and bovine coronavirus. & 从生物系统上来说，感染小鼠肝脏和中枢神经系统的小鼠肝炎病毒（鼠冠状病毒），与人冠状病毒 OC43 和猪冠状病毒有关。 & [`What is the relationship between mouse hepatitis virus and human coronavirus OC43?', `What is the relationship between mouse hepatitis virus and bovine coronavirus?', `What organs does mouse hepatitis virus infect in a mouse?'] & [`Mouse hepatitis virus is phylogenetically related to human coronavirus OC43', `Mouse hepatitis virus is phylogenetically related to bovine coronavirus', `The liver and the central nervous system"] & [`Mouse hepatitis virus is related to human coronavirus OC43', `There is no relationship mentioned between mouse hepatitis virus and bovine coronavirus', `The liver and central nervous system'] \\

    \specialrule{1.3pt}{0pt}{0pt}
    \end{tabular}
}
\caption{Qualitative Examples of questions and answer pairs generated by \textsc{AskQE} for \textsc{En-Zh} language pair. \textbf{$X_{\mathrm{src}}$}: Source sentence; \textbf{$Y_{\mathrm{tgt}}$}: Perturbed MT output; \textbf{$Q_{\mathrm{src}}$}: Questions generated from source; \textbf{$A_{\mathrm{src}}$}: Answers generated from source; \textbf{$A_{\mathrm{bt}}$}: Answers generated from the backtranslated MT output.} 
\label{tab:qualitative_zh}
\end{table*}

\end{CJK*}

\begin{table*}
\centering
\resizebox{400pt}{!}{%
    \begin{tabular}{llllllll}
    \specialrule{1.3pt}{0pt}{0pt}
    \textbf{Language} & \textbf{Severity} & \textbf{Perturbation} & \textbf{\textsc{xCOMET-QE} (↑)} & \textbf{\textsc{MetricX-QE} (↓)} & \textbf{BT-Score (↑)} \\
    \toprule

    \multirow{8}{*}{\textsc{\textbf{En-Es}}} & \multirow{5}{*}{\hl{Minor}} & Spelling & 0.926 & 1.832 & 0.924 \\
     &  & Word Order  & 0.910 & 2.044 & 0.916 \\
     &  & Synonym & 0.882 & 3.325 & 0.925 \\
     &  & Intensifier & 0.908 & 2.175 & 0.917 \\
     &  & Expansion (No Impact) & 0.904 & 2.394 & 0.920 \\
     \cmidrule{2-6}
     & \multirow{3}{*}{\hlred{Critical}} & Expansion (Impact) & 0.885 & 2.519 & 0.905 \\
     &  & Omission & 0.870 & 3.462 & 0.901 \\
     &  & Alteration & 0.712 & 6.510 & 0.871 \\
     \midrule
     
    \multirow{8}{*}{\textsc{\textbf{En-Fr}}} & \multirow{5}{*}{\hl{Minor}} & Spelling & 0.910 & 2.020 & 0.926 \\
     &  & Word Order  & 0.889 & 2.324 & 0.912 \\
     &  & Synonym & 0.845 & 3.682 & 0.917 \\
     &  & Intensifier & 0.889 & 2.320 & 0.920 \\
     &  & Expansion (No Impact) & 0.873 & 2.676 & 0.920 \\
     \cmidrule{2-6}
     & \multirow{3}{*}{\hlred{Critical}} & Expansion (Impact) & 0.842 & 2.897 & 0.904 \\
     &  & Omission & 0.830 & 3.730 & 0.903 \\
     &  & Alteration & 0.573 & 6.718 & 0.857 \\
     \midrule
     
    \multirow{8}{*}{\textsc{\textbf{En-Hi}}} & \multirow{5}{*}{\hl{Minor}} & Spelling & 0.630 & 3.334 & 0.924 \\
     &  & Word Order  & 0.631 & 3.946 & 0.902 \\
     &  & Synonym & 0.611 & 3.237 & 0.927 \\
     &  & Intensifier & 0.595 & 3.359 & 0.921 \\
     &  & Expansion (No Impact) & 0.571 & 3.605 & 0.920 \\
     \cmidrule{2-6}
     & \multirow{3}{*}{\hlred{Critical}} & Expansion (Impact) & 0.534 & 4.141 & 0.899 \\
     &  & Omission & 0.517 & 4.599 & 0.884 \\
     &  & Alteration & 0.491 & 5.391 & 0.890 \\
     \midrule
     
    \multirow{8}{*}{\textsc{\textbf{En-Tl}}} & \multirow{5}{*}{\hl{Minor}} & Spelling & 0.771 & 2.548 & 0.939 \\
     &  & Word Order  & 0.730 & 2.830 & 0.918 \\
     &  & Synonym & 0.738 & 3.463 & 0.943 \\
     &  & Intensifier & 0.734 & 2.804 & 0.928 \\
     &  & Expansion (No Impact) & 0.723 & 3.274 & 0.933 \\
     \cmidrule{2-6}
     & \multirow{3}{*}{\hlred{Critical}} & Expansion (Impact) & 0.687 & 3.336 & 0.911 \\
     &  & Omission & 0.705 & 4.085 & 0.902 \\
     &  & Alteration & 0.593 & 5.887 & 0.893 \\
     \midrule
     
    \multirow{8}{*}{\textsc{\textbf{En-Zh}}} & \multirow{5}{*}{\hl{Minor}} & Spelling & 0.839 & 1.626 & 0.894 \\
     &  & Word Order  & 0.818 & 2.113 & 0.877 \\
     &  & Synonym & 0.802 & 2.929 & 0.874 \\
     &  & Intensifier & 0.794 & 1.741 & 0.887 \\
     &  & Expansion (No Impact) & 0.782 & 1.823 & 0.887 \\
     \cmidrule{2-6}
     & \multirow{3}{*}{\hlred{Critical}} & Expansion (Impact) & 0.781 & 1.899 & 0.874 \\
     &  & Omission & 0.768 & 2.428 & 0.863 \\
     &  & Alteration & 0.666 & 3.856 & 0.850 \\
    
    \specialrule{1.3pt}{0pt}{0pt}
    \end{tabular}
}
\caption{Average metric scores for each of three QE metric: \textsc{xCOMET-QE}, \textsc{MetricX-QE}, BT-Score.} 
\label{tab:raw_qe}
\end{table*}

\definecolor{mygray}{RGB}{220,220,220} % Light gray

\begin{table*}[!htp]
\centering
\resizebox{\linewidth}{!}{%
    \begin{tabular}{l l l}
    \specialrule{1.3pt}{0pt}{0pt}
    \textbf{Desiderata} & \textbf{Level} & \textbf{Description} \\
    \toprule
    \rowcolors{2}{mygray}{white}

    \textbf{Empty} & I & Number of empty questions. \\
    \textbf{Duplicate} & I & Number of duplicated questions. \\
    \textbf{Diversity} & I & Output diversity of questions measured by average Sentence-BERT similarity \citep{reimers-gurevych-2019-sentence}. \\
    \textbf{Answerability} & Q & Answerability of question given the source measured by SelfCheckGPT \citep{manakul-etal-2023-selfcheckgpt}. \\
    \textbf{Readability} & Q & Readability of question measured by Flesch Reading Ease score \citep{Flesch1948}. \\
    \textbf{Answerability} & Q & Answerability of question given the source measured by SelfCheckGPT \citep{manakul-etal-2023-selfcheckgpt}. \\
    % \textbf{Error Coverage} & Q & Ratio of questions targeting the MT error measured by LLM-as-a-judge. \\

    \specialrule{1.3pt}{0pt}{0pt}
    \end{tabular}
}
\caption{Six quality desiderata. \textbf{Level:} Level of measurement (\textbf{I:} Instance, \textbf{Q:} Question-level).}
\label{tab:desiderata}
\end{table*}
\begin{table*}
\centering
\resizebox{\linewidth}{!}{%
    \begin{tabular}{lllllllll}
    \specialrule{1.3pt}{0pt}{0pt}
    \textbf{Model} & \textbf{Variant} & \textbf{Avg.\# Q} & \textbf{Empty (↓)} & \textbf{Duplicate (↓)}& \textbf{Diversity (↑)}& \textbf{Answerability (↑)} & \textbf{Readability (↓)}\\
    \toprule

    \multirow{3}{*}{\textbf{\textsc{Gemma-2 9b}}} & Vanilla & 3.04 & 0 & 0 & 0.534 & 88.70 & 66.44 \\
    & NLI & 2.92 & 0 & 0 & 0.559 & 90.33 & 68.28 \\
    & SRL & 2.86 & 0 & 0 & 0.559 & 91.17 & 66.87 \\
    \midrule

    \multirow{3}{*}{\textbf{\textsc{Gemma-2 27b}}} & Vanilla & 2.82 & 0 & 0 & 0.539 & 90.25 & 68.25 \\
    & NLI & 2.43 & 0 & 1 & 0.531 & 91.19 & 69.41 \\
    & SRL & 2.81 & 0 & 0 & 0.500 & 90.27 & 71.57 \\
    \midrule

    \multirow{3}{*}{\textbf{\textsc{LLaMA-3 8b}}} & Vanilla & 2.18 & 0 & 0 & 0.597 & 84.88 & 69.73 \\
    & NLI & 4.21 & 0 & 1 & 0.633 & 90.19 & \textbf{60.92} \\
    & SRL & 4.21 & 0 & 1 & 0.575 & 87.32 & 66.57 \\
    \midrule

    \multirow{3}{*}{\textbf{\textsc{LLaMA-3 70b}}} & Vanilla & 5.03 & 0 & 0 & 0.520 & 86.14 & 68.98 \\
    & NLI & 3.37 & 0 & 0 & \textbf{0.634} & \textbf{92.92} & 65.89 \\
    & SRL & 5.05 & 0 & 1 & 0.574 & 88.31 & 66.78 \\
    \midrule

    \multirow{3}{*}{\textbf{\textsc{Yi-1.5 9b}}} & Vanilla & 2.82 & 0 & 0 & 0.569 & 90.42 & 63.96 \\
    & NLI & 2.78 & 0 & 2 & 0.586 & 89.97 & 66.09 \\
    & SRL & 3.33 & 0 & 0 & 0.586 & 90.59 & 63.62 \\
    
    \specialrule{1.3pt}{0pt}{0pt}
    \end{tabular}
}
\caption{Desiderata evaluation for each LLM configuration (\textbf{Model, Variant}). \textsc{LLaMA-3 70b} with NLI has the highest diversity and answerability score.} 
\label{tab:detailed_desiderata}
\end{table*}

\begin{table*}[h]
    \centering
    \resizebox{\textwidth}{!}{%
    \begin{tabular}{clllllll}
    \specialrule{1.3pt}{0pt}{0pt}
    \textbf{Language} & \textbf{Severity} & \textbf{Perturbation} & \textbf{\textsc{F1}} & \textbf{\textsc{EM}} & \textbf{\textsc{chrF}} & \textbf{\textsc{BLEU}} & \textbf{\textsc{SBERT}} \\
    \toprule

    \multirow{8}{*}{\textsc{\textbf{En-Es}}} & \multirow{5}{*}{\hl{Minor}} & Spelling & 0.697 / 0.739 / 0.726 & 0.413 / 0.430 / 0.429 & 72.89 / 76.58 / 75.80 & 57.18 / 60.89 / 60.09 & 0.862 / 0.888 / 0.877 \\
 &  & Word Order & 0.678 / 0.715 / 0.712 & 0.389 / 0.402 / 0.410 & 71.41 / 74.58 / 74.37 & 54.60 / 57.64 / 57.73 & 0.850 / 0.874 / 0.869 \\
 &  & Synonym & 0.677 / 0.720 / 0.706 & 0.388 / 0.399 / 0.406 & 70.36 / 73.88 / 73.20 & 55.03 / 58.39 / 57.99 & 0.848 / 0.874 / 0.864 \\
 &  & Intensifier & 0.665 / 0.697 / 0.690 & 0.341 / 0.358 / 0.358 & 70.69 / 73.60 / 72.60 & 51.60 / 54.82 / 53.88 & 0.845 / 0.867 / 0.858 \\
 &  & Expansion (No Impact) & 0.663 / 0.710 / 0.686 & 0.346 / 0.367 / 0.358 & 70.46 / 73.99 / 72.95 & 52.12 / 55.34 / 54.43 & 0.848 / 0.871 / 0.857 \\
 \cmidrule{2-8}
 & \multirow{3}{*}{\hlred{Critical}} & Expansion (Impact) & 0.645 / 0.689 / 0.670 & 0.315 / 0.340 / 0.333 & 68.82 / 72.70 / 70.94 & 48.79 / 52.38 / 50.92 & 0.830 / 0.859 / 0.844 \\
 &  & Omission & 0.625 / 0.670 / 0.645 & 0.336 / 0.352 / 0.343 & 65.93 / 69.53 / 67.44 & 49.42 / 52.85 / 51.22 & 0.818 / 0.843 / 0.826 \\
 &  & Alteration & 0.577 / 0.605 / 0.597 & 0.298 / 0.293 / 0.294 & 62.42 / 64.64 / 63.78 & 44.83 / 46.64 / 46.31 & 0.779 / 0.803 / 0.798 \\
\midrule
 \multirow{8}{*}{\textsc{\textbf{En-Fr}}} & \multirow{5}{*}{\hl{Minor}} & Spelling & 0.693 / 0.730 / 0.881 & 0.410 / 0.415 / 0.420 & 72.36 / 75.26 / 74.87 & 56.72 / 59.51 / 59.25 & 0.859 / 0.864 / 0.876 \\
 &  & Word Order & 0.669 / 0.689 / 0.860 & 0.379 / 0.365 / 0.381 & 69.54 / 71.56 / 72.46 & 54.08 / 54.00 / 55.15 & 0.845 / 0.860 / 0.863 \\
 &  & Synonym & 0.668 / 0.702 / 0.864 & 0.379 / 0.374 / 0.378 & 70.14 / 71.75 / 71.27 & 53.32 / 56.09 / 55.77 & 0.845 / 0.881 / 0.858 \\
 &  & Intensifier & 0.667 / 0.699 / 0.868 & 0.350 / 0.363 / 0.350 & 70.01 / 72.94 / 72.62 & 52.17 / 54.79 / 53.83 & 0.846 / 0.866 / 0.862 \\
 &  & Expansion (No Impact) & 0.666 / 0.695 / 0.866 & 0.367 / 0.351 / 0.358 & 70.09 / 72.69 / 72.16 & 52.76 / 54.00 / 53.99 & 0.845 / 0.868 / 0.863 \\
 \cmidrule{2-8}
 & \multirow{3}{*}{\hlred{Critical}} & Expansion (Impact) & 0.641 / 0.668 / 0.846 & 0.334 / 0.329 / 0.335 & 68.34 / 70.84 / 70.59 & 49.37 / 50.61 / 50.81 & 0.829 / 0.846 / 0.843 \\
 &  & Omission & 0.641 / 0.670 / 0.848 & 0.346 / 0.345 / 0.342 & 66.92 / 68.85 / 68.04 & 50.82 / 52.26 / 51.53 & 0.831 / 0.848 / 0.840 \\
 &  & Alteration & 0.541 / 0.561 / 0.770 & 0.268 / 0.264 / 0.269 & 58.17 / 59.87 / 60.75 & 41.66 / 42.66 / 43.42 & 0.751 / 0.770 / 0.772 \\
\midrule
 \multirow{8}{*}{\textsc{\textbf{En-Hi}}} & \multirow{5}{*}{\hl{Minor}} & Spelling & 0.704 / 0.737 / 0.732 & 0.429 / 0.439 / 0.440 & 74.25 / 76.90 / 76.78 & 58.99 / 61.96 / 61.37 & 0.864 / 0.882 / 0.881 \\
 &  & Word Order & 0.683 / 0.673 / 0.698 & 0.387 / 0.375 / 0.385 & 70.98 / 70.67 / 72.50 & 55.75 / 53.93 / 56.95 & 0.821 / 0.836 / 0.841 \\
 &  & Synonym & 0.644 / 0.709 / 0.670 & 0.361 / 0.388 / 0.366 & 67.84 / 73.42 / 70.62 & 51.56 / 58.00 / 53.36 & 0.842 / 0.857 / 0.851 \\
 &  & Intensifier & 0.682 / 0.706 / 0.715 & 0.385 / 0.375 / 0.407 & 71.87 / 74.57 / 75.12 & 55.58 / 56.78 / 58.62 & 0.840 / 0.867 / 0.860 \\
 &  & Expansion (No Impact) & 0.666 / 0.718 / 0.692 & 0.360 / 0.407 / 0.368 & 70.80 / 75.19 / 73.33 & 53.20 / 59.10 / 55.54 & 0.847 / 0.870 / 0.869 \\
 \cmidrule{2-8}
 & \multirow{3}{*}{\hlred{Critical}} & Expansion (Impact) & 0.653 / 0.687 / 0.675 & 0.351 / 0.366 / 0.357 & 69.26 / 72.46 / 71.64 & 51.33 / 53.95 / 53.01 & 0.825 / 0.845 / 0.843 \\
 &  & Omission & 0.605 / 0.634 / 0.613 & 0.339 / 0.343 / 0.331 & 63.73 / 66.12 / 64.76 & 48.51 / 50.91 / 48.91 & 0.794 / 0.808 / 0.802 \\
 &  & Alteration & 0.606 / 0.640 / 0.639 & 0.327 / 0.331 / 0.335 & 64.76 / 67.58 / 67.64 & 48.57 / 51.24 / 51.17 & 0.795 / 0.818 / 0.819 \\
\midrule
 \multirow{8}{*}{\textsc{\textbf{En-Tl}}} & \multirow{5}{*}{\hl{Minor}} & Spelling & 0.753 / 0.783 / 0.775 & 0.497 / 0.507 / 0.501 & 77.31 / 80.11 / 79.80 & 64.92 / 67.72 / 66.90 & 0.885 / 0.900 / 0.897 \\
 &  & Word Order & 0.718 / 0.729 / 0.747 & 0.442 / 0.445 / 0.450 & 73.64 / 75.61 / 76.58 & 61.20 / 61.23 / 63.71 & 0.857 / 0.871 / 0.880 \\
 &  & Synonym & 0.707 / 0.754 / 0.740 & 0.440 / 0.457 / 0.454 & 73.68 / 76.65 / 76.74 & 59.49 / 64.40 / 62.43 & 0.860 / 0.881 / 0.881 \\
 &  & Intensifier & 0.718 / 0.735 / 0.749 & 0.417 / 0.401 / 0.430 & 74.87 / 76.82 / 77.68 & 59.02 / 60.25 / 61.66 & 0.864 / 0.881 / 0.883 \\
 &  & Expansion (No Impact) & 0.706 / 0.750 / 0.735 & 0.405 / 0.422 / 0.410 & 74.48 / 77.55 / 76.99 & 58.16 / 61.48 / 60.63 & 0.866 / 0.887 / 0.886 \\
 \cmidrule{2-8}
 & \multirow{3}{*}{\hlred{Critical}} & Expansion (Impact) & 0.690 / 0.729 / 0.725 & 0.376 / 0.387 / 0.400 & 72.88 / 76.04 / 75.90 & 54.95 / 58.17 / 58.37 & 0.848 / 0.870 / 0.868 \\
 &  & Omission & 0.657 / 0.694 / 0.666 & 0.391 / 0.399 / 0.379 & 68.09 / 70.29 / 68.70 & 54.50 / 56.94 / 54.24 & 0.823 / 0.842 / 0.829 \\
 &  & Alteration & 0.620 / 0.648 / 0.657 & 0.337 / 0.342 / 0.360 & 65.68 / 68.01 / 69.03 & 50.27 / 52.64 / 53.70 & 0.801 / 0.817 / 0.827 \\
\midrule
 \multirow{8}{*}{\textsc{\textbf{En-Zh}}} & \multirow{5}{*}{\hl{Minor}} & Spelling & 0.592 / 0.613 / 0.620 & 0.296 / 0.295 / 0.306 & 64.70 / 65.85 / 66.93 & 46.12 / 47.42 / 48.43 & 0.807 / 0.821 / 0.822 \\
 &  & Word Order & 0.592 / 0.603 / 0.617 & 0.298 / 0.294 / 0.311 & 64.15 / 64.83 / 66.35 & 45.98 / 46.49 / 48.24 & 0.804 / 0.812 / 0.819 \\
 &  & Synonym & 0.588 / 0.617 / 0.613 & 0.299 / 0.304 / 0.308 & 64.10 / 65.93 / 66.58 & 45.73 / 47.75 / 47.92 & 0.804 / 0.823 / 0.818 \\
 &  & Intensifier & 0.581 / 0.604 / 0.607 & 0.286 / 0.281 / 0.292 & 64.07 / 65.67 / 66.52 & 44.65 / 45.57 / 46.52 & 0.802 / 0.820 / 0.815 \\
 &  & Expansion (No Impact) & 0.579 / 0.611 / 0.603 & 0.278 / 0.290 / 0.282 & 63.93 / 65.99 / 65.98 & 44.16 / 46.53 / 45.79 & 0.801 / 0.821 / 0.818 \\
 \cmidrule{2-8}
 & \multirow{3}{*}{\hlred{Critical}} & Expansion (Impact) & 0.557 / 0.589 / 0.578 & 0.252 / 0.265 / 0.263 & 61.98 / 64.28 / 63.99 & 40.81 / 43.12 / 42.61 & 0.783 / 0.806 / 0.797 \\
 &  & Omission & 0.537 / 0.562 / 0.554 & 0.262 / 0.270 / 0.264 & 58.53 / 60.07 / 59.83 & 41.30 / 42.93 / 42.35 & 0.765 / 0.782 / 0.775 \\
 &  & Alteration & 0.514 / 0.529 / 0.526 & 0.233 / 0.229 / 0.224 & 57.01 / 57.84 / 58.12 & 38.83 / 39.43 / 39.00 & 0.747 / 0.763 / 0.761 \\

    \specialrule{1.3pt}{0pt}{0pt}
    \end{tabular}

}
    \caption{Detailed results of \askqe using \textsc{Gemma-2 9b}. Each cell presents scores for Vanilla / SRL / NLI variant. Each metric quantifies the overlap between answers derived from the source sentence ($A_{\mathrm{src}}$) and from the backtranslated MT output ($A_{\mathrm{bt}}$). \textbf{\textsc{SBERT}}: \textsc{SentenceBERT}.}
    \label{tab:detailed_gemma_9b}
\end{table*}

\begin{table*}[h]
    \centering
    \resizebox{\textwidth}{!}{%
    \begin{tabular}{clllllll}
    \specialrule{1.3pt}{0pt}{0pt}
    \textbf{Language} & \textbf{Severity} & \textbf{Perturbation} & \textbf{\textsc{F1}} & \textbf{\textsc{EM}} & \textbf{\textsc{chrF}} & \textbf{\textsc{BLEU}} & \textbf{\textsc{SBERT}} \\
    \toprule

    \multirow{8}{*}{\textsc{\textbf{En-Es}}} & \multirow{5}{*}{\hl{Minor}} & Spelling & 0.735 / 0.742 / 0.735 & 0.434 / 0.425 / 0.409 & 76.29 / 77.00 / 76.71 & 60.71 / 60.77 / 59.82 & 0.886 / 0.897 / 0.895 \\
 &  & Word Order & 0.713 / 0.726 / 0.715 & 0.395 / 0.403 / 0.384 & 73.70 / 75.50 / 75.16 & 58.04 / 58.38 / 57.24 & 0.874 / 0.887 / 0.883 \\
 &  & Synonym & 0.711 / 0.723 / 0.713 & 0.406 / 0.396 / 0.376 & 74.33 / 74.41 / 74.09 & 57.55 / 58.67 / 57.37 & 0.875 / 0.884 / 0.879 \\
 &  & Intensifier & 0.701 / 0.714 / 0.704 & 0.363 / 0.362 / 0.343 & 73.80 / 74.75 / 74.52 & 54.79 / 55.46 / 53.92 & 0.866 / 0.881 / 0.876 \\
 &  & Expansion (No Impact) & 0.697 / 0.704 / 0.701 & 0.362 / 0.354 / 0.341 & 73.85 / 74.37 / 74.76 & 55.10 / 54.86 / 54.53 & 0.868 / 0.881 / 0.877 \\
 \cmidrule{2-8}
 & \multirow{3}{*}{\hlred{Critical}} & Expansion (Impact) & 0.682 / 0.687 / 0.687 & 0.342 / 0.331 / 0.323 & 72.22 / 72.75 / 73.06 & 52.14 / 51.97 / 51.31 & 0.853 / 0.865 / 0.863 \\
 &  & Omission & 0.657 / 0.676 / 0.661 & 0.347 / 0.353 / 0.334 & 68.30 / 70.09 / 68.95 & 51.95 / 53.31 / 51.81 & 0.834 / 0.851 / 0.841 \\
 &  & Alteration & 0.614 / 0.624 / 0.619 & 0.297 / 0.300 / 0.288 & 65.52 / 66.82 / 66.67 & 47.59 / 48.05 / 47.36 & 0.803 / 0.815 / 0.814 \\
\midrule
 \multirow{8}{*}{\textsc{\textbf{En-Fr}}} & \multirow{5}{*}{\hl{Minor}} & Spelling & 0.720 / 0.727 / 0.726 & 0.408 / 0.399 / 0.393 & 74.96 / 75.30 / 75.52 & 58.63 / 58.82 / 58.38 & 0.881 / 0.885 / 0.887 \\
 &  & Word Order & 0.705 / 0.708 / 0.709 & 0.381 / 0.377 / 0.372 & 72.57 / 73.25 / 73.41 & 56.62 / 55.87 / 55.41 & 0.869 / 0.874 / 0.876 \\
 &  & Synonym & 0.701 / 0.703 / 0.707 & 0.381 / 0.369 / 0.360 & 72.88 / 72.19 / 72.81 & 55.40 / 55.87 / 55.69 & 0.871 / 0.869 / 0.876 \\
 &  & Intensifier & 0.698 / 0.703 / 0.709 & 0.352 / 0.342 / 0.337 & 72.93 / 73.53 / 74.03 & 54.13 / 53.99 / 54.24 & 0.869 / 0.879 / 0.875 \\
 &  & Expansion (No Impact) & 0.693 / 0.703 / 0.697 & 0.359 / 0.351 / 0.337 & 73.05 / 73.85 / 73.44 & 54.32 / 54.68 / 53.63 & 0.869 / 0.874 / 0.879 \\
 \cmidrule{2-8}
 & \multirow{3}{*}{\hlred{Critical}} & Expansion (Impact) & 0.670 / 0.677 / 0.678 & 0.337 / 0.325 / 0.320 & 70.79 / 71.72 / 71.68 & 50.84 / 51.04 / 50.46 & 0.844 / 0.854 / 0.856 \\
 &  & Omission & 0.656 / 0.669 / 0.659 & 0.335 / 0.335 / 0.316 & 67.75 / 68.95 / 68.20 & 50.98 / 52.13 / 50.46 & 0.836 / 0.847 / 0.842 \\
 &  & Alteration & 0.565 / 0.587 / 0.584 & 0.264 / 0.270 / 0.258 & 60.45 / 62.49 / 62.45 & 42.69 / 44.32 / 43.65 & 0.765 / 0.784 / 0.783 \\
\midrule
 \multirow{8}{*}{\textsc{\textbf{En-Hi}}} & \multirow{5}{*}{\hl{Minor}} & Spelling & 0.740 / 0.742 / 0.744 & 0.440 / 0.432 / 0.415 & 77.81 / 77.96 / 78.08 & 62.24 / 62.16 / 61.83 & 0.889 / 0.894 / 0.898 \\
 &  & Word Order & 0.713 / 0.708 / 0.714 & 0.395 / 0.379 / 0.371 & 74.04 / 73.39 / 74.04 & 58.60 / 57.64 / 57.66 & 0.842 / 0.848 / 0.846 \\
 &  & Synonym & 0.677 / 0.679 / 0.678 & 0.371 / 0.369 / 0.350 & 70.82 / 71.23 / 71.07 & 53.88 / 54.26 / 53.46 & 0.864 / 0.863 / 0.865 \\
 &  & Intensifier & 0.715 / 0.718 / 0.724 & 0.403 / 0.394 / 0.388 & 75.35 / 75.59 / 76.21 & 58.73 / 58.60 / 58.82 & 0.870 / 0.873 / 0.875 \\
 &  & Expansion (No Impact) & 0.706 / 0.707 / 0.712 & 0.369 / 0.365 / 0.353 & 74.89 / 75.02 / 75.21 & 56.54 / 56.36 / 56.36 & 0.873 / 0.876 / 0.882 \\
 \cmidrule{2-8}
 & \multirow{3}{*}{\hlred{Critical}} & Expansion (Impact) & 0.687 / 0.690 / 0.699 & 0.368 / 0.359 / 0.341 & 72.88 / 73.00 / 73.78 & 54.38 / 53.85 / 54.14 & 0.846 / 0.851 / 0.858 \\
 &  & Omission & 0.618 / 0.629 / 0.612 & 0.333 / 0.329 / 0.310 & 64.84 / 66.15 / 64.60 & 49.40 / 50.02 / 48.21 & 0.796 / 0.814 / 0.798 \\
 &  & Alteration & 0.638 / 0.658 / 0.657 & 0.332 / 0.334 / 0.318 & 67.69 / 69.35 / 69.61 & 51.17 / 52.41 / 51.97 & 0.813 / 0.827 / 0.833 \\
\midrule
 \multirow{8}{*}{\textsc{\textbf{En-Tl}}} & \multirow{5}{*}{\hl{Minor}} & Spelling & 0.780 / 0.777 / 0.794 & 0.497 / 0.477 / 0.484 & 80.00 / 79.53 / 81.33 & 67.40 / 66.47 / 67.83 & 0.899 / 0.901 / 0.910 \\
 &  & Word Order & 0.758 / 0.757 / 0.771 & 0.454 / 0.440 / 0.449 & 77.04 / 77.15 / 78.47 & 64.50 / 64.34 / 65.32 & 0.880 / 0.881 / 0.889 \\
 &  & Synonym & 0.742 / 0.737 / 0.745 & 0.456 / 0.438 / 0.432 & 76.55 / 76.16 / 77.42 & 62.54 / 61.48 / 61.84 & 0.883 / 0.891 / 0.897 \\
 &  & Intensifier & 0.750 / 0.747 / 0.767 & 0.432 / 0.403 / 0.423 & 77.84 / 77.43 / 79.42 & 62.13 / 60.72 / 62.85 & 0.885 / 0.888 / 0.895 \\
 &  & Expansion (No Impact) & 0.750 / 0.737 / 0.755 & 0.426 / 0.388 / 0.395 & 78.04 / 77.17 / 78.73 & 62.25 / 59.82 / 61.44 & 0.886 / 0.892 / 0.901 \\
 \cmidrule{2-8}
 & \multirow{3}{*}{\hlred{Critical}} & Expansion (Impact) & 0.726 / 0.726 / 0.734 & 0.394 / 0.374 / 0.377 & 75.67 / 75.92 / 76.78 & 58.47 / 57.79 / 58.24 & 0.870 / 0.875 / 0.882 \\
 &  & Omission & 0.665 / 0.694 / 0.677 & 0.382 / 0.386 / 0.372 & 68.25 / 70.57 / 69.22 & 54.68 / 56.82 / 55.06 & 0.818 / 0.846 / 0.830 \\
 &  & Alteration & 0.667 / 0.675 / 0.678 & 0.362 / 0.347 / 0.351 & 69.60 / 70.86 / 71.17 & 54.70 / 54.61 / 55.08 & 0.823 / 0.838 / 0.838 \\
\midrule
 \multirow{8}{*}{\textsc{\textbf{En-Zh}}} & \multirow{5}{*}{\hl{Minor}} & Spelling & 0.628 / 0.621 / 0.632 & 0.325 / 0.295 / 0.294 & 67.21 / 66.27 / 67.81 & 49.22 / 47.73 / 48.28 & 0.823 / 0.825 / 0.835 \\
 &  & Word Order & 0.620 / 0.617 / 0.626 & 0.311 / 0.301 / 0.299 & 66.67 / 66.44 / 67.61 & 48.51 / 47.66 / 47.99 & 0.831 / 0.824 / 0.835 \\
 &  & Synonym & 0.622 / 0.615 / 0.620 & 0.308 / 0.292 / 0.281 & 66.45 / 66.25 / 67.30 & 48.56 / 47.37 / 47.21 & 0.828 / 0.830 / 0.840 \\
 &  & Intensifier & 0.620 / 0.613 / 0.625 & 0.305 / 0.278 / 0.282 & 67.07 / 66.59 / 67.86 & 47.68 / 46.15 / 46.80 & 0.820 / 0.830 / 0.840 \\
 &  & Expansion (No Impact) & 0.611 / 0.611 / 0.619 & 0.294 / 0.278 / 0.266 & 66.44 / 66.60 / 67.58 & 46.82 / 46.08 / 45.79 & 0.828 / 0.828 / 0.840 \\
 \cmidrule{2-8}
 & \multirow{3}{*}{\hlred{Critical}} & Expansion (Impact) & 0.595 / 0.590 / 0.599 & 0.273 / 0.255 / 0.254 & 65.24 / 64.76 / 66.26 & 43.94 / 42.60 / 43.07 & 0.811 / 0.813 / 0.825 \\
 &  & Omission & 0.548 / 0.563 / 0.557 & 0.265 / 0.263 / 0.241 & 58.80 / 60.17 / 59.62 & 41.62 / 42.70 / 40.85 & 0.769 / 0.784 / 0.783 \\
 &  & Alteration & 0.550 / 0.544 / 0.545 & 0.251 / 0.228 / 0.222 & 59.80 / 59.86 / 60.39 & 41.48 / 40.12 / 39.74 & 0.771 / 0.773 / 0.781 \\

    \specialrule{1.3pt}{0pt}{0pt}
    \end{tabular}

}
    \caption{Detailed results of \askqe using \textsc{Gemma-2 27b}. Each cell presents scores for Vanilla / SRL / NLI variant.}
    \label{tab:detailed_gemma_27b}
\end{table*}

\begin{table*}[h]
    \centering
    \resizebox{\textwidth}{!}{%
    \begin{tabular}{clllllll}
    \specialrule{1.3pt}{0pt}{0pt}
    \textbf{Language} & \textbf{Severity} & \textbf{Perturbation} & \textbf{\textsc{F1}} & \textbf{\textsc{EM}} & \textbf{\textsc{chrF}} & \textbf{\textsc{BLEU}} & \textbf{\textsc{SBERT}} \\
    \toprule
    
    \multirow{8}{*}{\textsc{\textbf{En-Es}}} & \multirow{5}{*}{\hl{Minor}} & Spelling & 0.696 / 0.703 / 0.715 & 0.357 / 0.373 / 0.381 & 72.06 / 72.43 / 73.89 & 54.32 / 55.68 / 56.73 & 0.830 / 0.828 / 0.833 \\
 &  & Word Order & 0.673 / 0.680 / 0.684 & 0.326 / 0.346 / 0.352 & 70.02 / 70.46 / 71.59 & 50.94 / 52.62 / 52.95 & 0.832 / 0.831 / 0.839 \\
 &  & Synonym & 0.672 / 0.677 / 0.680 & 0.321 / 0.338 / 0.337 & 69.38 / 69.47 / 70.04 & 51.46 / 52.69 / 52.89 & 0.842 / 0.845 / 0.856 \\
 &  & Intensifier & 0.660 / 0.665 / 0.678 & 0.289 / 0.303 / 0.307 & 69.32 / 69.24 / 71.02 & 48.22 / 49.48 / 50.21 & 0.827 / 0.825 / 0.837 \\
 &  & Expansion (No Impact) & 0.650 / 0.662 / 0.664 & 0.279 / 0.307 / 0.296 & 68.45 / 69.38 / 70.16 & 47.82 / 50.07 / 49.36 & 0.818 / 0.826 / 0.832 \\
 \cmidrule{2-8}
 & \multirow{3}{*}{\hlred{Critical}} & Expansion (Impact) & 0.634 / 0.641 / 0.660 & 0.263 / 0.275 / 0.289 & 67.01 / 67.18 / 69.10 & 44.94 / 46.36 / 47.62 & 0.805 / 0.806 / 0.823 \\
 &  & Omission & 0.611 / 0.638 / 0.637 & 0.273 / 0.299 / 0.294 & 63.76 / 65.75 / 65.55 & 45.15 / 48.22 / 48.04 & 0.786 / 0.800 / 0.802 \\
 &  & Alteration & 0.549 / 0.539 / 0.557 & 0.210 / 0.209 / 0.218 & 59.18 / 58.29 / 60.90 & 38.81 / 38.14 / 39.75 & 0.733 / 0.723 / 0.744 \\
\midrule
 \multirow{8}{*}{\textsc{\textbf{En-Fr}}} & \multirow{5}{*}{\hl{Minor}} & Spelling & 0.691 / 0.693 / 0.694 & 0.346 / 0.360 / 0.353 & 71.22 / 71.56 / 71.91 & 53.68 / 54.77 / 54.05 & 0.838 / 0.803 / 0.808 \\
 &  & Word Order & 0.668 / 0.671 / 0.672 & 0.320 / 0.329 / 0.330 & 69.01 / 68.97 / 69.07 & 50.10 / 51.92 / 51.35 & 0.830 / 0.806 / 0.814 \\
 &  & Synonym & 0.665 / 0.662 / 0.671 & 0.311 / 0.317 / 0.329 & 68.41 / 68.74 / 69.58 & 50.54 / 50.26 / 50.74 & 0.826 / 0.820 / 0.830 \\
 &  & Intensifier & 0.662 / 0.671 / 0.670 & 0.284 / 0.303 / 0.298 & 68.61 / 69.41 / 70.00 & 48.31 / 50.12 / 49.31 & 0.823 / 0.800 / 0.812 \\
 &  & Expansion (No Impact) & 0.653 / 0.656 / 0.669 & 0.284 / 0.301 / 0.305 & 68.24 / 68.63 / 69.95 & 47.91 / 49.14 / 49.49 & 0.822 / 0.802 / 0.807 \\
 \cmidrule{2-8}
 & \multirow{3}{*}{\hlred{Critical}} & Expansion (Impact) & 0.628 / 0.630 / 0.643 & 0.262 / 0.276 / 0.270 & 66.16 / 66.50 / 67.77 & 44.64 / 45.58 / 45.67 & 0.765 / 0.782 / 0.799 \\
 &  & Omission & 0.625 / 0.637 / 0.635 & 0.281 / 0.299 / 0.292 & 64.05 / 65.47 / 65.57 & 46.32 / 48.09 / 47.33 & 0.746 / 0.776 / 0.778 \\
 &  & Alteration & 0.516 / 0.501 / 0.517 & 0.187 / 0.185 / 0.183 & 56.11 / 54.64 / 55.96 & 35.72 / 34.92 / 35.71 & 0.693 / 0.701 / 0.722 \\
\midrule
 \multirow{8}{*}{\textsc{\textbf{En-Hi}}} & \multirow{5}{*}{\hl{Minor}} & Spelling & 0.697 / 0.699 / 0.706 & 0.369 / 0.383 / 0.381 & 72.96 / 72.79 / 73.59 & 55.66 / 56.88 / 56.57 & 0.796 / 0.763 / 0.768 \\
 &  & Word Order & 0.666 / 0.670 / 0.680 & 0.318 / 0.333 / 0.331 & 69.03 / 69.21 / 69.83 & 51.70 / 52.93 / 53.08 & 0.788 / 0.766 / 0.773 \\
 &  & Synonym & 0.643 / 0.644 / 0.644 & 0.302 / 0.318 / 0.306 & 67.17 / 67.32 / 67.33 & 48.25 / 49.39 / 48.48 & 0.785 / 0.779 / 0.788 \\
 &  & Intensifier & 0.674 / 0.681 / 0.686 & 0.328 / 0.348 / 0.336 & 70.47 / 70.95 / 71.63 & 51.86 / 53.63 / 53.09 & 0.782 / 0.760 / 0.772 \\
 &  & Expansion (No Impact) & 0.665 / 0.673 / 0.684 & 0.303 / 0.321 / 0.317 & 70.18 / 71.16 / 71.57 & 50.13 / 51.98 / 51.92 & 0.781 / 0.761 / 0.767 \\
 \cmidrule{2-8}
 & \multirow{3}{*}{\hlred{Critical}} & Expansion (Impact) & 0.634 / 0.649 / 0.663 & 0.282 / 0.305 / 0.309 & 66.78 / 68.11 / 69.83 & 46.83 / 48.80 / 49.74 & 0.727 / 0.743 / 0.759 \\
 &  & Omission & 0.580 / 0.606 / 0.594 & 0.265 / 0.289 / 0.280 & 61.07 / 63.37 / 62.22 & 43.40 / 46.60 / 45.04 & 0.708 / 0.737 / 0.739 \\
 &  & Alteration & 0.577 / 0.586 / 0.604 & 0.250 / 0.261 / 0.258 & 61.88 / 62.65 / 64.18 & 42.77 / 44.12 / 45.14 & 0.659 / 0.666 / 0.686 \\
\midrule
 \multirow{8}{*}{\textsc{\textbf{En-Tl}}} & \multirow{5}{*}{\hl{Minor}} & Spelling & 0.748 / 0.746 / 0.759 & 0.435 / 0.436 / 0.451 & 76.47 / 76.50 / 77.77 & 61.91 / 62.19 / 63.44 & 0.813 / 0.779 / 0.784 \\
 &  & Word Order & 0.723 / 0.722 / 0.737 & 0.393 / 0.401 / 0.414 & 73.28 / 73.62 / 74.89 & 59.03 / 59.72 / 60.68 & 0.805 / 0.782 / 0.789 \\
 &  & Synonym & 0.706 / 0.695 / 0.715 & 0.388 / 0.381 / 0.395 & 72.56 / 72.07 / 73.66 & 56.32 / 56.06 / 57.64 & 0.801 / 0.795 / 0.805 \\
 &  & Intensifier & 0.713 / 0.704 / 0.728 & 0.358 / 0.359 / 0.377 & 73.70 / 73.07 / 75.44 & 55.83 / 55.76 / 57.54 & 0.798 / 0.776 / 0.788 \\
 &  & Expansion (No Impact) & 0.699 / 0.698 / 0.716 & 0.333 / 0.350 / 0.356 & 72.94 / 72.89 / 74.57 & 54.11 / 55.15 / 56.13 & 0.797 / 0.778 / 0.783 \\
 \cmidrule{2-8}
 & \multirow{3}{*}{\hlred{Critical}} & Expansion (Impact) & 0.683 / 0.677 / 0.690 & 0.326 / 0.321 / 0.325 & 71.04 / 70.86 / 72.48 & 51.91 / 52.06 / 52.16 & 0.742 / 0.759 / 0.775 \\
 &  & Omission & 0.633 / 0.650 / 0.637 & 0.314 / 0.335 / 0.325 & 64.78 / 66.57 / 65.29 & 49.11 / 51.36 / 50.18 & 0.723 / 0.753 / 0.755 \\
 &  & Alteration & 0.588 / 0.596 / 0.615 & 0.255 / 0.272 / 0.277 & 62.27 / 63.16 / 65.23 & 44.20 / 45.68 / 46.59 & 0.672 / 0.680 / 0.700 \\
\midrule
 \multirow{8}{*}{\textsc{\textbf{En-Zh}}} & \multirow{5}{*}{\hl{Minor}} & Spelling & 0.588 / 0.592 / 0.600 & 0.244 / 0.255 / 0.262 & 63.31 / 63.31 / 64.56 & 42.80 / 43.76 / 44.22 & 0.871 / 0.869 / 0.875 \\
 &  & Word Order & 0.587 / 0.594 / 0.598 & 0.238 / 0.260 / 0.255 & 62.51 / 63.06 / 63.86 & 42.49 / 44.08 / 43.70 & 0.873 / 0.872 / 0.881 \\
 &  & Synonym & 0.579 / 0.588 / 0.594 & 0.239 / 0.258 / 0.250 & 62.51 / 62.79 / 64.03 & 41.82 / 43.39 / 43.43 & 0.884 / 0.887 / 0.898 \\
 &  & Intensifier & 0.585 / 0.588 / 0.598 & 0.232 / 0.244 / 0.247 & 63.15 / 63.15 / 64.77 & 41.61 / 42.54 / 42.90 & 0.868 / 0.866 / 0.879 \\
 &  & Expansion (No Impact) & 0.574 / 0.582 / 0.593 & 0.219 / 0.235 / 0.237 & 62.37 / 62.84 / 64.69 & 40.40 / 41.91 / 42.30 & 0.859 / 0.868 / 0.873 \\
 \cmidrule{2-8}
 & \multirow{3}{*}{\hlred{Critical}} & Expansion (Impact) & 0.561 / 0.564 / 0.574 & 0.208 / 0.223 / 0.221 & 61.21 / 61.31 / 63.12 & 38.38 / 39.15 / 39.52 & 0.845 / 0.847 / 0.864 \\
 &  & Omission & 0.527 / 0.556 / 0.534 & 0.206 / 0.243 / 0.225 & 56.47 / 58.81 / 57.49 & 37.26 / 40.61 / 38.32 & 0.825 / 0.840 / 0.842 \\
 &  & Alteration & 0.492 / 0.490 / 0.503 & 0.170 / 0.175 / 0.178 & 54.69 / 54.23 / 56.03 & 33.54 / 33.89 / 34.42 & 0.770 / 0.759 / 0.782 \\

    \specialrule{1.3pt}{0pt}{0pt}
    \end{tabular}

}
    \caption{Detailed results of \askqe using \textsc{LLaMA-3 8b}. Each cell presents scores for Vanilla / SRL / NLI variant.}
    \label{tab:detailed_llama_8b}
\end{table*}

\begin{table*}[h]
    \centering
    \resizebox{\textwidth}{!}{%
    \begin{tabular}{clllllll}
    \specialrule{1.3pt}{0pt}{0pt}
    \textbf{Language} & \textbf{Severity} & \textbf{Perturbation} & \textbf{\textsc{F1}} & \textbf{\textsc{EM}} & \textbf{\textsc{chrF}} & \textbf{\textsc{BLEU}} & \textbf{\textsc{SBERT}} \\
    \toprule
    
    \multirow{9}{*}{\textsc{\textbf{En-Es}}} & \multirow{5}{*}{\hl{Minor}} & Spelling & 0.648 / 0.655 / 0.840 & 0.341 / 0.338 / 0.704 & 68.58 / 69.21 / 85.28 & 51.50 / 51.73 / 78.10 & 0.809 / 0.818 / 0.916 \\
 &  & Word Order & 0.628 / 0.639 / 0.820 & 0.318 / 0.331 / 0.679 & 66.88 / 67.73 / 83.43 & 49.06 / 50.00 / 75.80 & 0.798 / 0.805 / 0.904 \\
 &  & Synonym & 0.627 / 0.628 / 0.816 & 0.326 / 0.308 / 0.676 & 66.30 / 66.10 / 82.71 & 49.64 / 48.71 / 75.63 & 0.797 / 0.802 / 0.904 \\
 &  & Intensifier & 0.631 / 0.626 / 0.812 & 0.305 / 0.286 / 0.655 & 67.45 / 66.60 / 82.85 & 48.39 / 46.74 / 73.96 & 0.799 / 0.800 / 0.900 \\
 &  & Expansion (No Impact) & 0.620 / 0.620 / 0.814 & 0.298 / 0.284 / 0.665 & 66.74 / 66.53 / 82.90 & 47.59 / 46.63 / 74.83 & 0.801 / 0.801 / 0.897 \\
 \cmidrule{2-8}

 & \multirow{3}{*}{\hlred{Critical}} & Expansion (Impact) & 0.606 / 0.609 / 0.809 & 0.287 / 0.265 / 0.653 & 65.37 / 65.31 / 82.46 & 45.45 / 44.26 / 73.49 & 0.786 / 0.787 / 0.896 \\
 &  & Omission & 0.571 / 0.587 / 0.746 & 0.279 / 0.280 / 0.602 & 61.21 / 62.30 / 75.83 & 44.13 / 44.81 / 68.33 & 0.760 / 0.765 / 0.864 \\
 &  & Alteration & 0.442 / 0.430 / 0.475 & 0.188 / 0.180 / 0.360 & 49.31 / 48.21 / 49.77 & 32.83 / 31.62 / 42.37 & 0.636 / 0.618 / 0.762 \\
\midrule

 \multirow{9}{*}{\textsc{\textbf{En-Fr}}} & \multirow{5}{*}{\hl{Minor}} & Spelling & 0.635 / 0.645 / 0.825 & 0.336 / 0.328 / 0.689 & 67.58 / 68.06 / 83.81 & 50.56 / 50.80 / 76.66 & 0.805 / 0.810 / 0.907 \\
 &  & Word Order & 0.628 / 0.628 / 0.818 & 0.323 / 0.310 / 0.681 & 66.54 / 66.14 / 82.93 & 48.93 / 48.23 / 75.65 & 0.802 / 0.798 / 0.902 \\
 &  & Synonym & 0.615 / 0.622 / 0.816 & 0.305 / 0.286 / 0.672 & 65.18 / 65.98 / 83.05 & 48.35 / 46.88 / 74.94 & 0.795 / 0.792 / 0.892 \\
 &  & Intensifier & 0.620 / 0.627 / 0.810 & 0.291 / 0.285 / 0.670 & 66.10 / 66.35 / 82.66 & 47.10 / 47.08 / 74.55 & 0.797 / 0.800 / 0.897 \\
 &  & Expansion (No Impact) & 0.613 / 0.619 / 0.796 & 0.300 / 0.300 / 0.652 & 65.63 / 64.77 / 80.76 & 47.21 / 47.97 / 73.44 & 0.797 / 0.799 / 0.903 \\
 \cmidrule{2-8}
 
 & \multirow{3}{*}{\hlred{Critical}} & Expansion (Impact) & 0.607 / 0.600 / 0.797 & 0.288 / 0.277 / 0.650 & 64.87 / 64.53 / 81.30 & 45.51 / 44.57 / 72.61 & 0.783 / 0.776 / 0.890 \\
 &  & Omission & 0.585 / 0.591 / 0.746 & 0.293 / 0.280 / 0.605 & 61.78 / 61.90 / 75.90 & 45.22 / 44.99 / 68.46 & 0.763 / 0.767 / 0.863 \\
 &  & Alteration & 0.409 / 0.384 / 0.408 & 0.175 / 0.159 / 0.305 & 46.52 / 43.43 / 43.17 & 30.03 / 27.90 / 36.10 & 0.609 / 0.575 / 0.729 \\
\midrule

 \multirow{9}{*}{\textsc{\textbf{En-Hi}}} & \multirow{5}{*}{\hl{Minor}} & Spelling & 0.657 / 0.651 / 0.810 & 0.365 / 0.352 / 0.681 & 70.23 / 69.62 / 83.24 & 53.52 / 53.04 / 75.73 & 0.820 / 0.811 / 0.907 \\
 &  & Word Order & 0.637 / 0.637 / 0.810 & 0.342 / 0.326 / 0.676 & 67.91 / 68.37 / 82.97 & 51.12 / 50.59 / 75.29 & 0.759 / 0.772 / 0.866 \\
 &  & Synonym & 0.623 / 0.618 / 0.798 & 0.327 / 0.308 / 0.665 & 66.03 / 65.59 / 81.83 & 50.01 / 48.93 / 73.96 & 0.783 / 0.780 / 0.879 \\
 &  & Intensifier & 0.625 / 0.619 / 0.778 & 0.316 / 0.295 / 0.645 & 67.46 / 66.94 / 79.57 & 49.04 / 48.00 / 72.47 & 0.794 / 0.792 / 0.893 \\
 &  & Expansion (No Impact) & 0.591 / 0.602 / 0.759 & 0.303 / 0.303 / 0.630 & 62.89 / 64.30 / 77.92 & 45.93 / 46.93 / 70.26 & 0.796 / 0.802 / 0.901 \\
 \cmidrule{2-8}
 
 & \multirow{3}{*}{\hlred{Critical}} & Expansion (Impact) & 0.605 / 0.607 / 0.767 & 0.304 / 0.293 / 0.627 & 65.39 / 65.29 / 79.20 & 46.80 / 46.48 / 70.14 & 0.775 / 0.776 / 0.879 \\
 &  & Omission & 0.540 / 0.541 / 0.670 & 0.274 / 0.265 / 0.547 & 58.08 / 58.37 / 69.29 & 42.20 / 42.01 / 61.61 & 0.717 / 0.723 / 0.825 \\
 &  & Alteration & 0.516 / 0.492 / 0.590 & 0.251 / 0.223 / 0.469 & 56.37 / 53.86 / 61.34 & 40.07 / 37.60 / 53.87 & 0.700 / 0.676 / 0.809 \\
\midrule

 \multirow{9}{*}{\textsc{\textbf{En-Tl}}} & \multirow{5}{*}{\hl{Minor}} & Spelling & 0.686 / 0.690 / 0.832 & 0.384 / 0.389 / 0.702 & 72.10 / 72.10 / 83.73 & 56.54 / 57.44 / 78.18 & 0.827 / 0.823 / 0.906 \\
 &  & Word Order & 0.664 / 0.665 / 0.815 & 0.366 / 0.354 / 0.682 & 69.68 / 69.31 / 82.54 & 54.81 / 54.66 / 76.35 & 0.812 / 0.801 / 0.902 \\
 &  & Synonym & 0.661 / 0.652 / 0.813 & 0.377 / 0.354 / 0.676 & 69.71 / 68.20 / 82.05 & 54.19 / 52.62 / 76.19 & 0.813 / 0.812 / 0.900 \\
 &  & Intensifier & 0.665 / 0.667 / 0.814 & 0.351 / 0.342 / 0.667 & 70.09 / 70.03 / 82.58 & 53.18 / 53.00 / 75.35 & 0.819 / 0.810 / 0.900 \\
 &  & Expansion (No Impact) & 0.671 / 0.656 / 0.814 & 0.358 / 0.323 / 0.663 & 71.04 / 69.61 / 82.59 & 54.12 / 51.99 / 75.40 & 0.813 / 0.814 / 0.898 \\
 \cmidrule{2-8}
 
 & \multirow{3}{*}{\hlred{Critical}} & Expansion (Impact) & 0.642 / 0.638 / 0.796 & 0.318 / 0.303 / 0.650 & 68.66 / 67.97 / 80.89 & 50.18 / 49.49 / 73.17 & 0.798 / 0.795 / 0.888 \\
 &  & Omission & 0.586 / 0.599 / 0.718 & 0.305 / 0.309 / 0.590 & 61.60 / 62.41 / 72.45 & 46.83 / 47.59 / 66.75 & 0.743 / 0.751 / 0.847 \\
 &  & Alteration & 0.508 / 0.489 / 0.539 & 0.251 / 0.228 / 0.430 & 55.66 / 53.31 / 55.57 & 39.88 / 37.94 / 49.45 & 0.681 / 0.659 / 0.778 \\
\midrule

 \multirow{9}{*}{\textsc{\textbf{En-Zh}}} & \multirow{5}{*}{\hl{Minor}} & Spelling & 0.554 / 0.542 / 0.770 & 0.267 / 0.237 / 0.633 & 61.02 / 59.26 / 79.05 & 42.34 / 40.40 / 70.77 & 0.753 / 0.747 / 0.880 \\
 &  & Word Order & 0.546 / 0.543 / 0.764 & 0.263 / 0.250 / 0.628 & 60.24 / 59.78 / 78.79 & 41.60 / 40.91 / 70.29 & 0.762 / 0.751 / 0.879 \\
 &  & Synonym & 0.552 / 0.541 / 0.762 & 0.266 / 0.237 / 0.626 & 61.31 / 59.28 / 78.26 & 42.02 / 40.64 / 70.35 & 0.754 / 0.753 / 0.876 \\
 &  & Intensifier & 0.548 / 0.541 / 0.762 & 0.257 / 0.237 / 0.631 & 60.91 / 59.59 / 78.46 & 41.15 / 40.09 / 70.60 & 0.763 / 0.755 / 0.879 \\
 &  & Expansion (No Impact) & 0.541 / 0.539 / 0.759 & 0.259 / 0.232 / 0.620 & 59.77 / 60.01 / 78.13 & 41.24 / 39.51 / 70.07 & 0.767 / 0.753 / 0.876 \\
 \cmidrule{2-8}
 
 & \multirow{3}{*}{\hlred{Critical}} & Expansion (Impact) & 0.534 / 0.528 / 0.757 & 0.236 / 0.221 / 0.616 & 59.69 / 58.92 / 78.30 & 38.93 / 37.69 / 69.01 & 0.750 / 0.742 / 0.872 \\
 &  & Omission & 0.480 / 0.489 / 0.658 & 0.224 / 0.214 / 0.539 & 53.18 / 53.45 / 68.28 & 36.00 / 36.14 / 60.49 & 0.695 / 0.699 / 0.817 \\
 &  & Alteration & 0.420 / 0.394 / 0.470 & 0.182 / 0.148 / 0.357 & 47.79 / 45.69 / 49.43 & 30.98 / 27.86 / 42.00 & 0.637 / 0.611 / 0.758 \\

    \specialrule{1.3pt}{0pt}{0pt}
    \end{tabular}

}
    \caption{Detailed results of \askqe using \textsc{LLaMA-3 70b}. Each cell presents scores for Vanilla / SRL / NLI variant.}
    \label{tab:detailed_llama_70b}
\end{table*}

\begin{table*}[h]
    \centering
    \resizebox{\textwidth}{!}{%
    \begin{tabular}{clllllll}
    \specialrule{1.3pt}{0pt}{0pt}
    \textbf{Language} & \textbf{Severity} & \textbf{Perturbation} & \textbf{\textsc{F1}} & \textbf{\textsc{EM}} & \textbf{\textsc{chrF}} & \textbf{\textsc{BLEU}} & \textbf{\textsc{SBERT}} \\
    \toprule
    
    \multirow{8}{*}{\textsc{\textbf{En-Es}}} & \multirow{5}{*}{\hl{Minor}} & Spelling & 0.645 / 0.621 / 0.805 & 0.339 / 0.299 / 0.674 & 70.55 / 67.55 / 82.74 & 52.55 / 49.27 / 75.16 & 0.824 / 0.804 / 0.891 \\
 &  & Word Order & 0.620 / 0.616 / 0.793 & 0.313 / 0.296 / 0.660 & 68.10 / 67.09 / 81.57 & 49.61 / 48.17 / 73.86 & 0.809 / 0.799 / 0.886 \\
 &  & Synonym & 0.617 / 0.614 / 0.783 & 0.307 / 0.293 / 0.650 & 67.45 / 65.88 / 80.71 & 49.47 / 48.51 / 73.09 & 0.807 / 0.794 / 0.885 \\
 &  & Intensifier & 0.615 / 0.601 / 0.798 & 0.293 / 0.257 / 0.658 & 67.64 / 65.41 / 82.27 & 47.89 / 45.42 / 73.99 & 0.808 / 0.793 / 0.891 \\
 &  & Expansion (No Impact) & 0.609 / 0.605 / 0.795 & 0.294 / 0.265 / 0.655 & 67.33 / 65.99 / 82.15 & 47.76 / 46.39 / 73.82 & 0.806 / 0.792 / 0.892 \\
 \cmidrule{2-8}
 & \multirow{3}{*}{\hlred{Critical}} & Expansion (Impact) & 0.600 / 0.577 / 0.801 & 0.284 / 0.249 / 0.661 & 66.65 / 63.40 / 82.03 & 46.33 / 42.99 / 73.75 & 0.798 / 0.771 / 0.896 \\
 &  & Omission & 0.570 / 0.575 / 0.721 & 0.272 / 0.258 / 0.594 & 62.44 / 62.31 / 74.94 & 44.70 / 44.07 / 66.76 & 0.770 / 0.770 / 0.855 \\
 &  & Alteration & 0.554 / 0.529 / 0.536 & 0.264 / 0.223 / 0.398 & 61.54 / 59.03 / 55.66 & 43.47 / 39.92 / 47.98 & 0.750 / 0.732 / 0.804 \\
\midrule
 \multirow{8}{*}{\textsc{\textbf{En-Fr}}} & \multirow{5}{*}{\hl{Minor}} & Spelling & 0.644 / 0.611 / 0.808 & 0.345 / 0.289 / 0.678 & 69.91 / 66.67 / 83.09 & 52.63 / 48.09 / 75.38 & 0.830 / 0.800 / 0.894 \\
 &  & Word Order & 0.618 / 0.598 / 0.798 & 0.300 / 0.276 / 0.676 & 67.29 / 64.43 / 81.75 & 48.71 / 46.59 / 74.80 & 0.808 / 0.790 / 0.894 \\
 &  & Synonym & 0.604 / 0.597 / 0.798 & 0.293 / 0.282 / 0.672 & 65.45 / 64.78 / 81.91 & 48.10 / 46.14 / 74.44 & 0.795 / 0.788 / 0.889 \\
 &  & Intensifier & 0.611 / 0.599 / 0.790 & 0.294 / 0.268 / 0.657 & 67.68 / 65.17 / 81.84 & 47.99 / 45.69 / 73.51 & 0.812 / 0.799 / 0.904 \\
 &  & Expansion (No Impact) & 0.617 / 0.607 / 0.793 & 0.305 / 0.275 / 0.660 & 68.17 / 65.84 / 81.51 & 48.99 / 46.76 / 73.60 & 0.807 / 0.789 / 0.891 \\
 \cmidrule{2-8}
 & \multirow{3}{*}{\hlred{Critical}} & Expansion (Impact) & 0.589 / 0.566 / 0.779 & 0.271 / 0.244 / 0.646 & 65.11 / 62.36 / 80.18 & 45.27 / 42.07 / 71.93 & 0.787 / 0.763 / 0.881 \\
 &  & Omission & 0.582 / 0.568 / 0.746 & 0.275 / 0.250 / 0.610 & 63.07 / 61.11 / 76.52 & 45.04 / 43.06 / 69.11 & 0.777 / 0.763 / 0.866 \\
 &  & Alteration & 0.539 / 0.505 / 0.480 & 0.249 / 0.218 / 0.371 & 59.91 / 56.09 / 49.86 & 41.64 / 38.13 / 43.31 & 0.744 / 0.712 / 0.776 \\
\midrule
 \multirow{8}{*}{\textsc{\textbf{En-Hi}}} & \multirow{5}{*}{\hl{Minor}} & Spelling & 0.641 / 0.615 / 0.806 & 0.351 / 0.311 / 0.686 & 71.28 / 68.15 / 82.93 & 53.16 / 49.54 / 75.96 & 0.824 / 0.802 / 0.891 \\
 &  & Word Order & 0.615 / 0.597 / 0.769 & 0.313 / 0.285 / 0.654 & 67.51 / 65.06 / 80.23 & 50.06 / 47.21 / 72.37 & 0.784 / 0.765 / 0.872 \\
 &  & Synonym & 0.600 / 0.571 / 0.753 & 0.308 / 0.270 / 0.640 & 66.07 / 63.36 / 78.06 & 47.99 / 44.53 / 70.65 & 0.794 / 0.779 / 0.879 \\
 &  & Intensifier & 0.620 / 0.601 / 0.799 & 0.310 / 0.291 / 0.680 & 69.76 / 66.25 / 82.74 & 50.12 / 47.44 / 74.77 & 0.808 / 0.788 / 0.897 \\
 &  & Expansion (No Impact) & 0.618 / 0.598 / 0.790 & 0.301 / 0.277 / 0.664 & 68.95 / 66.42 / 82.00 & 49.18 / 46.81 / 74.04 & 0.815 / 0.789 / 0.902 \\
 \cmidrule{2-8}
 & \multirow{3}{*}{\hlred{Critical}} & Expansion (Impact) & 0.596 / 0.567 / 0.761 & 0.293 / 0.269 / 0.640 & 66.37 / 63.50 / 79.41 & 46.56 / 43.97 / 70.59 & 0.791 / 0.761 / 0.872 \\
 &  & Omission & 0.543 / 0.538 / 0.692 & 0.267 / 0.249 / 0.580 & 60.63 / 59.73 / 72.32 & 43.16 / 42.00 / 64.47 & 0.742 / 0.735 / 0.843 \\
 &  & Alteration & 0.581 / 0.551 / 0.652 & 0.290 / 0.255 / 0.538 & 65.03 / 61.03 / 67.53 & 46.63 / 43.17 / 60.51 & 0.776 / 0.745 / 0.837 \\
\midrule
 \multirow{8}{*}{\textsc{\textbf{En-Tl}}} & \multirow{5}{*}{\hl{Minor}} & Spelling & 0.663 / 0.660 / 0.823 & 0.355 / 0.348 / 0.695 & 72.59 / 71.03 / 84.25 & 54.89 / 54.64 / 77.29 & 0.835 / 0.820 / 0.911 \\
 &  & Word Order & 0.652 / 0.633 / 0.813 & 0.357 / 0.322 / 0.696 & 70.24 / 68.05 / 83.03 & 54.39 / 51.85 / 76.70 & 0.821 / 0.792 / 0.895 \\
 &  & Synonym & 0.648 / 0.624 / 0.799 & 0.343 / 0.329 / 0.678 & 70.53 / 67.91 / 82.42 & 52.74 / 50.94 / 75.43 & 0.817 / 0.802 / 0.899 \\
 &  & Intensifier & 0.641 / 0.637 / 0.817 & 0.330 / 0.310 / 0.684 & 70.35 / 69.08 / 83.88 & 51.85 / 50.79 / 76.62 & 0.829 / 0.805 / 0.899 \\
 &  & Expansion (No Impact) & 0.654 / 0.632 / 0.809 & 0.337 / 0.309 / 0.670 & 71.81 / 68.96 / 83.72 & 53.26 / 50.68 / 75.47 & 0.821 / 0.806 / 0.905 \\
 \cmidrule{2-8}
 & \multirow{3}{*}{\hlred{Critical}} & Expansion (Impact) & 0.622 / 0.615 / 0.804 & 0.301 / 0.293 / 0.676 & 68.96 / 67.23 / 82.54 & 49.02 / 48.34 / 74.88 & 0.806 / 0.784 / 0.896 \\
 &  & Omission & 0.578 / 0.593 / 0.703 & 0.300 / 0.292 / 0.580 & 63.24 / 63.81 / 72.53 & 46.60 / 47.47 / 65.66 & 0.757 / 0.766 / 0.852 \\
 &  & Alteration & 0.598 / 0.573 / 0.625 & 0.306 / 0.271 / 0.504 & 65.72 / 62.59 / 64.52 & 48.58 / 45.67 / 57.47 & 0.777 / 0.748 / 0.834 \\
\midrule
 \multirow{8}{*}{\textsc{\textbf{En-Zh}}} & \multirow{5}{*}{\hl{Minor}} & Spelling & 0.545 / 0.521 / 0.752 & 0.255 / 0.219 / 0.635 & 61.75 / 58.97 / 77.93 & 42.92 / 39.78 / 70.42 & 0.762 / 0.745 / 0.878 \\
 &  & Word Order & 0.544 / 0.520 / 0.730 & 0.247 / 0.222 / 0.610 & 61.29 / 58.35 / 76.87 & 42.35 / 39.50 / 68.02 & 0.762 / 0.743 / 0.864 \\
 &  & Synonym & 0.540 / 0.517 / 0.750 & 0.253 / 0.221 / 0.636 & 61.53 / 58.65 / 78.45 & 42.34 / 39.34 / 70.47 & 0.760 / 0.745 / 0.871 \\
 &  & Intensifier & 0.551 / 0.519 / 0.738 & 0.263 / 0.212 / 0.617 & 63.10 / 58.76 / 77.48 & 43.09 / 38.88 / 68.64 & 0.766 / 0.739 / 0.875 \\
 &  & Expansion (No Impact) & 0.545 / 0.518 / 0.752 & 0.245 / 0.220 / 0.640 & 62.26 / 58.24 / 78.95 & 41.84 / 38.81 / 70.36 & 0.775 / 0.745 / 0.865 \\
 \cmidrule{2-8}
 & \multirow{3}{*}{\hlred{Critical}} & Expansion (Impact) & 0.519 / 0.488 / 0.748 & 0.221 / 0.193 / 0.630 & 59.89 / 55.84 / 77.86 & 38.49 / 35.32 / 69.24 & 0.751 / 0.715 / 0.869 \\
 &  & Omission & 0.499 / 0.485 / 0.685 & 0.228 / 0.201 / 0.571 & 56.80 / 54.83 / 71.11 & 38.67 / 36.53 / 63.60 & 0.726 / 0.712 / 0.832 \\
 &  & Alteration & 0.500 / 0.472 / 0.543 & 0.217 / 0.180 / 0.417 & 57.58 / 54.42 / 56.70 & 38.14 / 34.93 / 49.21 & 0.726 / 0.706 / 0.813 \\

    \specialrule{1.3pt}{0pt}{0pt}
    \end{tabular}

}
    \caption{Detailed results of \askqe using \textsc{Yi-1.5 9b}. Each cell presents scores for Vanilla / SRL / NLI variant.}
    \label{tab:detailed_yi_9b}
\end{table*}

\clearpage

\begin{table*}[!htp]
\centering
\resizebox{\linewidth}{!}{%
    \begin{tabular}{cllllllll}
    \specialrule{1.3pt}{0pt}{0pt}
    \textbf{Language} & \textbf{QE Metric} & \textbf{\textsc{F1}} & \textbf{\textsc{EM}} & \textbf{\textsc{chrF}} & \textbf{\textsc{BLEU}} & \textbf{\textsc{SBERT}}\\
    \toprule

    \multirow{3}{*}{\textsc{\textbf{En-Es}}} & \textsc{xCOMET-QE} & 0.862 / 0.892 / 0.849 & 0.608 / 0.733 / 0.695 & 0.845 / 0.874 / 0.832 & 0.749 / 0.794 / 0.751 & 0.944 / 0.947 / 0.902 \\
 & \textsc{MetricX-QE} & -0.819 / -0.846 / -0.812 & -0.530 / -0.658 / -0.630 & -0.814 / -0.838 / -0.804 & -0.681 / -0.724 / -0.689 & -0.925 / -0.931 / -0.893 \\
 & \textsc{BT-Score} & 0.970 / 0.973 / 0.955 & 0.792 / 0.876 / 0.848 & 0.952 / 0.960 / 0.948 & 0.911 / 0.932 / 0.907 & 0.982 / 0.976 / 0.958 \\
\midrule
\multirow{3}{*}{\textsc{\textbf{En-Fr}}} & \textsc{xCOMET-QE} & 0.936 / 0.920 / 0.912 & 0.832 / 0.804 / 0.787 & 0.930 / 0.910 / 0.899 & 0.882 / 0.850 / 0.844 & 0.990 / 0.939 / 0.987 \\
 & \textsc{MetricX-QE} & -0.862 / -0.837 / -0.839 & -0.734 / -0.692 / -0.685 & -0.864 / -0.842 / -0.841 & -0.789 / -0.749 / -0.752 & -0.948 / -0.863 / -0.960 \\
 & \textsc{BT-Score} & 0.960 / 0.959 / 0.942 & 0.879 / 0.869 / 0.836 & 0.949 / 0.945 / 0.926 & 0.927 / 0.917 / 0.899 & 0.990 / 0.966 / 0.992 \\
\midrule
\multirow{3}{*}{\textsc{\textbf{En-Hi}}} & \textsc{xCOMET-QE} & 0.768 / 0.741 / 0.667 & 0.805 / 0.766 / 0.740 & 0.710 / 0.668 / 0.612 & 0.818 / 0.814 / 0.740 & 0.748 / 0.703 / 0.610 \\
 & \textsc{MetricX-QE} & -0.934 / -0.916 / -0.865 & -0.853 / -0.884 / -0.833 & -0.896 / -0.879 / -0.838 & -0.894 / -0.892 / -0.843 & -0.898 / -0.858 / -0.809 \\
 & \textsc{BT-Score} & 0.959 / 0.960 / 0.949 & 0.855 / 0.866 / 0.874 & 0.950 / 0.940 / 0.924 & 0.943 / 0.955 / 0.945 & 0.938 / 0.932 / 0.896 \\
\midrule
\multirow{3}{*}{\textsc{\textbf{En-Tl}}} & \textsc{xCOMET-QE} & 0.795 / 0.818 / 0.697 & 0.791 / 0.800 / 0.709 & 0.709 / 0.700 / 0.604 & 0.815 / 0.818 / 0.708 & 0.831 / 0.816 / 0.698 \\
 & \textsc{MetricX-QE} & -0.821 / -0.834 / -0.805 & -0.660 / -0.657 / -0.660 & -0.813 / -0.808 / -0.768 & -0.743 / -0.742 / -0.732 & -0.929 / -0.920 / -0.867 \\
 & \textsc{BT-Score} & 0.936 / 0.929 / 0.916 & 0.846 / 0.827 / 0.847 & 0.908 / 0.909 / 0.880 & 0.931 / 0.936 / 0.930 & 0.905 / 0.912 / 0.895 \\
\midrule
\multirow{3}{*}{\textsc{\textbf{En-Zh}}} & \textsc{xCOMET-QE} & 0.729 / 0.812 / 0.728 & 0.706 / 0.847 / 0.759 & 0.692 / 0.759 / 0.690 & 0.675 / 0.785 / 0.704 & 0.764 / 0.769 / 0.727 \\
 & \textsc{MetricX-QE} & -0.658 / -0.755 / -0.656 & -0.559 / -0.709 / -0.630 & -0.668 / -0.751 / -0.674 & -0.536 / -0.654 / -0.578 & -0.721 / -0.733 / -0.702 \\
 & \textsc{BT-Score} & 0.867 / 0.915 / 0.863 & 0.767 / 0.847 / 0.803 & 0.870 / 0.911 / 0.866 & 0.786 / 0.859 / 0.805 & 0.906 / 0.907 / 0.903 \\

    \specialrule{1.3pt}{0pt}{0pt}
    \end{tabular}
}
\caption{Correlation analysis of \askqe using \textsc{Gemma-2 9b}. Each cell presents Pearson correlation coefficients for Vanilla / SRL / NLI. \textsc{xCOMET-QE} (↑), \textsc{MetricX-QE} (↓): Computed between source $X_{\mathrm{src}}$ and perturbed MT output $Y_{\mathrm{tgt}}$; BT-Score (↑): Computed between source $X_{\mathrm{src}}$ and backtranslated MT output $Y_{\mathrm{bt}}$. \textbf{\textsc{SBERT}}: \textsc{SentenceBERT}.}
\label{tab:correlation_gemma_9b}
\end{table*}
\begin{table*}[!htp]
\centering
\resizebox{\linewidth}{!}{%
    \begin{tabular}{cllllllll}
    \specialrule{1.3pt}{0pt}{0pt}
    \textbf{Language} & \textbf{QE Metric} & \textbf{\textsc{F1}} & \textbf{\textsc{EM}} & \textbf{\textsc{chrF}} & \textbf{\textsc{BLEU}} & \textbf{\textsc{SBERT}}\\
    \toprule
    
    \multirow{3}{*}{\textsc{\textbf{En-Es}}} & \textsc{xCOMET-QE} & 0.843 / 0.864 / 0.851 & 0.700 / 0.680 / 0.676 & 0.819 / 0.833 / 0.806 & 0.746 / 0.748 / 0.737 & 0.905 / 0.936 / 0.906 \\
 & \textsc{MetricX-QE} & -0.801 / -0.816 / -0.812 & -0.628 / -0.602 / -0.599 & -0.793 / -0.800 / -0.787 & -0.680 / -0.678 / -0.669 & -0.890 / -0.921 / -0.904 \\
 & \textsc{BT-Score} & 0.963 / 0.965 / 0.963 & 0.860 / 0.832 / 0.837 & 0.941 / 0.948 / 0.926 & 0.914 / 0.897 / 0.902 & 0.972 / 0.976 / 0.957 \\
\midrule
\multirow{3}{*}{\textsc{\textbf{En-Fr}}} & \textsc{xCOMET-QE} & 0.940 / 0.925 / 0.923 & 0.843 / 0.804 / 0.799 & 0.916 / 0.898 / 0.891 & 0.880 / 0.856 / 0.851 & 0.970 / 0.983 / 0.966 \\
 & \textsc{MetricX-QE} & -0.885 / -0.861 / -0.874 & -0.762 / -0.710 / -0.719 & -0.870 / -0.849 / -0.851 & -0.806 / -0.767 / -0.777 & -0.937 / -0.962 / -0.943 \\
 & \textsc{BT-Score} & 0.966 / 0.953 / 0.951 & 0.884 / 0.843 / 0.834 & 0.947 / 0.929 / 0.926 & 0.931 / 0.905 / 0.904 & 0.990 / 0.991 / 0.986 \\
\midrule
\multirow{3}{*}{\textsc{\textbf{En-Hi}}} & \textsc{xCOMET-QE} & 0.683 / 0.644 / 0.573 & 0.740 / 0.694 / 0.740 & 0.626 / 0.591 / 0.527 & 0.757 / 0.730 / 0.681 & 0.668 / 0.674 / 0.550 \\
 & \textsc{MetricX-QE} & -0.894 / -0.837 / -0.805 & -0.870 / -0.829 / -0.858 & -0.852 / -0.803 / -0.766 & -0.875 / -0.825 / -0.815 & -0.862 / -0.833 / -0.754 \\
 & \textsc{BT-Score} & 0.959 / 0.945 / 0.926 & 0.882 / 0.859 / 0.907 & 0.937 / 0.915 / 0.899 & 0.960 / 0.939 / 0.946 & 0.940 / 0.914 / 0.889 \\
\midrule
\multirow{3}{*}{\textsc{\textbf{En-Tl}}} & \textsc{xCOMET-QE} & 0.658 / 0.773 / 0.670 & 0.729 / 0.769 / 0.749 & 0.561 / 0.614 / 0.547 & 0.683 / 0.777 / 0.696 & 0.656 / 0.743 / 0.621 \\
 & \textsc{MetricX-QE} & -0.772 / -0.801 / -0.758 & -0.684 / -0.628 / -0.656 & -0.731 / -0.750 / -0.711 & -0.707 / -0.694 / -0.685 & -0.842 / -0.872 / -0.808 \\
 & \textsc{BT-Score} & 0.942 / 0.955 / 0.952 & 0.890 / 0.791 / 0.872 & 0.896 / 0.918 / 0.894 & 0.953 / 0.922 / 0.953 & 0.885 / 0.938 / 0.895 \\
\midrule
\multirow{3}{*}{\textsc{\textbf{En-Zh}}} & \textsc{xCOMET-QE} & 0.602 / 0.743 / 0.716 & 0.600 / 0.727 / 0.698 & 0.560 / 0.647 / 0.601 & 0.575 / 0.700 / 0.665 & 0.566 / 0.699 / 0.629 \\
 & \textsc{MetricX-QE} & -0.616 / -0.711 / -0.716 & -0.533 / -0.601 / -0.642 & -0.618 / -0.682 / -0.660 & -0.525 / -0.587 / -0.608 & -0.601 / -0.691 / -0.657 \\
 & \textsc{BT-Score} & 0.835 / 0.902 / 0.909 & 0.740 / 0.758 / 0.814 & 0.821 / 0.875 / 0.859 & 0.791 / 0.817 / 0.848 & 0.812 / 0.882 / 0.858 \\

    \specialrule{1.3pt}{0pt}{0pt}
    \end{tabular}
}
\caption{Correlation analysis of \askqe using \textsc{Gemma-2 27b}. Each cell presents Pearson correlation for Vanilla / SRL / NLI.}
\label{tab:correlation_gemma_27b}
\end{table*}
\begin{table*}[!htp]
\centering
\resizebox{\linewidth}{!}{%
    \begin{tabular}{cllllllll}
    \specialrule{1.3pt}{0pt}{0pt}
    \textbf{Language} & \textbf{QE Metric} & \textbf{\textsc{F1}} & \textbf{\textsc{EM}} & \textbf{\textsc{chrF}} & \textbf{\textsc{BLEU}} & \textbf{\textsc{SBERT}}\\
    \toprule

    \multirow{3}{*}{\textsc{\textbf{En-Es}}} & \textsc{xCOMET-QE} & 0.740 / 0.887 / 0.957 & 0.567 / 0.729 / 0.960 & 0.716 / 0.844 / 0.962 & 0.642 / 0.785 / 0.962 & 0.890 / 0.935 / 0.946 \\
 & \textsc{MetricX-QE} & -0.703 / -0.842 / -0.937 & -0.502 / -0.668 / -0.934 & -0.692 / -0.806 / -0.940 & -0.585 / -0.722 / -0.938 & -0.899 / -0.909 / -0.950 \\
 & \textsc{BT-Score} & 0.904 / 0.978 / 0.928 & 0.796 / 0.847 / 0.933 & 0.877 / 0.956 / 0.932 & 0.853 / 0.922 / 0.939 & 0.935 / 0.975 / 0.910 \\
\midrule
\multirow{3}{*}{\textsc{\textbf{En-Fr}}} & \textsc{xCOMET-QE} & 0.802 / 0.916 / 0.984 & 0.599 / 0.821 / 0.988 & 0.764 / 0.883 / 0.980 & 0.713 / 0.846 / 0.986 & 0.910 / 0.953 / 0.986 \\
 & \textsc{MetricX-QE} & -0.719 / -0.856 / -0.945 & -0.481 / -0.753 / -0.955 & -0.703 / -0.829 / -0.939 & -0.619 / -0.774 / -0.948 & -0.933 / -0.928 / -0.960 \\
 & \textsc{BT-Score} & 0.861 / 0.963 / 0.965 & 0.707 / 0.878 / 0.971 & 0.830 / 0.936 / 0.964 & 0.804 / 0.918 / 0.970 & 0.933 / 0.981 / 0.973 \\
\midrule
\multirow{3}{*}{\textsc{\textbf{En-Hi}}} & \textsc{xCOMET-QE} & 0.559 / 0.761 / 0.714 & 0.563 / 0.739 / 0.729 & 0.541 / 0.697 / 0.714 & 0.644 / 0.789 / 0.740 & 0.499 / 0.720 / 0.675 \\
 & \textsc{MetricX-QE} & -0.781 / -0.888 / -0.959 & -0.701 / -0.853 / -0.971 & -0.743 / -0.870 / -0.962 & -0.766 / -0.859 / -0.970 & -0.701 / -0.858 / -0.895 \\
 & \textsc{BT-Score} & 0.929 / 0.990 / 0.873 & 0.818 / 0.920 / 0.871 & 0.908 / 0.957 / 0.858 & 0.929 / 0.974 / 0.889 & 0.860 / 0.943 / 0.877 \\
\midrule
\multirow{3}{*}{\textsc{\textbf{En-Tl}}} & \textsc{xCOMET-QE} & 0.535 / 0.749 / 0.785 & 0.645 / 0.766 / 0.812 & 0.453 / 0.672 / 0.775 & 0.562 / 0.756 / 0.810 & 0.557 / 0.816 / 0.747 \\
 & \textsc{MetricX-QE} & -0.682 / -0.781 / -0.929 & -0.608 / -0.700 / -0.943 & -0.641 / -0.772 / -0.928 & -0.614 / -0.710 / -0.942 & -0.771 / -0.892 / -0.919 \\
 & \textsc{BT-Score} & 0.898 / 0.950 / 0.852 & 0.884 / 0.853 / 0.850 & 0.859 / 0.922 / 0.843 & 0.916 / 0.945 / 0.856 & 0.826 / 0.939 / 0.869 \\
\midrule
\multirow{3}{*}{\textsc{\textbf{En-Zh}}} & \textsc{xCOMET-QE} & 0.619 / 0.694 / 0.872 & 0.596 / 0.745 / 0.872 & 0.552 / 0.601 / 0.860 & 0.576 / 0.628 / 0.872 & 0.625 / 0.635 / 0.782 \\
 & \textsc{MetricX-QE} & -0.594 / -0.601 / -0.841 & -0.473 / -0.621 / -0.832 & -0.585 / -0.534 / -0.835 & -0.462 / -0.468 / -0.828 & -0.674 / -0.549 / -0.784 \\
 & \textsc{BT-Score} & 0.865 / 0.857 / 0.845 & 0.715 / 0.816 / 0.836 & 0.826 / 0.809 / 0.845 & 0.765 / 0.749 / 0.847 & 0.880 / 0.832 / 0.902 \\

    \specialrule{1.3pt}{0pt}{0pt}
    \end{tabular}
}
\caption{Correlation analysis of \askqe using \textsc{LLaMA-3 8b}. Each cell presents Pearson correlation for Vanilla / SRL / NLI.}
\label{tab:correlation_llama_8b}
\end{table*}
\begin{table*}[!htp]
\centering
\resizebox{\linewidth}{!}{%
    \begin{tabular}{cllllllll}
    \specialrule{1.3pt}{0pt}{0pt}
    \textbf{Language} & \textbf{QE Metric} & \textbf{\textsc{F1}} & \textbf{\textsc{EM}} & \textbf{\textsc{chrF}} & \textbf{\textsc{BLEU}} & \textbf{\textsc{SBERT}}\\
    \toprule

    \multirow{3}{*}{\textsc{\textbf{En-Es}}} & \textsc{xCOMET-QE} & 0.954 / 0.956 / 0.969 & 0.918 / 0.858 / 0.962 & 0.951 / 0.950 / 0.967 & 0.937 / 0.914 / 0.967 & 0.989 / 0.991 / 0.983 \\
 & \textsc{MetricX-QE} & -0.919 / -0.913 / -0.936 & -0.864 / -0.793 / -0.924 & -0.921 / -0.912 / -0.937 & -0.888 / -0.855 / -0.930 & -0.965 / -0.963 / -0.964 \\
 & \textsc{BT-Score} & 0.971 / 0.958 / 0.943 & 0.974 / 0.916 / 0.956 & 0.963 / 0.955 / 0.942 & 0.984 / 0.958 / 0.953 & 0.958 / 0.955 / 0.955 \\
\midrule
\multirow{3}{*}{\textsc{\textbf{En-Fr}}} & \textsc{xCOMET-QE} & 0.969 / 0.968 / 0.972 & 0.916 / 0.925 / 0.968 & 0.963 / 0.960 / 0.970 & 0.956 / 0.952 / 0.972 & 0.990 / 0.993 / 0.989 \\
 & \textsc{MetricX-QE} & -0.917 / -0.908 / -0.926 & -0.831 / -0.843 / -0.920 & -0.914 / -0.904 / -0.925 & -0.888 / -0.878 / -0.922 & -0.954 / -0.953 / -0.962 \\
 & \textsc{BT-Score} & 0.952 / 0.960 / 0.952 & 0.908 / 0.922 / 0.952 & 0.952 / 0.953 / 0.951 & 0.954 / 0.956 / 0.955 & 0.972 / 0.978 / 0.971 \\
\midrule
\multirow{3}{*}{\textsc{\textbf{En-Hi}}} & \textsc{xCOMET-QE} & 0.749 / 0.719 / 0.697 & 0.810 / 0.757 / 0.720 & 0.715 / 0.705 / 0.687 & 0.814 / 0.782 / 0.727 & 0.723 / 0.701 / 0.674 \\
 & \textsc{MetricX-QE} & -0.963 / -0.969 / -0.961 & -0.951 / -0.942 / -0.969 & -0.935 / -0.958 / -0.955 & -0.962 / -0.975 / -0.973 & -0.923 / -0.940 / -0.916 \\
 & \textsc{BT-Score} & 0.916 / 0.850 / 0.828 & 0.903 / 0.813 / 0.838 & 0.908 / 0.851 / 0.820 & 0.939 / 0.876 / 0.849 & 0.884 / 0.813 / 0.861 \\
\midrule
\multirow{3}{*}{\textsc{\textbf{En-Tl}}} & \textsc{xCOMET-QE} & 0.823 / 0.883 / 0.864 & 0.849 / 0.902 / 0.882 & 0.770 / 0.835 / 0.852 & 0.855 / 0.904 / 0.881 & 0.843 / 0.877 / 0.863 \\
 & \textsc{MetricX-QE} & -0.919 / -0.925 / -0.949 & -0.874 / -0.830 / -0.941 & -0.901 / -0.914 / -0.951 & -0.905 / -0.888 / -0.948 & -0.968 / -0.964 / -0.972 \\
 & \textsc{BT-Score} & 0.884 / 0.867 / 0.809 & 0.887 / 0.856 / 0.819 & 0.877 / 0.871 / 0.805 & 0.917 / 0.905 / 0.822 & 0.858 / 0.846 / 0.824 \\
\midrule
\multirow{3}{*}{\textsc{\textbf{En-Zh}}} & \textsc{xCOMET-QE} & 0.793 / 0.866 / 0.854 & 0.806 / 0.856 / 0.859 & 0.775 / 0.831 / 0.856 & 0.796 / 0.857 / 0.858 & 0.802 / 0.851 / 0.820 \\
 & \textsc{MetricX-QE} & -0.799 / -0.833 / -0.842 & -0.747 / -0.798 / -0.839 & -0.801 / -0.834 / -0.846 & -0.765 / -0.787 / -0.838 & -0.823 / -0.828 / -0.822 \\
 & \textsc{BT-Score} & 0.868 / 0.862 / 0.854 & 0.861 / 0.810 / 0.844 & 0.870 / 0.869 / 0.851 & 0.872 / 0.845 / 0.854 & 0.880 / 0.858 / 0.882 \\

    \specialrule{1.3pt}{0pt}{0pt}
    \end{tabular}
}
\caption{Correlation analysis of \askqe using \textsc{LLaMA-3 70b}. Each cell presents Pearson correlation for Vanilla / SRL / NLI.}
\label{tab:correlation_llama_70b}
\end{table*}
\begin{table*}[!htp]
\centering
\resizebox{\linewidth}{!}{%
    \begin{tabular}{cllllllll}
    \specialrule{1.3pt}{0pt}{0pt}
    \textbf{Language} & \textbf{QE Metric} & \textbf{\textsc{F1}} & \textbf{\textsc{EM}} & \textbf{\textsc{chrF}} & \textbf{\textsc{BLEU}} & \textbf{\textsc{SBERT}}\\
    \toprule

    \multirow{3}{*}{\textsc{\textbf{En-Es}}} & \textsc{xCOMET-QE} & 0.868 / 0.914 / 0.885 & 0.734 / 0.794 / 0.755 & 0.854 / 0.901 / 0.838 & 0.788 / 0.850 / 0.811 & 0.912 / 0.939 / 0.911 \\
 & \textsc{MetricX-QE} & -0.825 / -0.856 / -0.835 & -0.659 / -0.715 / -0.684 & -0.820 / -0.851 / -0.802 & -0.721 / -0.775 / -0.738 & -0.879 / -0.888 / -0.871 \\
 & \textsc{BT-Score} & 0.968 / 0.974 / 0.959 & 0.876 / 0.908 / 0.874 & 0.959 / 0.975 / 0.940 & 0.931 / 0.950 / 0.929 & 0.978 / 0.981 / 0.969 \\
\midrule
\multirow{3}{*}{\textsc{\textbf{En-Fr}}} & \textsc{xCOMET-QE} & 0.932 / 0.947 / 0.950 & 0.833 / 0.870 / 0.882 & 0.914 / 0.941 / 0.935 & 0.874 / 0.899 / 0.903 & 0.874 / 0.941 / 0.912 \\
 & \textsc{MetricX-QE} & -0.864 / -0.874 / -0.887 & -0.734 / -0.768 / -0.793 & -0.857 / -0.875 / -0.879 & -0.784 / -0.807 / -0.817 & -0.861 / -0.877 / -0.859 \\
 & \textsc{BT-Score} & 0.954 / 0.967 / 0.968 & 0.859 / 0.907 / 0.911 & 0.943 / 0.964 / 0.956 & 0.913 / 0.938 / 0.939 & 0.923 / 0.961 / 0.935 \\
\midrule
\multirow{3}{*}{\textsc{\textbf{En-Hi}}} & \textsc{xCOMET-QE} & 0.743 / 0.783 / 0.690 & 0.800 / 0.815 / 0.792 & 0.704 / 0.728 / 0.608 & 0.813 / 0.862 / 0.788 & 0.747 / 0.740 / 0.704 \\
 & \textsc{MetricX-QE} & -0.951 / -0.969 / -0.907 & -0.898 / -0.920 / -0.908 & -0.916 / -0.928 / -0.851 & -0.934 / -0.952 / -0.913 & -0.948 / -0.921 / -0.936 \\
 & \textsc{BT-Score} & 0.953 / 0.931 / 0.950 & 0.885 / 0.864 / 0.873 & 0.952 / 0.931 / 0.914 & 0.955 / 0.934 / 0.959 & 0.826 / 0.686 / 0.723 \\
\midrule
\multirow{3}{*}{\textsc{\textbf{En-Tl}}} & \textsc{xCOMET-QE} & 0.815 / 0.853 / 0.761 & 0.828 / 0.847 / 0.816 & 0.731 / 0.779 / 0.651 & 0.834 / 0.857 / 0.806 & 0.854 / 0.908 / 0.863 \\
 & \textsc{MetricX-QE} & -0.869 / -0.848 / -0.829 & -0.758 / -0.719 / -0.727 & -0.838 / -0.836 / -0.784 & -0.808 / -0.781 / -0.770 & -0.915 / -0.912 / -0.908 \\
 & \textsc{BT-Score} & 0.933 / 0.949 / 0.955 & 0.852 / 0.875 / 0.895 & 0.924 / 0.947 / 0.923 & 0.935 / 0.941 / 0.952 & 0.905 / 0.834 / 0.846 \\
\midrule
\multirow{3}{*}{\textsc{\textbf{En-Zh}}} & \textsc{xCOMET-QE} & 0.784 / 0.878 / 0.767 & 0.756 / 0.848 / 0.813 & 0.698 / 0.830 / 0.688 & 0.764 / 0.845 / 0.767 & 0.786 / 0.825 / 0.777 \\
 & \textsc{MetricX-QE} & -0.736 / -0.787 / -0.760 & -0.624 / -0.685 / -0.710 & -0.695 / -0.785 / -0.734 & -0.654 / -0.695 / -0.694 & -0.699 / -0.718 / -0.694 \\
 & \textsc{BT-Score} & 0.892 / 0.867 / 0.900 & 0.783 / 0.726 / 0.803 & 0.865 / 0.888 / 0.874 & 0.845 / 0.811 / 0.857 & 0.825 / 0.794 / 0.788 \\

    \specialrule{1.3pt}{0pt}{0pt}
    \end{tabular}
}
\caption{Correlation analysis of \askqe using \textsc{Yi-1.5 9b}. Each cell presents Pearson correlation for Vanilla / SRL / NLI.}
\label{tab:correlation_yi_9b}
\end{table*}
\clearpage

\onecolumn
\section{Question Categorization Prompt}
\label{appendix:categorization_prompt}
We use the annotation guidelines and few-shot examples from \citet{cao-wang-2021-controllable}.
\begin{prompt}[title={Prompt: Question Categorization}]
\textbf{Task:} You will be given a question. Your goal is to annotate the question type. The question type reflects the nature of the question. It is NOT determined by the interrogative word of the question. There are 10 question types in total. The definition for each type is shown in the following, along with examples per question type. During annotation, you can label two most-confident types when no clear decision can be made for the most probable type. Output you answer in Python list format without giving any additional explanation. \\ \\
*** Question Type Starts *** \\
1. Verification: Asking for the truthfulness of an event or a concept. \\
- Is Michael Jackson an African American? \\
- Could stress, anxiety, or worry cause cholesterol levels to rise? \\ \\
2. Disjunctive: Asking for the true one given multiple events or concepts, where comparison among options is not needed. \\
- Is Michael Jackson an African American or Latino? \\
- When you get a spray-on tan does someone put it on you or does a machine do it? \\ \\
3. Concept: Asking for a definition of an event or a concept. \\
- Who said the sun never sets on the British empire? \\
- Where do dolphins have hair at? \\ \\
4. Extent: Asking for the extent or quantity of an event or a concept. \\
- How long does gum stay in your system? \\
- To what extent is the Renewable Fuel Standard accurate nationwide? \\ \\
5. Example: Asking for example(s) or instance(s) of an event or a concept. \\
- What are some examples to support or contradict this? \\
- What countries/regions throughout the world do not celebrate the Christmas holidays? \\ \\
6. Comparison: Asking for comparison among multiple events or concepts. \\
- What is the best tinted facial moisturizer? \\
- In what hilariously inaccurate ways is your job/career portrayed on television or in movies? \\ \\
7. Cause: Asking for the cause or reason for an event or a concept. \\
- Why are parents strick on girls than boys? \\
- What makes nerve agents like 'Novichok' so hard to produce and why can only a handful of laboratories create them? \\ \\
8. Consequence: Asking for the consequences or results of an event. \\
- What are the negative consequences for the services if they do not evaluate their programs? \\
- What would happen if employers violate the legislation? \\ \\
9. Procedural: Asking for the procedures, tools, or methods by which a certain outcome is achieved. \\
- How did the Amish resist assimilation into the current social status in the U.S? \\
- How astronomers detect a nebula when there are no stars illuminating it? \\ \\
10. Judgmental: Asking for the opinions of the answerer's own. \\
- Do you think that it’s acceptable to call off work for a dying-dead pet? \\
- How old is too old for a guy to still live with his mother? \\
*** Question Type Ends *** \\ \\
\textbf{Question:} \texttt{\{question\}} \\
\textbf{Question type(s):}
\end{prompt}

\end{document}